\documentclass{article}

\usepackage[main,final]{neurips_2026}

\usepackage[utf8]{inputenc}
\usepackage[T1]{fontenc}
\usepackage{amsmath}
\usepackage{amsfonts}
\usepackage{graphicx}
\usepackage{booktabs}
\usepackage{nicefrac}
\usepackage{microtype}
\usepackage{xcolor}         
\usepackage{array}
\usepackage{tabularx}
\newcolumntype{Y}{>{\raggedright\arraybackslash}X}
\usepackage{url}
\usepackage{caption}
\usepackage{wrapfig}
\usepackage{array}
\usepackage{tabularx}
\usepackage{makecell}
\usepackage{enumitem}
\usepackage{placeins}
\usepackage{listings}
\usepackage{tikz}
\usetikzlibrary{positioning,calc,fit,arrows.meta}
\usepackage[most]{tcolorbox}
\usepackage{hyperref}
\usepackage[table]{xcolor}
\usepackage{multirow}
\providecommand{\cref}[1]{\ref{#1}}
\providecommand{\Cref}[1]{\ref{#1}}

\definecolor{codegray}{rgb}{0.5,0.5,0.5}
\definecolor{codepurple}{rgb}{0.58,0,0.82}
\definecolor{backcolour}{rgb}{0.95,0.95,0.92}

\usepackage{listings}
\usepackage{xcolor}

\lstdefinestyle{schema}{
  basicstyle=\ttfamily\scriptsize\color{black},
  backgroundcolor=\color{white},
  columns=fullflexible,
  keepspaces=true,
  showstringspaces=false,
  breaklines=true,
  breakatwhitespace=false,
  frame=leftline,
  framerule=0.35pt,
  framesep=6pt,
  xleftmargin=8pt,
  xrightmargin=0pt,
  aboveskip=4pt,
  belowskip=4pt,
  numbers=none,
  numberstyle=\tiny\color{black},
  rulecolor=\color{black!45},
  stringstyle=\color{black},
  keywordstyle=\color{black},
  commentstyle=\color{black},
  identifierstyle=\color{black}
}
\begin{document}

\title{VidUEU-Agent: A Data-Curation Agent for Multimodal Understanding, Editing, and Unified Tasks}

\author{%
Tengju Ru$^{1,*}$ \quad Weitong Lian$^{1,*}$ \quad Zecong Tang$^{2,*,\ddagger}$ \quad Lingyi Meng$^{1,*}$ \\[3pt]
Haoran Li$^{2}$ \quad Zhejun Cui$^{2}$ \quad Yichen Zhu$^{1}$ \quad Hangshuo Cao$^{1}$ \\[3pt]
Qi Kang$^{1}$ \quad Yechi Liu$^{3}$ \quad Kaixuan Wang$^{2}$ \quad Yu-Jie Yuan$^{4}$ \\[3pt]
Chunwei Wang$^{4}$ \quad Yu Zhang$^{1,\dagger}$ \quad Bo Dai$^{2,\dagger}$ \\[5pt]
{\normalfont\footnotesize $^{1}$ Zhejiang University \quad $^{2}$ The University of Hong Kong} \\[2pt]
{\normalfont\footnotesize $^{3}$ Institute of Automation, Chinese Academy of Sciences \quad $^{4}$ Huawei Technologies Ltd.} \\[3pt]
{\normalfont\footnotesize \texttt{zhangyu80@zju.edu.cn} \quad \texttt{bdai@hku.hk}} \\[2pt]
{\normalfont\footnotesize $^{*}$ Equal contribution. \quad $^{\ddagger}$ Project lead. \quad $^{\dagger}$ Corresponding authors.}%
}

\maketitle

\vspace{-2em}
{\centering
\small \textbf{GitHub:} \url{https://github.com/LRuca/VidUEU-Agent}\par
}

\begin{abstract}
The advancement of multimodal learning is heavily constrained by the scarcity and high annotation cost of high-quality instruction data.
While raw videos offer an abundant and dynamic source of visual knowledge, transforming them into precise instruction data remains highly challenging.
To bridge this gap, we present VidUEU-Agent, an automated pipeline that synthesizes instruction data from raw videos for multimodal understanding, editing, and unified tasks.
VidUEU-Agent follows a four-stage pipeline: it performs global perception to filter low-quality shots, mines keyframes and multimodal signals, constructs metadata and routes the metadata to synthesize instruction samples.
More importantly, our agent framework can autonomously iterate the data distribution based on downstream training results, thereby optimizing data quality, balance, and training effectiveness.
Scaling this pipeline, we construct VidUEU-100K, a large-scale dataset of 100K training samples spanning multiple domains and tasks, providing a robust foundation for developing comprehensive multimodal capabilities.
Extensive experiments demonstrate that fine-tuning baseline models on VidUEU-100K yields significant and consistent performance gains across diverse representative benchmarks, validating the effectiveness of our proposed agent framework and dataset.

\end{abstract}

\section{Introduction}
\label{sec:intro}


With the rapid advancement of large language models (LLMs), multimodal large language models (MLLMs) have made significant strides \cite{yin_survey_2024}.
These models use vision encoders to bridge multimodal inputs, enabling joint understanding and reasoning \cite{wu2023multimodal,liang2024comprehensive}.
Concurrently, multimodal generative models have also evolved rapidly, expanding capabilities from high-fidelity text-to-image synthesis to controllable text-to-video generation\cite{zhang2023survey,chen2025comprehensive,lipman2022flow}.
However, generative models and MLLMs have progressed along separate tracks, making it challenging to unify understanding and generation in one system \cite{zhao2025unified}.
Unified multimodal models (UMMs) provide a consistent framework for multimodal understanding and generation, showing great potential for multi-step reasoning and controllable image editing \cite{yang2025survey,chen2025blip3,wu2025omnigen2,deng2025emergingpropertiesunifiedmultimodal}.

The advancement of multimodal capabilities relies on the availability of high-quality datasets \cite{yang2025survey}.
For UMMs, there remains a lack of a scalable and reproducible paradigm for producing such high-quality instruction data while minimizing manual annotation cost \cite{yin_survey_2024,deng_enhancing_2024}.
Raw videos provide a scalable source of multimodal knowledge, covering broad domains and temporally ordered interactions beyond static images~\cite{tong2025thinkingvideovideogeneration}; however, they are also redundant, noisy, and unstructured, and do not directly provide task-oriented instruction data.
The key challenge is therefore to convert large-scale raw videos into instruction data that jointly supports understanding and generation while remaining suitable for stable model training.

\begin{figure}[!htbp]
  \centering
  \includegraphics[width=1.0\linewidth]{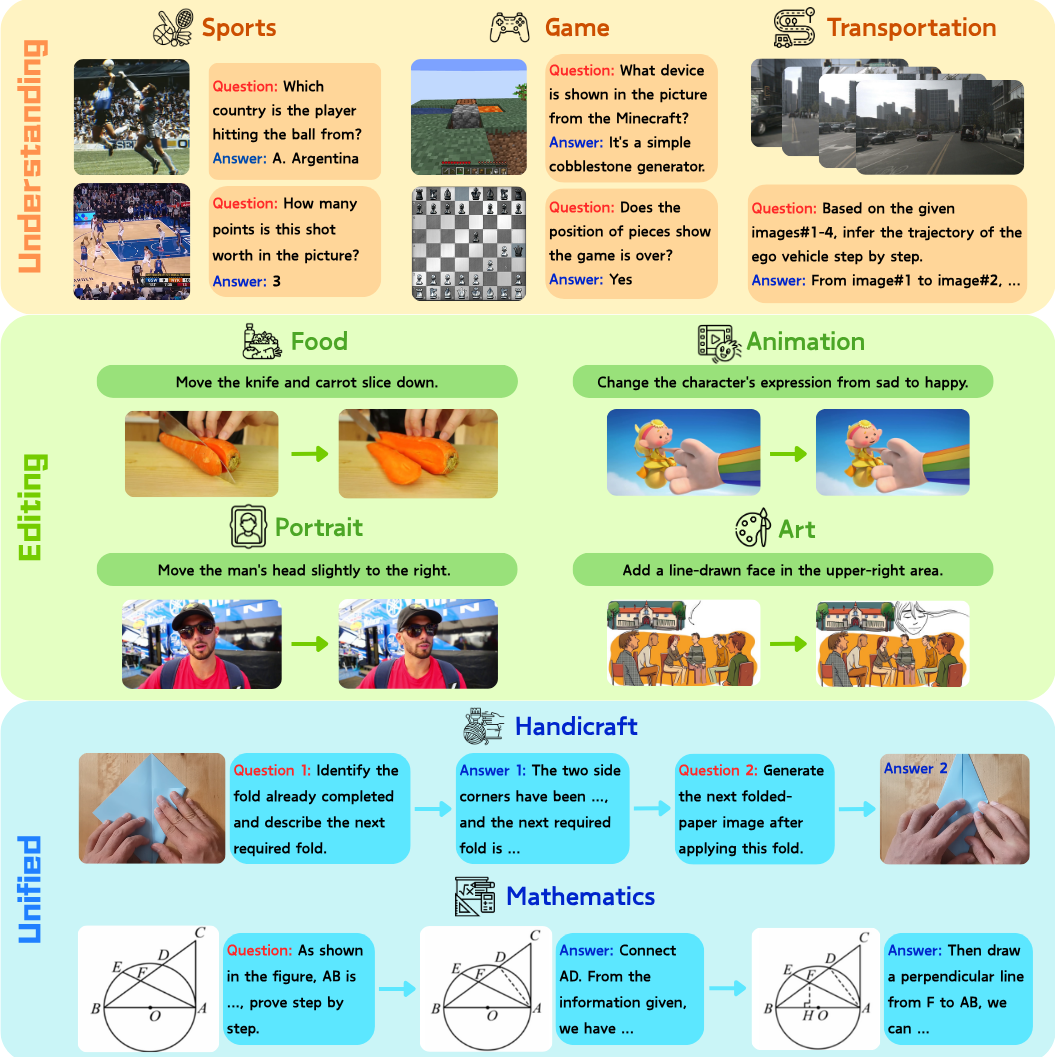}
  \caption{Example samples of VidUEU-100K. We illustrate three capability paradigms: Understanding (top), Editing (middle), and Unified (bottom).}
  \label{fig:agent_overview}
\end{figure}

In recent years, LLM-based agent systems have demonstrated strong planning and tool-use capabilities, enabling them to decompose complex goals into executable steps and to achieve large-scale data generation, verification, and filtering \cite{Guo2024LargeLM,xu2025llm,kumar2024mmctagent,zhang2025vc}.
By interacting with external tools and making feedback-driven decisions, agents can automate complex data-curation workflows
\cite{tang2025synthesizingposttrainingdatallms,riaz-etal-2025-metasynth}.
These properties make agent systems a fit for transforming raw videos into structured instruction data.

Motivated by the strengths of agent systems and video data, we introduce VidUEU-Agent, an automated data synthesis framework that leverages agent planning and tool use to convert raw videos into multimodal instruction data.
VidUEU-Agent follows a four-stage pipeline: it performs global perception to obtain coarse video understanding and filter low-quality shots, mines keyframes and multimodal signals through frame filtering and tool use, constructs metadata from the keyframe sequence and filtered signals, and routes the metadata to synthesize instruction samples for understanding, editing, and unified tasks.
VidUEU-Agent further autonomously iterates the data distribution based on downstream training results, improving data quality, balance, and training effectiveness.

Further, we scale this pipeline to construct VidUEU-100K, a large-scale high-quality multimodal instruction dataset built from 8,268 diverse videos collected from the web and open datasets.
The dataset contains nearly 20k dense metadata annotations, enabling 40K understanding samples, 30K editing samples, and 30K unified samples, and it spans nine task paradigms and five scenario categories, ensuring comprehensive and balanced data for improving the overall capabilities of UMMs.
We further fine-tune a series of baseline models on VidUEU-100K and validate the effectiveness of our data through multiple benchmarks and case studies.

In summary, our contributions are as follows:
\begin{itemize}
    \item We propose VidUEU-Agent, a scalable and automated framework that leverages the planning and tool-using capabilities of agents to synthesize high-quality instruction data for multimodal understanding, editing, and unified tasks from video sources.

    \item By scaling this pipeline, we construct VidUEU-100K, a large-scale dataset featuring 20k dense metadata annotations and 100K diverse instruction samples, covering diverse domains and complex tasks across understanding, editing, and unified capabilities.

    \item We fine-tune a series of baseline models on VidUEU-100K, where extensive experiments across multiple benchmarks demonstrate the effectiveness of our data in enhancing model performance in various multimodal tasks.
\end{itemize}

\section{Related Work}

\subsection{Data Construction based on Agent Systems}

Early methods often used powerful LLMs to directly generate instruction--response pair \cite{xu2023wizardlm,liu2023visual,zhao2024long}.
Despite their efficiency, this approach lacks grounding in verifiable external signals, introducing uncertainty in data reliability and controllability
\cite{ji2023survey,rawte2023survey,wang2023self}.
Recent trends shift toward vision agents and tool use \cite{gao2025multimodalagenttuningbuilding,deng2025toolscope}.
Such systems equip LLMs with perception tools for performing complex visual reasoning and information extraction to improve reliability and interpretability 
\cite{reddy2025orionunifiedvisualagent,deng2025toolscope}.
With this tool-use capability, agent-based data synthesis is becoming a promising approach for automated construction of multimodal datasets \cite{sun2025datasetagent}.
Prior work has explored using model agents to automatically collect bounding boxes, rewrite image captions, or synthesize question--answer pairs, greatly reducing reliance on manual annotation \cite{chen2023sharegpt4vimprovinglargemultimodal,joshi2025text}.
LLaVA-Video~\cite{zhang2025llavavideovideoinstructiontuning} further synthesizes video-understanding QA from sampled frames, but remains focused on video understanding tasks.
However, existing methods struggle to jointly synthesize understanding, editing, and unified instruction data for UMMs, which hinders the development of these models across broader tasks \cite{zhao2025unified}.

\subsection{Multimodal Understanding and Generation Datasets}

Large-scale, high-quality instruction datasets have become one of the key drivers of recent progress in multimodal models \cite{li2023m3itlargescaledatasetmultimodal,chen2023sharegpt4vimprovinglargemultimodal,liu2024improvedbaselinesvisualinstruction}.
In multimodal understanding, existing work enhances model capabilities in visual question answering and fine-grained recognition \cite{liu2023visual,han2024parameter,yue2024mmmu,chen2024m3cot,liu2024mmbench,ru2026generationknowsunlockingfinegrained}. 
In multimodal generation, existing work advances high-fidelity generation and controllable editing \cite{zhao2023unleashing,podell2023sdxl,cao2025image,yu2023capsfusion,li2024laion}.
Recently, the multimodal field has shifted toward unified multimodal models~\cite{yang2025survey, zhang2026pelicanunify10unifiedembodied}.
Accordingly, unified architectures demand stable generalization across understanding and generation capabilities \cite{tong2025metamorph,zhuang2025vargptunifiedunderstandinggeneration,meng2026imugbenchbenchmarkingunifiedmultimodal}.
However, current image-text datasets are primarily sourced from static images, inherently lacking real-world dynamics
\cite{chen2025comm,shen2026danqinguptodatelargescalechinese}.
In contrast, videos, featuring inherent temporal dynamics and causal interactions, serve as a more promising medium for high-quality instruction data~\cite{tong2025thinkingvideovideogeneration}.

\section{VidUEU: Agent \& Dataset}
\label{sec:vidueu_agent_dataset}
\begin{figure}[tb]
  \centering
  \includegraphics[width=1.0\linewidth]{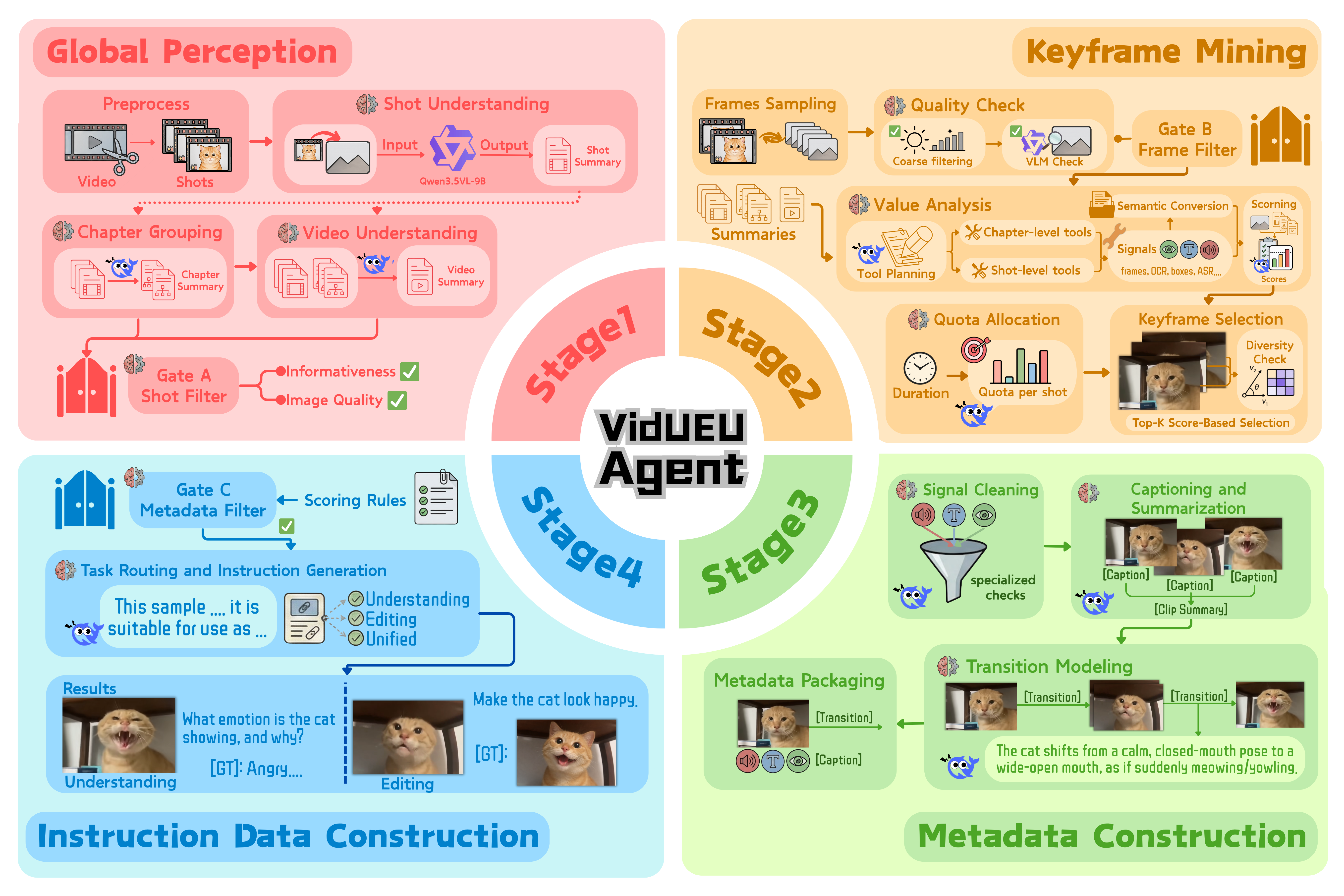}
  \caption{Overview of VidUEU-Agent. The agent performs \textbf{global perception} with quality gating, then conducts \textbf{keyframe mining} using multimodal signals and tool verification, builds \textbf{metadata}, and finally synthesizes \textbf{instruction data} for understanding, editing, and unified tasks.}
  \label{fig:pipeline}
\end{figure}

\subsection{VidUEU-Agent}
We propose \textbf{VidUEU-Agent}, a data synthesis agent for raw videos, designed to curate instruction data for multimodal understanding, editing, and unified tasks.
VidUEU-Agent first parses raw videos through global perception, removes low-quality shots, and then constructs metadata from selected keyframes and filtered multimodal signals.
It then maps each metadata entry to suitable instruction data formats, ensures data quality through multi-stage quality screening and consistency filtering, and autonomously iterates the data distribution using downstream training feedback.

\subsubsection{Stage 1: Global Perception}
As a preprocessing step, VidUEU-Agent first uses PySceneDetect \cite{castellano_pyscenedetect} to detect shot boundaries and split each long video into a sequence of shot-level clips.
It then performs global perception at three levels.
At the shot level, VidUEU-Agent samples the middle frame of each shot and prompts a VLM to generate a shot summary.
At the chapter level, the agent reviews all shot summaries of the video, groups adjacent shots according to their scene content, maps them into several chapters, and generates a chapter summary for each chapter.
At the video level, the agent summarizes the whole video based on all shot-level and chapter-level summaries to obtain a video summary.
Finally, VidUEU-Agent applies Gate A, namely the shot filter, to examine each shot according to informativeness and image quality rules, and removes low-quality shots from the subsequent stages.
The prompts for three-level summarization, as well as chapter construction, are provided in Appendix~\ref{app:stage1_global_perception}.


\subsubsection{Stage 2: Keyframe Mining}
We perform keyframe mining on the shots that pass Gate A.
\textbf{Frame Sampling.}
For each retained shot, VidUEU-Agent samples frames at a fixed interval and applies Gate B frame filtering, which combines tool-based quality statistics with VLM-based check to remove low-quality frames.
For each frame that passes Gate B, VidUEU-Agent generates a caption.
\textbf{Tool Planning.}
Based on the retained frames and the summaries from global perception, VidUEU-Agent performs hierarchical tool planning by assigning tools at both the chapter level and the shot level to collect targeted information.
Following this plan, VidUEU-Agent applies the selected tools to every retained frame in the corresponding shot.
\textbf{Frame Scoring and Keyframe Selection.}
VidUEU-Agent verbalizes the tool signals and scores each frame using these signals together with the three-level summaries. Based on the video duration, VidUEU-Agent sets the keyframe budget and allocates it across shots using the summaries. For each shot, VidUEU-Agent keeps the top-scoring candidates within the assigned quota, computes their image embeddings, measures cosine similarities, and applies weighted facility-location greedy selection to choose high-quality, semantically representative keyframes.
The prompts for tool planning, signal verbalization, frame scoring, and quota allocation are provided in Appendix~\ref{app:stage2_keyframe_mining}.
The tool pool, verbalization rules, and scoring rules are described in Appendix~\ref{app:tool_invocation_system}.


\label{sec:method}

\subsubsection{Stage 3: Metadata Construction}
To construct high-quality metadata and mitigate noise from unreliable signals, the agent first performs specialized checks on the raw multimodal signals bound to the keyframe sequence.
For each keyframe, the agent performs specialized checks on the associated signals using predefined rules and determines which signals to retain.
Then, based on the keyframe sequence and the filtered signals, it generates transitions describing changes between consecutive keyframes and a clip summary that describes the video clip based on the keyframe captions and verbalized tool signals, and verifies the quality and factuality of the generated text using the signals.
Finally, it binds the keyframes, their captions, the keyframe-bound multimodal signals, the transitions, and the clip summary to form a metadata list for the video clip.
The prompts for clip summary generation and pairwise transition generation are provided in Appendix~\ref{app:stage3_metadata_construction}, and the metadata schema is given in Appendix~\ref{app:metadata_schema_definition}.

\subsubsection{Stage 4: Instruction Data Construction}

\begin{figure}[!t]
\centering
\includegraphics[width=\textwidth]{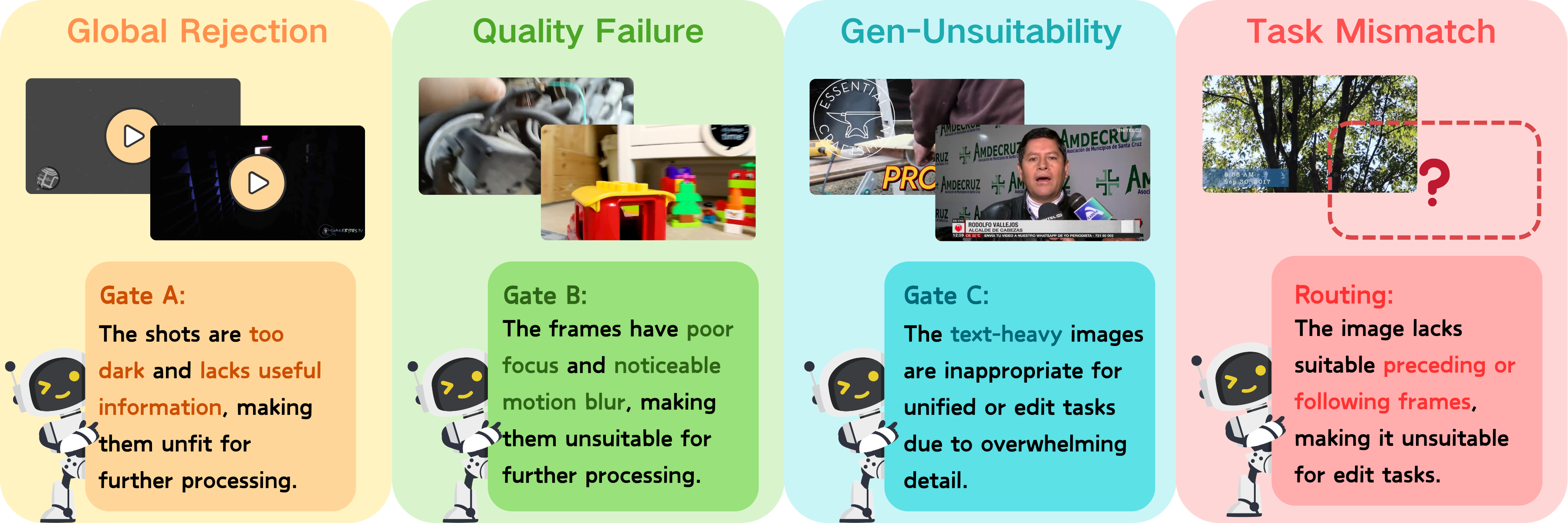}
\caption{Failure Case Examples: Illustrating four common rejected cases in the processing pipeline: Global Rejection, Quality Failure, Gen-Unsuitability, and Task Mismatch.}
\label{fig:failure_case}
\end{figure}

For the generated metadata, the agent applies Gate C to perform quality filtering and categorization.
Specifically, this gate calls a VLM to score metadata by overall information content, logical coherence, and overall quality, outputs a decision with reasons about its suitability as instruction data, and removes unsuitable metadata.
Subsequently, for each metadata entry, the agent determines the suitable task type(s) and outputs a task list with one or more of understanding, editing, and unified.
The prompts for task routing and instruction generation are provided in Appendix~\ref{app:stage4_instruction_data_construction}.
Finally, the agent matches instruction templates based on the metadata and its task list, and requires the model to cite the signal IDs used during instruction generation to improve traceability, thereby producing high-quality instruction data.
Figure~\ref{fig:failure_case} illustrates representative rejected cases across Gate A, Gate B, and Gate C, covering global rejection, low-quality visual inputs, generation-unsuitable frames, and task-routing mismatches.
Instruction-data synthesis templates and examples are described in Appendix~\ref{app:instruction_synthesis_templates}, and all quality-control filtering rules are provided in Appendix~\ref{app:quality_control_filtering_rules}.

\subsubsection{Autonomous Data Iteration}
To ensure the final 100K samples are high-quality, balanced, and effective for model training, VidUEU-Agent employs a self-iterative data refinement loop.
In each iteration, we train models on 30K samples, evaluate them on representative benchmarks, and return results to the agent.
The agent reflects on the data's weaknesses, compares new results with the baseline and previous iterations, and updates the target distribution over capabilities, domains, and task types.
Under the current policy, the agent adjusts video-type sampling in global perception to prioritize underrepresented types, shifting the domain distribution.
In keyframe mining, the agent modifies sampling density and keyframe budget, and adds policy-aware weights to keyframe scoring to align with the desired keyframe distribution.
During instruction data construction, task routing adjusts capability and task-type distributions, and quota-based sampling down-samples types whose ratios exceed the target ratios.
The agent audits the realized distribution and records statistics, feedback, and policy changes to guide the next iteration.
The final 100K dataset is scaled from the Iteration 4's target distribution.

\subsection{VidUEU-100K}
\label{sec:dataset}

Based on the above framework, with DeepSeek-V3 as the primary agent LLM~\cite{deepseekai2025deepseekv3technicalreport} and Qwen3.5-9B as the VLM~\cite{qwen3.5}, we construct the \textbf{VidUEU-100K} dataset.
It contains 100K high-quality instruction-tuning samples, including 40K understanding samples, 30K editing samples, and 30K unified samples, as well as 20k fine-grained metadata.

\subsubsection{Data Collection \& Sources}
\label{subsec:sources}

Large-scale video sources and a scalable agent pipeline enable VidUEU-100K to cover diverse scenarios, including rare and long-tail visual cases.
We source videos from large-scale video datasets, including OpenVid-1M \cite{nan2024openvid} and Panda-70M \cite{chen2024panda}, together with videos from \textit{Wikimedia} and YouTube.
To improve coverage in specific domains, we further use specialized open-source datasets such as FineGym \cite{shao2020finegym}, BDD100K \cite{yu2020bdd100k}, and CodeVidCL \cite{raja2025dataset}.
To build VidUEU-100K, we use 8,268 raw source videos, from which VidUEU-Agent derives approximately 8,505 semantic segments for dataset instance generation.

\subsubsection{Dataset Composition \& Statistics}
\label{subsec:stats}
To ensure comprehensive and balanced data, we analyze VidUEU-100K from three complementary dimensions: capability, domain, and task type.
For capability, we divide the 100K instruction-tuning samples into \textbf{Understanding}, \textbf{Editing}, and \textbf{Unified}: Understanding covers fine-grained perception, visual reasoning, and multi-image logical consistency; Editing focuses on instruction-based image editing while preserving irrelevant regions; and Unified involves interleaved text--image tasks that require models to understand visual context, infer state transitions, generate target visual states, and select images consistent with the intended change.
For domain, we categorize samples into Technology \& Tutorial, Sports \& Action, Daily Life \& Human Activities, Nature \& Scenery, and Gaming \& ACG.
For task type, the agent generates nine categories: Fine-grained Perception, Visual Reasoning, Multi-Image Logic, Editing, Grounded Edit-to-Select Generation, Interleaved Understand--Recall--Generate--Select, Stepwise Sequence Generate-and-Select, Middle-Step Gap Generate-and-Fill, and Instructional Sequential Generate-and-Select.
Detailed definitions and distributions of all capability, domain, and task categories are provided in Appendix~\ref{app:dataset_taxonomy}.

\section{Experiment}
\label{sec:experiment}

\subsection{Experimental Setup}
We LoRA-fine-tune InternVL3-8B~\cite{zhu2025internvl3exploringadvancedtraining} on the understanding subset and Bagel-7B-MoT~\cite{deng2025emergingpropertiesunifiedmultimodal} separately on the three capability subsets of VidUEU-100K.
We then evaluate the corresponding capabilities on MMT-Bench~\cite{ying2024mmtbenchcomprehensivemultimodalbenchmark}, MME~\cite{fu2025mme}, LLaVA-Bench~\cite{liu2024improvedbaselinesvisualinstruction}, MMDU~\cite{NEURIPS2024_10570531}, GEdit-Bench-En~\cite{liu2025step1xeditpracticalframeworkgeneral}, and WEAVEbench~\cite{chow_weave_2025}.
Detailed settings are provided in Appendix~\ref{app:f1_training_configuration}.

\newpage

\begin{table}
  \caption{Comparison of fine-tuned InternVL3-8B \& Bagel, and baseline models on benchmarks.}
  \label{tab:sum_overall_mmtacc}
  \centering
  \begin{tabular}{clccccc}
    \toprule
    Type & Model & Size & MMT-Bench$\uparrow$ & MME$\uparrow$ & LLaVA-Bench$\uparrow$ & MMDU$\uparrow$ \\
    \midrule
    \multirow{7}{*}{\rotatebox[origin=c]{90}{\itshape Und. Only}}
    & Qwen2.5-VL & 3B  & 0.577 & 2133.6 & 72.3 & 42.5 \\
    & Qwen2.5-VL & 7B  & 0.607 & 2308.6 & \underline{\textbf{82.7}} & 59.5 \\
    & LLaMA-3.2 & 11B & 0.456 & 1761.7 & 73.6 & 37.0 \\
    & InternVL3 & 1B  & 0.512 & 1871.9 & 46.4 & 39.0 \\
    & InternVL3 & 2B  & 0.532 & 2157.9 & 62.1 & 45.6 \\
    \cmidrule(lr){2-7}
    & InternVL3 & 8B  & 0.546 & 2308.0 & 69.0 & 53.1 \\
    & \cellcolor{green!12}Ours(InternVL3) & \cellcolor{green!12}8B & \cellcolor{green!12}0.596 & \cellcolor{green!12}\underline{\textbf{2369.3}} & \cellcolor{green!12}70.8 & \cellcolor{green!12}\underline{\textbf{61.1}} \\
    \midrule
    \multirow{4}{*}{\rotatebox[origin=c]{90}{\itshape Unified}}
    & Janus-Pro & 7B & 0.480 & 1784.4 & 58.7 & 42.6 \\
    & TokenFlow-XL & 14B & 0.532 & 1937.0 & 55.2 & 25.7 \\
    \cmidrule(lr){2-7}
    & Bagel & 14B & 0.645 & 1573.6 & 47.5 & 28.5 \\
    & \cellcolor{green!12}Ours(Bagel) & \cellcolor{green!12}14B & \cellcolor{green!12}\underline{\textbf{0.651}} & \cellcolor{green!12}1900.2 & \cellcolor{green!12}57.9 & \cellcolor{green!12}33.2 \\
    \bottomrule
  \end{tabular}
\end{table}

\begin{table}[!t]
\centering
\begin{minipage}[!t]{0.49\textwidth}
  \vspace{-1.4em}
  \captionsetup{width=\linewidth,hypcap=false,justification=centering,singlelinecheck=true,skip=2pt}
  \captionof{table}{Comparison of fine-tuned Bagel and baseline models on GEdit-Bench-En.}
  \label{tab:gedit-bench-en}
  {\small
  \renewcommand{\arraystretch}{0.92}
  \setlength{\tabcolsep}{3.4pt}
  \centering
  \begin{tabular}{clccc}
    \toprule
    Type & Model & G\_SC$\uparrow$ & G\_PQ$\uparrow$ & G\_O$\uparrow$ \\
    \midrule
    \multirow{4}{0.75cm}{\centering\rotatebox[origin=c]{90}{\itshape Edit Only}}
    & Instruct-Pix2Pix & 3.575 & 5.491 & 3.684 \\
    & MagicBrush & 4.677 & 5.656 & 4.518 \\
    & AnyEdit & 3.178 & 5.820 & 3.212 \\
    & Step1X-Edit & 7.091 & 6.763 & 6.701 \\
    \midrule
    \multirow{4}{0.75cm}{\centering\raisebox{-1.15ex}{\rotatebox[origin=c]{90}{\itshape Unified}}}
    & Gemini-2.0-flash & 6.732 & 6.606 & 6.315 \\
    & OmniGen & 5.963 & 5.888 & 5.061 \\
    \cmidrule(lr){2-5}
    & Bagel & 7.358 & 6.831 & 6.527 \\
    & \cellcolor{green!12}Ours(Bagel) & \cellcolor{green!12}\underline{\textbf{7.565}} & \cellcolor{green!12}\underline{\textbf{7.328}} & \cellcolor{green!12}\underline{\textbf{6.951}} \\
    \bottomrule
  \end{tabular}
  }
\end{minipage}\hfill
\begin{minipage}[!htbp]{0.47\textwidth}
  \vspace{0em}
  
\subsection{Results on Understanding Capabilities}
We fine-tune InternVL3-8B and Bagel-7B-MoT on the understanding subset to evaluate the effect of VidUEU-100K on multimodal understanding.
As shown in Table~\ref{tab:sum_overall_mmtacc}, the data produced by VidUEU-Agent substantially improves the understanding ability of both the understanding-only model and the unified model.
For InternVL3-8B, the gains are most notable on MME (2369.3) and MMDU (61.1), achieving the best results among all compared models on both benchmarks. 
For Bagel, fine-

\end{minipage}

\par\vspace{-2.1em}
\end{table}

tuning improves all reported understanding scores and achieves the best MMT-Bench score (0.651).

Overall, these results show that the supervision generated by VidUEU-Agent effectively improves multimodal understanding and generalizes across different model families.
Particularly, since MMDU evaluates multi-image reasoning and answer consistency grounded in visual evidence, the MMDU improvements of InternVL3-8B and Bagel show that VidUEU-Agent can mine semantically related keyframes from videos and model their relations in instruction data.

\subsection{Results on Editing Capabilities}

We fine-tune Bagel on the editing subset to evaluate the effect of VidUEU-100K on instruction-based image editing.
As shown in Table~\ref{tab:gedit-bench-en}, the data produced by VidUEU-Agent improves Bagel on all GEdit-Bench metrics, reaching 7.565 on G\_SC, 7.328 on G\_PQ, and 6.951 on G\_O, achieving the best results among all com
-pared models.
These results show that VidUEU-100K improves both edit correctness and output quality, rather than only optimizing one aspect of image editing.

The two core metrics reveal a consistent but uneven gain pattern.
The improvement on G\_SC shows that VidUEU-Agent can extract meaningful visual changes from videos and convert them into instructions that guide the target edit more effectively.
The larger gain on G\_PQ further indicates that video state-transition supervision helps preserve image realism and local visual structure during editing.
This is because adjacent keyframes usually share most scene content but differ in specific objects, attributes, or states, matching the edit-and-preserve nature of instruction-based image editing.

\subsection{Results on Unified Capabilities}
We fine-tune Bagel on the unified subset to evaluate the effect of VidUEU-100K on interleaved understanding, generation, visual memory, and cross-turn reasoning.
As shown in Table~\ref{tab:WEAVEbench_comparison}, the data produced by VidUEU-Agent improves Bagel's Overall score on WEAVEbench from 0.446 to 0.501.
The fine-tuned model outperforms the Bagel and Bagel-Zebra baselines, reaching 0.480 on Logic and 0.419 on Game.

\begin{table}[!t]
  \caption{Comparison of fine-tuned Bagel and baseline models on WEAVEbench.}
  \label{tab:WEAVEbench_comparison}
  \centering
  \begin{tabular}{c l c c c c c c}
    \toprule
    Type & Model & Size & Science$\uparrow$ & Creation$\uparrow$ & Logic$\uparrow$ & Game$\uparrow$ & Overall$\uparrow$ \\
    \midrule
    \multirow{5}{1.2cm}{\centering\rotatebox[origin=c]{90}{\itshape Unified}}
    & OmniGen2    & 7B  & \underline{\textbf{0.511}} & \underline{\textbf{0.682}} & \underline{\textbf{0.551}}  & \underline{\textbf{0.511}} & \underline{\textbf{0.609}} \\
    & Ovis-U1     & 3B  & 0.402 & 0.557 & 0.364 & 0.357 & 0.422 \\
    & Bagel       & 14B & 0.378 & 0.475 & 0.406 & 0.365 & 0.446 \\
    & Bagel-Zebra & 14B & 0.399 & 0.456 & 0.393 & 0.396 & 0.449 \\
    \cmidrule(lr){2-8}
    & \cellcolor{green!12}Ours(Bagel) & \cellcolor{green!12}14B & \cellcolor{green!12}0.426 & \cellcolor{green!12}0.501 & \cellcolor{green!12}0.480 & \cellcolor{green!12}0.419 & \cellcolor{green!12}0.501 \\
    \bottomrule
  \end{tabular}
\end{table}

\begin{table}[!t]
\centering
\begin{minipage}[!t]{0.49\textwidth}
  \vspace{-1.4em}

  \captionsetup{width=\linewidth,hypcap=false,justification=centering,singlelinecheck=true,skip=2pt}
  \captionof{table}{Results of autonomous data iteration.}
  \label{tab:autonomous_iteration}
  {\small
  \renewcommand{\arraystretch}{0.92}
  \setlength{\tabcolsep}{4.0pt}
  \makebox[\linewidth][c]{%
  \begin{tabular}{lccc}
    \toprule
    Setting & MMDU$\uparrow$ & GEdit $\uparrow$ & WEAVE $\uparrow$ \\
    \midrule
    Iteration 1 & 29.7 & 6.541 & 0.448 \\
    Iteration 2 & 29.2 & 6.796 & 0.452 \\
    Iteration 3 & 31.4 & 6.873 & 0.458 \\
    Iteration 4 & 32.8 & 6.901 & 0.499 \\
    \midrule
    \cellcolor{green!12}Final & \cellcolor{green!12}\underline{\textbf{33.0}} & \cellcolor{green!12}\underline{\textbf{7.065}} & \cellcolor{green!12}\underline{\textbf{0.526}} \\
    \bottomrule
  \end{tabular}%
  }
  }

  \vspace{1.5em}

  \captionsetup{width=\linewidth,hypcap=false,justification=centering,singlelinecheck=true,skip=2pt}
  \captionof{table}{Ablation study of VidUEU-Agent.}
  \label{tab:ablation_vidueu_agent}
  {\small
  \renewcommand{\arraystretch}{0.92}
  \setlength{\tabcolsep}{4.0pt}
  \makebox[\linewidth][c]{%
  \begin{tabular}{lccc}
    \toprule
    Setting & MMDU$\uparrow$ & GEdit $\uparrow$ & WEAVE $\uparrow$ \\
    \midrule
    \cellcolor{green!12}Ours & \cellcolor{green!12}\underline{\textbf{33.2}} & \cellcolor{green!12}\underline{\textbf{6.951}} & \cellcolor{green!12}\underline{\textbf{0.501}} \\
    \midrule
    w/o A & 32.4 & 6.773 & 0.497 \\
    w/o B & 24.7 & 5.175 & 0.404 \\
    w/o S & 30.8 & 5.848 & 0.462 \\
    w/o C & 30.5 & 6.303 & 0.454 \\
    N & 22.0 & 5.041 & 0.393 \\
    w/o R & 29.4 & 6.031 & 0.470 \\
    \bottomrule
  \end{tabular}%
  }
  }
\end{minipage}\hfill
\begin{minipage}[!htbp]{0.47\textwidth}
  \vspace{0pt}

These results highlight temporal supervision in unified multimodal tasks.
The gains on Logic and Game suggest that the model tracks states, follows rules, models object relations, and reasons over events across turns.
This matches the structure of adjacent keyframes, which preserve scene continuity while showing changes in objects, attributes, actions, and outcomes.
The modest Creation gain suggests that such supervision is more effective for structured transitions than open-ended creative generation.

\subsection{Case Study}

For understanding,  Bagel-7B-MoT shows hallucination and cross-image misbinding, while the fine-tuned model relies more on the visible evidence.
For editing, Bagel often fails to follow background replacement instructions, whereas the fine-tuned model produces more realistic target edits while preserving irrelevant regions. 
\end{minipage}
\par\vspace{-2.2em}
\end{table}

For unified tasks, the fine-tuned model better executes edits and context use in early turns, but long interleaved image--text chains remain challenging because early errors may propagate to later turns.

Overall, VidUEU-100K improves instruction alignment, compositional binding, and local-edit preservation, while long-horizon interleaved interaction remains an open challenge.

\subsection{Results on Autonomous Data Iteration}

We further evaluate the autonomous data iteration process of VidUEU-Agent.
Iterations 1--4 each construct 30K samples and jointly fine-tune BAGEL from the same initial checkpoint before evaluation.
The benchmark results are then fed back into the agent, which updates the target distributions 
across capabilities, domains, and task types for the next iteration.
The Final 100K dataset is constructed by scaling the target distribution obtained in Iteration 4, and a single BAGEL model is jointly fine-tuned on the complete dataset.
Table~\ref{tab:autonomous_iteration} reports results on MMDU, GEdit-Bench-En, and WEAVEbench, whose aggregate test-split scores guide iteration and constitute development-guided evaluations; full results and feedback-free evaluations on MMBench, MMStar~\cite{chen2024rightwayevaluatinglarge}, and VSI-Bench~\cite{yang2025thinkingspacemultimodallarge} appear in Appendix~\ref{app:iteration_ablation_full}.

\begin{figure}
    \centering
    \includegraphics[width=1.0\linewidth]{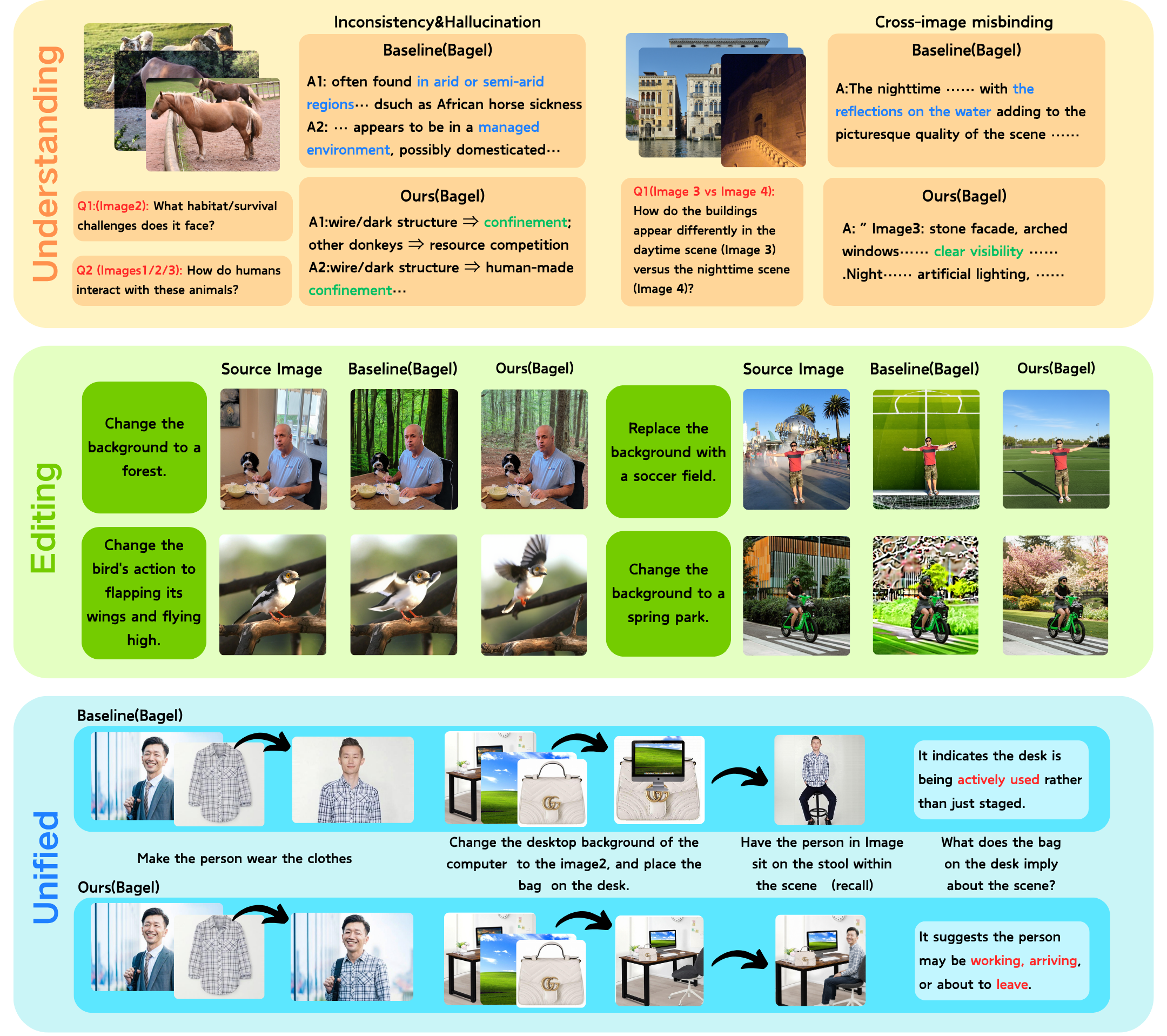}
    \caption{Case study examples comparing baseline models and VidUEU-100K fine-tuned models.}
    \label{fig:case_study}
    \vspace{-1em}
\end{figure}

The iteration results show a clear upward trend after repeated feedback-driven refinement.
Although Iteration 2 slightly decreases on MMDU, it improves on GEdit and WEAVE, suggesting that early distribution updates first benefit generation-related capabilities.
From Iteration 3 to Iteration 4, the gains become more consistent across all three benchmarks, indicating that the agent gradually improves the balance among capabilities, domains, and task types.
Final further improves upon Iteration 4 with 100K samples.

\subsection{Ablation Study}

We conduct ablations to evaluate key design choices in VidUEU-Agent.
Ablations train Bagel on 10K samples per capability; Ours uses separately trained Bagel models on the Final 100K subsets.
We separately remove Gate A shot filtering (A), Gate B frame quality filtering (B), score-and-similarity-based greedy keyframe sampling (S), and Gate C metadata filtering (C).
We compare with a no-agent variant (N) that uses fixed-interval sampled frames as metadata and directly uses a VLM to generate QA pairs without tool evidence.
Finally, we remove the consistency hard rule (R) requiring each instruction sample to cite used signal IDs.
Table~\ref{tab:ablation_vidueu_agent} reports the main results on MMDU, GEdit-Bench-En G\_O score, and WEAVEbench overall score; the full results with the other three understanding benchmarks are provided in Appendix~\ref{app:ablation_vidueu_agent_full}.
Baseline dataset comparisons appear in Appendix~\ref{app:baseline_dataset_comparison}.

The ablation results suggest that all components contribute to performance.
Removing Gate A causes only a mild drop, suggesting that the collected videos contain relatively few meaningless or extremely dark shots after source-level curation.
Removing Gate B clearly hurts all benchmarks, especially editing and unified tasks, showing that the agent must filter the many low-quality frames in raw videos.
Removing score-similarity greedy sampling hurts editing more, indicating that editing data benefits from keyframes that are both informative and non-redundant, rather than from dense but weakly differentiated frame pairs.
Removing Gate C degrades all tasks, showing that the final metadata filtering step is necessary to prevent inconsistent, incomplete, or task-mismatched evidence from entering instruction synthesis.
The no-agent variant drops the most, showing that direct VLM-based QA generation lacks the structured quality control, tool evidence, and task routing needed to construct reliable multimodal supervision.
Removing the consistency hard rule also hurts performance, indicating that evidence citation suppresses weakly grounded or low-quality samples.

\section{Conclusion}
\label{sec:conclusion}
In this paper, we study the bottleneck of high-quality instruction data for multimodal tasks, and propose a scalable agent-based solution that leverages raw videos as a rich and dynamic data source.
Specifically, VidUEU-Agent constructs instruction data through four stages: global perception, keyframe mining, metadata construction, and instruction data construction, and refines the data distribution through autonomous data iteration using downstream feedback.
Using VidUEU-Agent, we construct VidUEU-100K, a large-scale dataset with 100K instruction samples spanning understanding, editing, and unified capabilities.
Extensive experiments across representative benchmarks show that fine-tuning baseline models with data produced by VidUEU-Agent brings consistent performance improvements, validating the effectiveness and generalizability of the proposed agent pipeline.
We hope VidUEU-Agent and VidUEU-100K provide a reproducible and extensible foundation for scalable multimodal data construction, and facilitate the development of unified multimodal models that more reliably integrate understanding and generation.



\bibliographystyle{plainnat}
\bibliography{main}

@InProceedings{10.1007/978-3-030-58536-5_24,
author="Teed, Zachary
and Deng, Jia",
editor="Vedaldi, Andrea
and Bischof, Horst
and Brox, Thomas
and Frahm, Jan-Michael",
title="RAFT: Recurrent All-Pairs Field Transforms for Optical Flow",
booktitle="Computer Vision -- ECCV 2020",
year="2020",
publisher="Springer International Publishing",
address="Cham",
pages="402--419",
isbn="978-3-030-58536-5"
}

@software{jocher_2024_10983461,
  author       = {Jocher, Glenn and
                  Chaurasia, Ayush and
                  Qiu, Jing},
  title        = {Ultralytics YOLO},
  month        = apr,
  year         = 2024,
  publisher    = {Zenodo},
  version      = {v8.2.0},
  doi          = {10.5281/zenodo.10983461},
  url          = {https://doi.org/10.5281/zenodo.10983461},
}

@misc{cui2025paddleocr30technicalreport,
      title={PaddleOCR 3.0 Technical Report}, 
      author={Cheng Cui and Ting Sun and Manhui Lin and Tingquan Gao and Yubo Zhang and Jiaxuan Liu and Xueqing Wang and Zelun Zhang and Changda Zhou and Hongen Liu and Yue Zhang and Wenyu Lv and Kui Huang and Yichao Zhang and Jing Zhang and Jun Zhang and Yi Liu and Dianhai Yu and Yanjun Ma},
      year={2025},
      eprint={2507.05595},
      archivePrefix={arXiv},
      primaryClass={cs.CV},
      url={https://arxiv.org/abs/2507.05595}, 
}

@inproceedings{
fu2025mme,
title={{MME}: A Comprehensive Evaluation Benchmark for Multimodal Large Language Models},
author={Chaoyou Fu and Peixian Chen and Yunhang Shen and Yulei Qin and Mengdan Zhang and Xu Lin and Jinrui Yang and Xiawu Zheng and Ke Li and Xing Sun and Yunsheng Wu and Rongrong Ji and Caifeng Shan and Ran He},
booktitle={The Thirty-ninth Annual Conference on Neural Information Processing Systems Datasets and Benchmarks Track},
year={2025},
url={https://openreview.net/forum?id=DgH9YCsqWm}
}

@misc{ying2024mmtbenchcomprehensivemultimodalbenchmark,
      title={MMT-Bench: A Comprehensive Multimodal Benchmark for Evaluating Large Vision-Language Models Towards Multitask AGI}, 
      author={Kaining Ying and Fanqing Meng and Jin Wang and Zhiqian Li and Han Lin and Yue Yang and Hao Zhang and Wenbo Zhang and Yuqi Lin and Shuo Liu and Jiayi Lei and Quanfeng Lu and Runjian Chen and Peng Xu and Renrui Zhang and Haozhe Zhang and Peng Gao and Yali Wang and Yu Qiao and Ping Luo and Kaipeng Zhang and Wenqi Shao},
      year={2024},
      eprint={2404.16006},
      archivePrefix={arXiv},
      primaryClass={cs.CV},
      url={https://arxiv.org/abs/2404.16006}, 
}

@inproceedings{wu2023multimodal,
  title={Multimodal large language models: A survey},
  author={Wu, Jiayang and Gan, Wensheng and Chen, Zefeng and Wan, Shicheng and Yu, Philip S},
  booktitle={2023 IEEE International Conference on Big Data (BigData)},
  pages={2247--2256},
  year={2023},
  organization={IEEE}
}

@article{liang2024comprehensive,
  title={A comprehensive survey and guide to multimodal large language models in vision-language tasks},
  author={Liang, Chia Xin and Tian, Pu and Yin, Caitlyn Heqi and Yua, Yao and An-Hou, Wei and Ming, Li and Song, Xinyuan and Wang, Tianyang and Bi, Ziqian and Liu, Ming},
  journal={arXiv preprint arXiv:2411.06284},
  year={2024}
}

@article{chen2025comprehensive,
  title={Comprehensive exploration of diffusion models in image generation: a survey},
  author={Chen, Hang and Xiang, Qian and Hu, Jiaxin and Ye, Meilin and Yu, Chao and Cheng, Hao and Zhang, Lei},
  journal={Artificial Intelligence Review},
  volume={58},
  number={4},
  pages={99},
  year={2025},
  publisher={Springer}
}

@article{zhang2023survey,
  title={A survey of diffusion based image generation models: Issues and their solutions},
  author={Zhang, Tianyi and Wang, Zheng and Huang, Jing and Tasnim, Mohiuddin Muhammad and Shi, Wei},
  journal={arXiv preprint arXiv:2308.13142},
  year={2023}
}

@article{lipman2022flow,
  title={Flow matching for generative modeling},
  author={Lipman, Yaron and Chen, Ricky TQ and Ben-Hamu, Heli and Nickel, Maximilian and Le, Matt},
  journal={arXiv preprint arXiv:2210.02747},
  year={2022}
}

@article{zhao2025unified,
  title={Unified multimodal understanding and generation models: Advances, challenges, and opportunities},
  author={Zhao, Shanshan and Zhang, Xinjie and Guo, Jintao and Hu, Jiakui and Duan, Lunhao and Fu, Minghao and Chng, Yong Xien and Wang, Guo-Hua and Chen, Qing-Guo and Xu, Zhao and others},
  journal={arXiv preprint arXiv:2505.02567},
  year={2025}
}

@article{yang2025survey,
  title={A Survey of Unified Multimodal Understanding and Generation: Advances and Challenges},
  author={Yang, Yan and Tian, Haochen and Shi, Yang and Xie, Wulin and Zhang, Yi-Fan and Dong, Yuhao and Hu, Yibo and Wang, Liang and He, Ran and Shan, Caifeng and others},
  journal={Authorea Preprints},
  year={2025},
  publisher={Authorea}
}

@article{chen2025blip3,
  title={Blip3-o: A family of fully open unified multimodal models-architecture, training and dataset},
  author={Chen, Jiuhai and Xu, Zhiyang and Pan, Xichen and Hu, Yushi and Qin, Can and Goldstein, Tom and Huang, Lifu and Zhou, Tianyi and Xie, Saining and Savarese, Silvio and others},
  journal={arXiv preprint arXiv:2505.09568},
  year={2025}
}

@article{wu2025omnigen2,
  title={Omnigen2: Exploration to advanced multimodal generation},
  author={Wu, Chenyuan and Zheng, Pengfei and Yan, Ruiran and Xiao, Shitao and Luo, Xin and Wang, Yueze and Li, Wanli and Jiang, Xiyan and Liu, Yexin and Zhou, Junjie and others},
  journal={arXiv preprint arXiv:2506.18871},
  year={2025}
}

@misc{deng2025emergingpropertiesunifiedmultimodal,
      title={Emerging Properties in Unified Multimodal Pretraining}, 
      author={Chaorui Deng and Deyao Zhu and Kunchang Li and Chenhui Gou and Feng Li and Zeyu Wang and Shu Zhong and Weihao Yu and Xiaonan Nie and Ziang Song and Guang Shi and Haoqi Fan},
      year={2025},
      eprint={2505.14683},
      archivePrefix={arXiv},
      primaryClass={cs.CV},
      url={https://arxiv.org/abs/2505.14683}, 
}

@inproceedings{liu2024mmbench,
  title={Mmbench: Is your multi-modal model an all-around player?},
  author={Liu, Yuan and Duan, Haodong and Zhang, Yuanhan and Li, Bo and Zhang, Songyang and Zhao, Wangbo and Yuan, Yike and Wang, Jiaqi and He, Conghui and Liu, Ziwei and others},
  booktitle={European conference on computer vision},
  pages={216--233},
  year={2024},
  organization={Springer}
}

@inproceedings{yue2024mmmu,
  title={Mmmu: A massive multi-discipline multimodal understanding and reasoning benchmark for expert agi},
  author={Yue, Xiang and Ni, Yuansheng and Zhang, Kai and Zheng, Tianyu and Liu, Ruoqi and Zhang, Ge and Stevens, Samuel and Jiang, Dongfu and Ren, Weiming and Sun, Yuxuan and others},
  booktitle={Proceedings of the IEEE/CVF conference on computer vision and pattern recognition},
  pages={9556--9567},
  year={2024}
}

@misc{chow_weave_2025,
	title = {{WEAVE}: {Unleashing} and {Benchmarking} the {In}-context {Interleaved} {Comprehension} and {Generation}},
	shorttitle = {{WEAVE}},
	url = {http://arxiv.org/abs/2511.11434},
	doi = {10.48550/arXiv.2511.11434},
	urldate = {2026-02-07},
	publisher = {arXiv},
	author = {Chow, Wei and Pan, Jiachun and Liang, Yongyuan and Zhou, Mingze and Song, Xue and Jia, Liyu and Zhang, Saining and Tang, Siliang and Li, Juncheng and Zhang, Fengda and Wu, Weijia and Zhang, Hanwang and Chua, Tat-Seng},
	month = nov,
	year = {2025},
	note = {arXiv:2511.11434 [cs]},
}

@inproceedings{chen2024m3cot,
  title={M3cot: A novel benchmark for multi-domain multi-step multi-modal chain-of-thought},
  author={Chen, Qiguang and Qin, Libo and Zhang, Jin and Chen, Zhi and Xu, Xiao and Che, Wanxiang},
  booktitle={Proceedings of the 62nd Annual Meeting of the Association for Computational Linguistics (Volume 1: Long Papers)},
  pages={8199--8221},
  year={2024}
}

@article{li2024laion,
  title={LAION-SG: An Enhanced Large-Scale Dataset for Training Complex Image-Text Models with Structural Annotations},
  author={Li, Zejian and Meng, Chenye and Li, Yize and Yang, Ling and Zhang, Shengyuan and Ma, Jiarui and Li, Jiayi and Yang, Guang and Yang, Changyuan and Yang, Zhiyuan and others},
  journal={arXiv preprint arXiv:2412.08580},
  year={2024}
}

@article{yu2023capsfusion,
  title={CapsFusion: Rethinking Image-Text Data at Scale},
  author={Yu, Qiying and Sun, Quan and Zhang, Xiaosong and Cui, Yufeng and Zhang, Fan and Cao, Yue and Wang, Xinlong and Liu, Jingjing},
  journal={arXiv preprint arXiv:2310.20550},
  year={2023}
}

@inproceedings{deng_enhancing_2024,
	title = {Enhancing {Large} {Vision} {Language} {Models} with {Self}-{Training} on {Image} {Comprehension}},
	volume = {37},
	url = {https://proceedings.neurips.cc/paper_files/paper/2024/file/ed45d6a03de84cc650cae0655f699356-Paper-Conference.pdf},
	doi = {10.52202/079017-4175},
	booktitle = {Advances in {Neural} {Information} {Processing} {Systems}},
	publisher = {Curran Associates, Inc.},
	author = {Deng, Yihe and Lu, Pan and Yin, Fan and Hu, Ziniu and Shen, Sheng and Gu, Quanquan and Zou, James and Chang, Kai-Wei and Wang, Wei},
	editor = {Globerson, A. and Mackey, L. and Belgrave, D. and Fan, A. and Paquet, U. and Tomczak, J. and Zhang, C.},
	year = {2024},
	pages = {131369--131397},
}

@article{yin_survey_2024,
	title = {A survey on multimodal large language models},
	volume = {11},
	issn = {2095-5138},
	url = {https://doi.org/10.1093/nsr/nwae403},
	doi = {10.1093/nsr/nwae403},
	number = {12},
	journal = {National Science Review},
	author = {Yin, Shukang and Fu, Chaoyou and Zhao, Sirui and Li, Ke and Sun, Xing and Xu, Tong and Chen, Enhong},
	month = nov,
	year = {2024},
	note = {\_eprint: https://academic.oup.com/nsr/article-pdf/11/12/nwae403/61201557/nwae403.pdf},
	pages = {nwae403},
}

@inproceedings{Guo2024LargeLM,
  title={Large Language Model based Multi-Agents: A Survey of Progress and Challenges},
  author={Taicheng Guo and Xiuying Chen and Yaqi Wang and Ruidi Chang and Shichao Pei and N. Chawla and Olaf Wiest and Xiangliang Zhang},
  booktitle={International Joint Conference on Artificial Intelligence},
  year={2024},
  url={https://api.semanticscholar.org/CorpusID:267412980}
}

@article{xu2025llm,
  title={LLM-Based Agents for Tool Learning: A Survey: W. Xu et al.},
  author={Xu, Weikai and Huang, Chengrui and Gao, Shen and Shang, Shuo},
  journal={Data Science and Engineering},
  pages={1--31},
  year={2025},
  publisher={Springer}
}

@misc{tang2025synthesizingposttrainingdatallms,
      title={Synthesizing Post-Training Data for LLMs through Multi-Agent Simulation}, 
      author={Shuo Tang and Xianghe Pang and Zexi Liu and Bohan Tang and Rui Ye and Tian Jin and Xiaowen Dong and Yanfeng Wang and Siheng Chen},
      year={2025},
      eprint={2410.14251},
      archivePrefix={arXiv},
      primaryClass={cs.AI},
      url={https://arxiv.org/abs/2410.14251}, 
}

@inproceedings{riaz-etal-2025-metasynth,
    title = "{M}eta{S}ynth: Meta-Prompting-Driven Agentic Scaffolds for Diverse Synthetic Data Generation",
    author = "Riaz, Haris  and
      Bhabesh, Sourav Sanjukta  and
      Arannil, Vinayak  and
      Ballesteros, Miguel  and
      Horwood, Graham",
    editor = "Che, Wanxiang  and
      Nabende, Joyce  and
      Shutova, Ekaterina  and
      Pilehvar, Mohammad Taher",
    booktitle = "Findings of the Association for Computational Linguistics: ACL 2025",
    month = jul,
    year = "2025",
    address = "Vienna, Austria",
    publisher = "Association for Computational Linguistics",
    url = "https://aclanthology.org/2025.findings-acl.962/",
    doi = "10.18653/v1/2025.findings-acl.962",
    pages = "18770--18803",
    ISBN = "979-8-89176-256-5"
}

@misc{li2023m3itlargescaledatasetmultimodal,
      title={M$^3$IT: A Large-Scale Dataset towards Multi-Modal Multilingual Instruction Tuning}, 
      author={Lei Li and Yuwei Yin and Shicheng Li and Liang Chen and Peiyi Wang and Shuhuai Ren and Mukai Li and Yazheng Yang and Jingjing Xu and Xu Sun and Lingpeng Kong and Qi Liu},
      year={2023},
      eprint={2306.04387},
      archivePrefix={arXiv},
      primaryClass={cs.CV},
      url={https://arxiv.org/abs/2306.04387}, 
}

@misc{chen2023sharegpt4vimprovinglargemultimodal,
      title={ShareGPT4V: Improving Large Multi-Modal Models with Better Captions}, 
      author={Lin Chen and Jinsong Li and Xiaoyi Dong and Pan Zhang and Conghui He and Jiaqi Wang and Feng Zhao and Dahua Lin},
      year={2023},
      eprint={2311.12793},
      archivePrefix={arXiv},
      primaryClass={cs.CV},
      url={https://arxiv.org/abs/2311.12793}, 
}

@article{liu2023visual,
  title={Visual instruction tuning},
  author={Liu, Haotian and Li, Chunyuan and Wu, Qingyang and Lee, Yong Jae},
  journal={Advances in neural information processing systems},
  volume={36},
  pages={34892--34916},
  year={2023}
}

@article{han2024parameter,
  title={Parameter-efficient fine-tuning for large models: A comprehensive survey},
  author={Han, Zeyu and Gao, Chao and Liu, Jinyang and Zhang, Jeff and Zhang, Sai Qian},
  journal={arXiv preprint arXiv:2403.14608},
  year={2024}
}

@inproceedings{zhao2023unleashing,
  title={Unleashing text-to-image diffusion models for visual perception},
  author={Zhao, Wenliang and Rao, Yongming and Liu, Zuyan and Liu, Benlin and Zhou, Jie and Lu, Jiwen},
  booktitle={Proceedings of the IEEE/CVF international conference on computer vision},
  pages={5729--5739},
  year={2023}
}

@article{podell2023sdxl,
  title={Sdxl: Improving latent diffusion models for high-resolution image synthesis},
  author={Podell, Dustin and English, Zion and Lacey, Kyle and Blattmann, Andreas and Dockhorn, Tim and M{\"u}ller, Jonas and Penna, Joe and Rombach, Robin},
  journal={arXiv preprint arXiv:2307.01952},
  year={2023}
}

@inproceedings{cao2025image,
  title={Image is all you need to empower large-scale diffusion models for in-domain generation},
  author={Cao, Pu and Zhou, Feng and Yang, Lu and Huang, Tianrui and Song, Qing},
  booktitle={Proceedings of the Computer Vision and Pattern Recognition Conference},
  pages={18358--18368},
  year={2025}
}

@inproceedings{tong2025metamorph,
  title={Metamorph: Multimodal understanding and generation via instruction tuning},
  author={Tong, Shengbang and Fan, David and Li, Jiachen and Xiong, Yunyang and Chen, Xinlei and Sinha, Koustuv and Rabbat, Michael and LeCun, Yann and Xie, Saining and Liu, Zhuang},
  booktitle={Proceedings of the IEEE/CVF International Conference on Computer Vision},
  pages={17001--17012},
  year={2025}
}

@misc{zhuang2025vargptunifiedunderstandinggeneration,
      title={VARGPT: Unified Understanding and Generation in a Visual Autoregressive Multimodal Large Language Model}, 
      author={Xianwei Zhuang and Yuxin Xie and Yufan Deng and Liming Liang and Jinghan Ru and Yuguo Yin and Yuexian Zou},
      year={2025},
      eprint={2501.12327},
      archivePrefix={arXiv},
      primaryClass={cs.CV},
      url={https://arxiv.org/abs/2501.12327}, 
}

@inproceedings{chen2025comm,
  title={Comm: A coherent interleaved image-text dataset for multimodal understanding and generation},
  author={Chen, Wei and Li, Lin and Yang, Yongqi and Wen, Bin and Yang, Fan and Gao, Tingting and Wu, Yu and Chen, Long},
  booktitle={Proceedings of the Computer Vision and Pattern Recognition Conference},
  pages={8073--8082},
  year={2025}
}

@misc{shen2026danqinguptodatelargescalechinese,
      title={DanQing: An Up-to-Date Large-Scale Chinese Vision-Language Pre-training Dataset}, 
      author={Hengyu Shen and Tiancheng Gu and Bin Qin and Lan Wu and Yuling Wu and Shuo Tan and Zelong Sun and Jun Wang and Nan Wu and Xiang An and Weidong Cai and Ziyong Feng and Kaicheng Yang},
      year={2026},
      eprint={2601.10305},
      archivePrefix={arXiv},
      primaryClass={cs.CV},
      url={https://arxiv.org/abs/2601.10305}, 
}

@misc{tong2025thinkingvideovideogeneration,
      title={Thinking with Video: Video Generation as a Promising Multimodal Reasoning Paradigm}, 
      author={Jingqi Tong and Yurong Mou and Hangcheng Li and Mingzhe Li and Yongzhuo Yang and Ming Zhang and Qiguang Chen and Tianyi Liang and Xiaomeng Hu and Yining Zheng and Xinchi Chen and Jun Zhao and Xuanjing Huang and Xipeng Qiu},
      year={2025},
      eprint={2511.04570},
      archivePrefix={arXiv},
      primaryClass={cs.CV},
      url={https://arxiv.org/abs/2511.04570}, 
}

@article{xu2023wizardlm,
  title={Wizardlm: Empowering large language models to follow complex instructions},
  author={Xu, Can and Sun, Qingfeng and Zheng, Kai and Geng, Xiubo and Zhao, Pu and Feng, Jiazhan and Tao, Chongyang and Jiang, Daxin},
  journal={arXiv preprint arXiv:2304.12244},
  year={2023}
}

@article{zhao2024long,
  title={Long is more for alignment: A simple but tough-to-beat baseline for instruction fine-tuning},
  author={Zhao, Hao and Andriushchenko, Maksym and Croce, Francesco and Flammarion, Nicolas},
  journal={arXiv preprint arXiv:2402.04833},
  year={2024}
}

@article{rawte2023survey,
  title={A survey of hallucination in large foundation models},
  author={Rawte, Vipula and Sheth, Amit and Das, Amitava},
  journal={arXiv preprint arXiv:2309.05922},
  year={2023}
}

@article{ji2023survey,
  title={Survey of hallucination in natural language generation},
  author={Ji, Ziwei and Lee, Nayeon and Frieske, Rita and Yu, Tiezheng and Su, Dan and Xu, Yan and Ishii, Etsuko and Bang, Ye Jin and Madotto, Andrea and Fung, Pascale},
  journal={ACM computing surveys},
  volume={55},
  number={12},
  pages={1--38},
  year={2023},
  publisher={ACM New York, NY}
}

@inproceedings{wang2023self,
  title={Self-instruct: Aligning language models with self-generated instructions},
  author={Wang, Yizhong and Kordi, Yeganeh and Mishra, Swaroop and Liu, Alisa and Smith, Noah A and Khashabi, Daniel and Hajishirzi, Hannaneh},
  booktitle={Proceedings of the 61st annual meeting of the association for computational linguistics (volume 1: long papers)},
  pages={13484--13508},
  year={2023}
}

@misc{gao2025multimodalagenttuningbuilding,
      title={Multi-modal Agent Tuning: Building a VLM-Driven Agent for Efficient Tool Usage}, 
      author={Zhi Gao and Bofei Zhang and Pengxiang Li and Xiaojian Ma and Tao Yuan and Yue Fan and Yuwei Wu and Yunde Jia and Song-Chun Zhu and Qing Li},
      year={2025},
      eprint={2412.15606},
      archivePrefix={arXiv},
      primaryClass={cs.AI},
      url={https://arxiv.org/abs/2412.15606}, 
}

@article{deng2025toolscope,
  title={ToolScope: An Agentic Framework for Vision-Guided and Long-Horizon Tool Use},
  author={Deng, Mengjie and Dong, Guanting and Dou, Zhicheng},
  journal={arXiv preprint arXiv:2510.27363},
  year={2025}
}

@article{zhang2025vc,
  title={VC-Agent: An Interactive Agent for Customized Video Dataset Collection},
  author={Zhang, Yidan and Xu, Mutian and Hao, Yiming and Zhou, Kun and Chang, Jiahao and Liu, Xiaoqiang and Wan, Pengfei and Fu, Hongbo and Han, Xiaoguang},
  journal={arXiv preprint arXiv:2509.21291},
  year={2025}
}

@article{kumar2024mmctagent,
  title={Mmctagent: Multi-modal critical thinking agent framework for complex visual reasoning},
  author={Kumar, Somnath and Gadhia, Yash and Ganu, Tanuja and Nambi, Akshay},
  journal={arXiv preprint arXiv:2405.18358},
  year={2024}
}

@misc{reddy2025orionunifiedvisualagent,
      title={Orion: A Unified Visual Agent for Multimodal Perception, Advanced Visual Reasoning and Execution}, 
      author={N Dinesh Reddy and Dylan Snyder and Lona Kiragu and Mirajul Mohin and Shahrear Bin Amin and Sudeep Pillai},
      year={2025},
      eprint={2511.14210},
      archivePrefix={arXiv},
      primaryClass={cs.CV},
      url={https://arxiv.org/abs/2511.14210}, 
}

@article{sun2025datasetagent,
  title={DatasetAgent: A Novel Multi-Agent System for Auto-Constructing Datasets from Real-World Images},
  author={Sun, Haoran and Bian, Haoyu and Zeng, Shaoning and Rao, Yunbo and Xu, Xu and Mei, Lin and Gou, Jianping},
  journal={arXiv preprint arXiv:2507.08648},
  year={2025}
}

@article{joshi2025text,
  title={Text-VQA Aug: Pipelined Harnessing of Large Multimodal Models for Automated Synthesis},
  author={Joshi, Soham and Mishra, Shwet Kamal and Gopalakrishnan, Viswanath},
  journal={arXiv preprint arXiv:2511.02046},
  year={2025}
}

@inproceedings{shao2020finegym,
  title={Finegym: A hierarchical video dataset for fine-grained action understanding},
  author={Shao, Dian and Zhao, Yue and Dai, Bo and Lin, Dahua},
  booktitle={Proceedings of the IEEE/CVF conference on computer vision and pattern recognition},
  pages={2616--2625},
  year={2020}
}

@inproceedings{yu2020bdd100k,
  title={Bdd100k: A diverse driving dataset for heterogeneous multitask learning},
  author={Yu, Fisher and Chen, Haofeng and Wang, Xin and Xian, Wenqi and Chen, Yingying and Liu, Fangchen and Madhavan, Vashisht and Darrell, Trevor},
  booktitle={Proceedings of the IEEE/CVF conference on computer vision and pattern recognition},
  pages={2636--2645},
  year={2020}
}

@inproceedings{raja2025dataset,
  title={A Dataset for Programming-based Instructional Video Classification and Question Answering},
  author={Raja, Sana Javaid and Zafar, Adeel and Shoaib, Aqsa},
  booktitle={Proceedings of the First Workshop of Evaluation of Multi-Modal Generation},
  pages={1--9},
  year={2025}
}

@misc{bai2025qwen25vltechnicalreport,
      title={Qwen2.5-VL Technical Report}, 
      author={Shuai Bai and Keqin Chen and Xuejing Liu and Jialin Wang and Wenbin Ge and Sibo Song and Kai Dang and Peng Wang and Shijie Wang and Jun Tang and Humen Zhong and Yuanzhi Zhu and Mingkun Yang and Zhaohai Li and Jianqiang Wan and Pengfei Wang and Wei Ding and Zheren Fu and Yiheng Xu and Jiabo Ye and Xi Zhang and Tianbao Xie and Zesen Cheng and Hang Zhang and Zhibo Yang and Haiyang Xu and Junyang Lin},
      year={2025},
      eprint={2502.13923},
      archivePrefix={arXiv},
      primaryClass={cs.CV},
      url={https://arxiv.org/abs/2502.13923}, 
}

@misc{grattafiori2024llama3herdmodels,
      title={The Llama 3 Herd of Models}, 
      author={Aaron Grattafiori and Abhimanyu Dubey and Abhinav Jauhri and Abhinav Pandey and Abhishek Kadian and Ahmad Al-Dahle and Aiesha Letman and Akhil Mathur and Alan Schelten and Alex Vaughan and Amy Yang and Angela Fan and Anirudh Goyal and Anthony Hartshorn and Aobo Yang and Archi Mitra and Archie Sravankumar and Artem Korenev and Arthur Hinsvark and Arun Rao and Aston Zhang and Aurelien Rodriguez and Austen Gregerson and Ava Spataru and Baptiste Roziere and Bethany Biron and Binh Tang and Bobbie Chern and Charlotte Caucheteux and Chaya Nayak and Chloe Bi and Chris Marra and Chris McConnell and Christian Keller and Christophe Touret and Chunyang Wu and Corinne Wong and Cristian Canton Ferrer and Cyrus Nikolaidis and Damien Allonsius and Daniel Song and Danielle Pintz and Danny Livshits and Danny Wyatt and David Esiobu and Dhruv Choudhary and Dhruv Mahajan and Diego Garcia-Olano and Diego Perino and Dieuwke Hupkes and Egor Lakomkin and Ehab AlBadawy and Elina Lobanova and Emily Dinan and Eric Michael Smith and Filip Radenovic and Francisco Guzmán and Frank Zhang and Gabriel Synnaeve and Gabrielle Lee and Georgia Lewis Anderson and Govind Thattai and Graeme Nail and Gregoire Mialon and Guan Pang and Guillem Cucurell and Hailey Nguyen and Hannah Korevaar and Hu Xu and Hugo Touvron and Iliyan Zarov and Imanol Arrieta Ibarra and Isabel Kloumann and Ishan Misra and Ivan Evtimov and Jack Zhang and Jade Copet and Jaewon Lee and Jan Geffert and Jana Vranes and Jason Park and Jay Mahadeokar and Jeet Shah and Jelmer van der Linde and Jennifer Billock and Jenny Hong and Jenya Lee and Jeremy Fu and Jianfeng Chi and Jianyu Huang and Jiawen Liu and Jie Wang and Jiecao Yu and Joanna Bitton and Joe Spisak and Jongsoo Park and Joseph Rocca and Joshua Johnstun and Joshua Saxe and Junteng Jia and Kalyan Vasuden Alwala and Karthik Prasad and Kartikeya Upasani and Kate Plawiak and Ke Li and Kenneth Heafield and Kevin Stone and Khalid El-Arini and Krithika Iyer and Kshitiz Malik and Kuenley Chiu and Kunal Bhalla and Kushal Lakhotia and Lauren Rantala-Yeary and Laurens van der Maaten and Lawrence Chen and Liang Tan and Liz Jenkins and Louis Martin and Lovish Madaan and Lubo Malo and Lukas Blecher and Lukas Landzaat and Luke de Oliveira and Madeline Muzzi and Mahesh Pasupuleti and Mannat Singh and Manohar Paluri and Marcin Kardas and Maria Tsimpoukelli and Mathew Oldham and Mathieu Rita and Maya Pavlova and Melanie Kambadur and Mike Lewis and Min Si and Mitesh Kumar Singh and Mona Hassan and Naman Goyal and Narjes Torabi and Nikolay Bashlykov and Nikolay Bogoychev and Niladri Chatterji and Ning Zhang and Olivier Duchenne and Onur Çelebi and Patrick Alrassy and Pengchuan Zhang and Pengwei Li and Petar Vasic and Peter Weng and Prajjwal Bhargava and Pratik Dubal and Praveen Krishnan and Punit Singh Koura and Puxin Xu and Qing He and Qingxiao Dong and Ragavan Srinivasan and Raj Ganapathy and Ramon Calderer and Ricardo Silveira Cabral and Robert Stojnic and Roberta Raileanu and Rohan Maheswari and Rohit Girdhar and Rohit Patel and Romain Sauvestre and Ronnie Polidoro and Roshan Sumbaly and Ross Taylor and Ruan Silva and Rui Hou and Rui Wang and Saghar Hosseini and Sahana Chennabasappa and Sanjay Singh and Sean Bell and Seohyun Sonia Kim and Sergey Edunov and Shaoliang Nie and Sharan Narang and Sharath Raparthy and Sheng Shen and Shengye Wan and Shruti Bhosale and Shun Zhang and Simon Vandenhende and Soumya Batra and Spencer Whitman and Sten Sootla and Stephane Collot and Suchin Gururangan and Sydney Borodinsky and Tamar Herman and Tara Fowler and Tarek Sheasha and Thomas Georgiou and Thomas Scialom and Tobias Speckbacher and Todor Mihaylov and Tong Xiao and Ujjwal Karn and Vedanuj Goswami and Vibhor Gupta and Vignesh Ramanathan and Viktor Kerkez and Vincent Gonguet and Virginie Do and Vish Vogeti and Vítor Albiero and Vladan Petrovic and Weiwei Chu and Wenhan Xiong and Wenyin Fu and Whitney Meers and Xavier Martinet and Xiaodong Wang and Xiaofang Wang and Xiaoqing Ellen Tan and Xide Xia and Xinfeng Xie and Xuchao Jia and Xuewei Wang and Yaelle Goldschlag and Yashesh Gaur and Yasmine Babaei and Yi Wen and Yiwen Song and Yuchen Zhang and Yue Li and Yuning Mao and Zacharie Delpierre Coudert and Zheng Yan and Zhengxing Chen and Zoe Papakipos and Aaditya Singh and Aayushi Srivastava and Abha Jain and Adam Kelsey and Adam Shajnfeld and Adithya Gangidi and Adolfo Victoria and Ahuva Goldstand and Ajay Menon and Ajay Sharma and Alex Boesenberg and Alexei Baevski and Allie Feinstein and Amanda Kallet and Amit Sangani and Amos Teo and Anam Yunus and Andrei Lupu and Andres Alvarado and Andrew Caples and Andrew Gu and Andrew Ho and Andrew Poulton and Andrew Ryan and Ankit Ramchandani and Annie Dong and Annie Franco and Anuj Goyal and Aparajita Saraf and Arkabandhu Chowdhury and Ashley Gabriel and Ashwin Bharambe and Assaf Eisenman and Azadeh Yazdan and Beau James and Ben Maurer and Benjamin Leonhardi and Bernie Huang and Beth Loyd and Beto De Paola and Bhargavi Paranjape and Bing Liu and Bo Wu and Boyu Ni and Braden Hancock and Bram Wasti and Brandon Spence and Brani Stojkovic and Brian Gamido and Britt Montalvo and Carl Parker and Carly Burton and Catalina Mejia and Ce Liu and Changhan Wang and Changkyu Kim and Chao Zhou and Chester Hu and Ching-Hsiang Chu and Chris Cai and Chris Tindal and Christoph Feichtenhofer and Cynthia Gao and Damon Civin and Dana Beaty and Daniel Kreymer and Daniel Li and David Adkins and David Xu and Davide Testuggine and Delia David and Devi Parikh and Diana Liskovich and Didem Foss and Dingkang Wang and Duc Le and Dustin Holland and Edward Dowling and Eissa Jamil and Elaine Montgomery and Eleonora Presani and Emily Hahn and Emily Wood and Eric-Tuan Le and Erik Brinkman and Esteban Arcaute and Evan Dunbar and Evan Smothers and Fei Sun and Felix Kreuk and Feng Tian and Filippos Kokkinos and Firat Ozgenel and Francesco Caggioni and Frank Kanayet and Frank Seide and Gabriela Medina Florez and Gabriella Schwarz and Gada Badeer and Georgia Swee and Gil Halpern and Grant Herman and Grigory Sizov and Guangyi and Zhang and Guna Lakshminarayanan and Hakan Inan and Hamid Shojanazeri and Han Zou and Hannah Wang and Hanwen Zha and Haroun Habeeb and Harrison Rudolph and Helen Suk and Henry Aspegren and Hunter Goldman and Hongyuan Zhan and Ibrahim Damlaj and Igor Molybog and Igor Tufanov and Ilias Leontiadis and Irina-Elena Veliche and Itai Gat and Jake Weissman and James Geboski and James Kohli and Janice Lam and Japhet Asher and Jean-Baptiste Gaya and Jeff Marcus and Jeff Tang and Jennifer Chan and Jenny Zhen and Jeremy Reizenstein and Jeremy Teboul and Jessica Zhong and Jian Jin and Jingyi Yang and Joe Cummings and Jon Carvill and Jon Shepard and Jonathan McPhie and Jonathan Torres and Josh Ginsburg and Junjie Wang and Kai Wu and Kam Hou U and Karan Saxena and Kartikay Khandelwal and Katayoun Zand and Kathy Matosich and Kaushik Veeraraghavan and Kelly Michelena and Keqian Li and Kiran Jagadeesh and Kun Huang and Kunal Chawla and Kyle Huang and Lailin Chen and Lakshya Garg and Lavender A and Leandro Silva and Lee Bell and Lei Zhang and Liangpeng Guo and Licheng Yu and Liron Moshkovich and Luca Wehrstedt and Madian Khabsa and Manav Avalani and Manish Bhatt and Martynas Mankus and Matan Hasson and Matthew Lennie and Matthias Reso and Maxim Groshev and Maxim Naumov and Maya Lathi and Meghan Keneally and Miao Liu and Michael L. Seltzer and Michal Valko and Michelle Restrepo and Mihir Patel and Mik Vyatskov and Mikayel Samvelyan and Mike Clark and Mike Macey and Mike Wang and Miquel Jubert Hermoso and Mo Metanat and Mohammad Rastegari and Munish Bansal and Nandhini Santhanam and Natascha Parks and Natasha White and Navyata Bawa and Nayan Singhal and Nick Egebo and Nicolas Usunier and Nikhil Mehta and Nikolay Pavlovich Laptev and Ning Dong and Norman Cheng and Oleg Chernoguz and Olivia Hart and Omkar Salpekar and Ozlem Kalinli and Parkin Kent and Parth Parekh and Paul Saab and Pavan Balaji and Pedro Rittner and Philip Bontrager and Pierre Roux and Piotr Dollar and Polina Zvyagina and Prashant Ratanchandani and Pritish Yuvraj and Qian Liang and Rachad Alao and Rachel Rodriguez and Rafi Ayub and Raghotham Murthy and Raghu Nayani and Rahul Mitra and Rangaprabhu Parthasarathy and Raymond Li and Rebekkah Hogan and Robin Battey and Rocky Wang and Russ Howes and Ruty Rinott and Sachin Mehta and Sachin Siby and Sai Jayesh Bondu and Samyak Datta and Sara Chugh and Sara Hunt and Sargun Dhillon and Sasha Sidorov and Satadru Pan and Saurabh Mahajan and Saurabh Verma and Seiji Yamamoto and Sharadh Ramaswamy and Shaun Lindsay and Shaun Lindsay and Sheng Feng and Shenghao Lin and Shengxin Cindy Zha and Shishir Patil and Shiva Shankar and Shuqiang Zhang and Shuqiang Zhang and Sinong Wang and Sneha Agarwal and Soji Sajuyigbe and Soumith Chintala and Stephanie Max and Stephen Chen and Steve Kehoe and Steve Satterfield and Sudarshan Govindaprasad and Sumit Gupta and Summer Deng and Sungmin Cho and Sunny Virk and Suraj Subramanian and Sy Choudhury and Sydney Goldman and Tal Remez and Tamar Glaser and Tamara Best and Thilo Koehler and Thomas Robinson and Tianhe Li and Tianjun Zhang and Tim Matthews and Timothy Chou and Tzook Shaked and Varun Vontimitta and Victoria Ajayi and Victoria Montanez and Vijai Mohan and Vinay Satish Kumar and Vishal Mangla and Vlad Ionescu and Vlad Poenaru and Vlad Tiberiu Mihailescu and Vladimir Ivanov and Wei Li and Wenchen Wang and Wenwen Jiang and Wes Bouaziz and Will Constable and Xiaocheng Tang and Xiaojian Wu and Xiaolan Wang and Xilun Wu and Xinbo Gao and Yaniv Kleinman and Yanjun Chen and Ye Hu and Ye Jia and Ye Qi and Yenda Li and Yilin Zhang and Ying Zhang and Yossi Adi and Youngjin Nam and Yu and Wang and Yu Zhao and Yuchen Hao and Yundi Qian and Yunlu Li and Yuzi He and Zach Rait and Zachary DeVito and Zef Rosnbrick and Zhaoduo Wen and Zhenyu Yang and Zhiwei Zhao and Zhiyu Ma},
      year={2024},
      eprint={2407.21783},
      archivePrefix={arXiv},
      primaryClass={cs.AI},
      url={https://arxiv.org/abs/2407.21783}, 
}

@misc{zhu2025internvl3exploringadvancedtraining,
      title={InternVL3: Exploring Advanced Training and Test-Time Recipes for Open-Source Multimodal Models}, 
      author={Jinguo Zhu and Weiyun Wang and Zhe Chen and Zhaoyang Liu and Shenglong Ye and Lixin Gu and Hao Tian and Yuchen Duan and Weijie Su and Jie Shao and Zhangwei Gao and Erfei Cui and Xuehui Wang and Yue Cao and Yangzhou Liu and Xingguang Wei and Hongjie Zhang and Haomin Wang and Weiye Xu and Hao Li and Jiahao Wang and Nianchen Deng and Songze Li and Yinan He and Tan Jiang and Jiapeng Luo and Yi Wang and Conghui He and Botian Shi and Xingcheng Zhang and Wenqi Shao and Junjun He and Yingtong Xiong and Wenwen Qu and Peng Sun and Penglong Jiao and Han Lv and Lijun Wu and Kaipeng Zhang and Huipeng Deng and Jiaye Ge and Kai Chen and Limin Wang and Min Dou and Lewei Lu and Xizhou Zhu and Tong Lu and Dahua Lin and Yu Qiao and Jifeng Dai and Wenhai Wang},
      year={2025},
      eprint={2504.10479},
      archivePrefix={arXiv},
      primaryClass={cs.CV},
      url={https://arxiv.org/abs/2504.10479}, 
}

@misc{radford2022whisper,
  doi = {10.48550/ARXIV.2212.04356},
  url = {https://arxiv.org/abs/2212.04356},
  author = {Radford, Alec and Kim, Jong Wook and Xu, Tao and Brockman, Greg and McLeavey, Christine and Sutskever, Ilya},
  title = {Robust Speech Recognition via Large-Scale Weak Supervision},
  publisher = {arXiv},
  year = {2022},
  copyright = {arXiv.org perpetual, non-exclusive license}
}

@inproceedings{NEURIPS2024_10570531,
 author = {Liu, Ziyu and Chu, Tao and Zang, Yuhang and Wei, Xilin and Dong, Xiaoyi and Zhang, Pan and Liang, Zijian and Xiong, Yuanjun and Qiao, Yu and Lin, Dahua and Wang, Jiaqi},
 booktitle = {Advances in Neural Information Processing Systems},
 doi = {10.52202/079017-0278},
 editor = {A. Globerson and L. Mackey and D. Belgrave and A. Fan and U. Paquet and J. Tomczak and C. Zhang},
 pages = {8698--8733},
 publisher = {Curran Associates, Inc.},
 title = {MMDU: A Multi-Turn Multi-Image Dialog Understanding Benchmark and Instruction-Tuning Dataset for LVLMs},
 url = {https://proceedings.neurips.cc/paper_files/paper/2024/file/1057053100de064a44286239724f7865-Paper-Datasets_and_Benchmarks_Track.pdf},
 volume = {37},
 year = {2024}
}

@misc{liu2024improvedbaselinesvisualinstruction,
      title={Improved Baselines with Visual Instruction Tuning}, 
      author={Haotian Liu and Chunyuan Li and Yuheng Li and Yong Jae Lee},
      year={2024},
      eprint={2310.03744},
      archivePrefix={arXiv},
      primaryClass={cs.CV},
      url={https://arxiv.org/abs/2310.03744}, 
}

@misc{wang2025ovisu1technicalreport,
      title={Ovis-U1 Technical Report}, 
      author={Guo-Hua Wang and Shanshan Zhao and Xinjie Zhang and Liangfu Cao and Pengxin Zhan and Lunhao Duan and Shiyin Lu and Minghao Fu and Xiaohao Chen and Jianshan Zhao and Yang Li and Qing-Guo Chen},
      year={2025},
      eprint={2506.23044},
      archivePrefix={arXiv},
      primaryClass={cs.CV},
      url={https://arxiv.org/abs/2506.23044}, 
}

@inproceedings{
  li2026zebracot,
  title={Zebra-CoT: A Dataset for Interleaved Vision-Language Reasoning},
  author={Ang Li and Charles Wang and Deqing Fu and Kaiyu Yue and Zikui Cai
          and Wang Bill Zhu and Ollie Liu and Peng Guo and Willie Neiswanger
          and Furong Huang and Tom Goldstein and Micah Goldblum},
  booktitle={The Fourteenth International Conference on Learning Representations},
  year={2026},
  url={https://openreview.net/forum?id=c6XIVI3TiQ}
}

@misc{liu2025step1xeditpracticalframeworkgeneral,
      title={Step1X-Edit: A Practical Framework for General Image Editing}, 
      author={Shiyu Liu and Yucheng Han and Peng Xing and Fukun Yin and Rui Wang and Wei Cheng and Jiaqi Liao and Yingming Wang and Honghao Fu and Chunrui Han and Guopeng Li and Yuang Peng and Quan Sun and Jingwei Wu and Yan Cai and Zheng Ge and Ranchen Ming and Lei Xia and Xianfang Zeng and Yibo Zhu and Binxing Jiao and Xiangyu Zhang and Gang Yu and Daxin Jiang},
      year={2025},
      eprint={2504.17761},
      archivePrefix={arXiv},
      primaryClass={cs.CV},
      url={https://arxiv.org/abs/2504.17761}, 
}

@misc{xiao2024omnigenunifiedimagegeneration,
      title={OmniGen: Unified Image Generation}, 
      author={Shitao Xiao and Yueze Wang and Junjie Zhou and Huaying Yuan and Xingrun Xing and Ruiran Yan and Chaofan Li and Shuting Wang and Tiejun Huang and Zheng Liu},
      year={2024},
      eprint={2409.11340},
      archivePrefix={arXiv},
      primaryClass={cs.CV},
      url={https://arxiv.org/abs/2409.11340}, 
}

@misc{zhang2024magicbrushmanuallyannotateddataset,
      title={MagicBrush: A Manually Annotated Dataset for Instruction-Guided Image Editing}, 
      author={Kai Zhang and Lingbo Mo and Wenhu Chen and Huan Sun and Yu Su},
      year={2024},
      eprint={2306.10012},
      archivePrefix={arXiv},
      primaryClass={cs.CV},
      url={https://arxiv.org/abs/2306.10012}, 
}

@misc{brooks2023instructpix2pixlearningfollowimage,
      title={InstructPix2Pix: Learning to Follow Image Editing Instructions}, 
      author={Tim Brooks and Aleksander Holynski and Alexei A. Efros},
      year={2023},
      eprint={2211.09800},
      archivePrefix={arXiv},
      primaryClass={cs.CV},
      url={https://arxiv.org/abs/2211.09800}, 
}

@article{nan2024openvid,
  title={Openvid-1m: A large-scale high-quality dataset for text-to-video generation},
  author={Nan, Kepan and Xie, Rui and Zhou, Penghao and Fan, Tiehan and Yang, Zhenheng and Chen, Zhijie and Li, Xiang and Yang, Jian and Tai, Ying},
  journal={arXiv preprint arXiv:2407.02371},
  year={2024}
}

@inproceedings{chen2024panda,
  title={Panda-70m: Captioning 70m videos with multiple cross-modality teachers},
  author={Chen, Tsai-Shien and Siarohin, Aliaksandr and Menapace, Willi and Deyneka, Ekaterina and Chao, Hsiang-wei and Jeon, Byung Eun and Fang, Yuwei and Lee, Hsin-Ying and Ren, Jian and Yang, Ming-Hsuan and others},
  booktitle={Proceedings of the IEEE/CVF Conference on Computer Vision and Pattern Recognition},
  pages={13320--13331},
  year={2024}
}

@misc{chen2025janusprounifiedmultimodalunderstanding,
      title={Janus-Pro: Unified Multimodal Understanding and Generation with Data and Model Scaling}, 
      author={Xiaokang Chen and Zhiyu Wu and Xingchao Liu and Zizheng Pan and Wen Liu and Zhenda Xie and Xingkai Yu and Chong Ruan},
      year={2025},
      eprint={2501.17811},
      archivePrefix={arXiv},
      primaryClass={cs.AI},
      url={https://arxiv.org/abs/2501.17811}, 
}

@misc{gemini2flashimage2025,
  title        = {Experiment with Gemini 2.0 Flash Native Image Generation},
  author       = {Kampf, Kat and Brichtova, Nicole},
  year         = {2025},
  month        = mar,
  howpublished = {\url{https://developers.googleblog.com/experiment-with-gemini-20-flash-native-image-generation/}},
  note         = {Google Developers Blog}
}

@article{yu2024anyedit,
  title         = {AnyEdit: Mastering Unified High-Quality Image Editing for Any Idea},
  author        = {Yu, Qifan and Chow, Wei and Yue, Zhongqi and Pan, Kaihang and Wu, Yang and Wan, Xiaoyang and Li, Juncheng and Tang, Siliang and Zhang, Hanwang and Zhuang, Yueting},
  journal       = {arXiv preprint arXiv:2411.15738},
  year          = {2024},
  eprint        = {2411.15738},
  archivePrefix = {arXiv},
  primaryClass  = {cs.CV},
  url           = {https://arxiv.org/abs/2411.15738}
}

@misc{qu2024tokenflowunifiedimagetokenizer,
      title={TokenFlow: Unified Image Tokenizer for Multimodal Understanding and Generation},
      author={Liao Qu and Huichao Zhang and Yiheng Liu and Xu Wang and Yi Jiang and Yiming Gao and Hu Ye and Daniel K. Du and Zehuan Yuan and Xinglong Wu},
      year={2024},
      eprint={2412.03069},
      archivePrefix={arXiv},
      primaryClass={cs.CV},
      url={https://arxiv.org/abs/2412.03069},
}

@misc{lu2024omniparserpurevisionbased,
      title={OmniParser for Pure Vision Based GUI Agent},
      author={Yadong Lu and Jianwei Yang and Yelong Shen and Ahmed Awadallah},
      year={2024},
      eprint={2408.00203},
      archivePrefix={arXiv},
      primaryClass={cs.CV},
      url={https://arxiv.org/abs/2408.00203},
}

@misc{castellano_pyscenedetect,
  author       = {Castellano, Brandon},
  title        = {PySceneDetect},
  howpublished = {\url{https://github.com/Breakthrough/PySceneDetect}},
}

@article{bai2025qwen3,
  title={Qwen3-vl technical report},
  author={Bai, Shuai and Cai, Yuxuan and Chen, Ruizhe and Chen, Keqin and Chen, Xionghui and Cheng, Zesen and Deng, Lianghao and Ding, Wei and Gao, Chang and Ge, Chunjiang and others},
  journal={arXiv preprint arXiv:2511.21631},
  year={2025}
}

@misc{qwen3.5,
    title  = {{Qwen3.5}: Towards Native Multimodal Agents},
    author = {{Qwen Team}},
    month  = {February},
    year   = {2026},
    url    = {https://qwen.ai/blog?id=qwen3.5}
}

@misc{openai2024gpt4ocard,
      title={GPT-4o System Card}, 
      author={OpenAI and : and Aaron Hurst and Adam Lerer and Adam P. Goucher and Adam Perelman and Aditya Ramesh and Aidan Clark and AJ Ostrow and Akila Welihinda and Alan Hayes and Alec Radford and Aleksander Mądry and Alex Baker-Whitcomb and Alex Beutel and Alex Borzunov and Alex Carney and Alex Chow and Alex Kirillov and Alex Nichol and Alex Paino and Alex Renzin and Alex Tachard Passos and Alexander Kirillov and Alexi Christakis and Alexis Conneau and Ali Kamali and Allan Jabri and Allison Moyer and Allison Tam and Amadou Crookes and Amin Tootoochian and Amin Tootoonchian and Ananya Kumar and Andrea Vallone and Andrej Karpathy and Andrew Braunstein and Andrew Cann and Andrew Codispoti and Andrew Galu and Andrew Kondrich and Andrew Tulloch and Andrey Mishchenko and Angela Baek and Angela Jiang and Antoine Pelisse and Antonia Woodford and Anuj Gosalia and Arka Dhar and Ashley Pantuliano and Avi Nayak and Avital Oliver and Barret Zoph and Behrooz Ghorbani and Ben Leimberger and Ben Rossen and Ben Sokolowsky and Ben Wang and Benjamin Zweig and Beth Hoover and Blake Samic and Bob McGrew and Bobby Spero and Bogo Giertler and Bowen Cheng and Brad Lightcap and Brandon Walkin and Brendan Quinn and Brian Guarraci and Brian Hsu and Bright Kellogg and Brydon Eastman and Camillo Lugaresi and Carroll Wainwright and Cary Bassin and Cary Hudson and Casey Chu and Chad Nelson and Chak Li and Chan Jun Shern and Channing Conger and Charlotte Barette and Chelsea Voss and Chen Ding and Cheng Lu and Chong Zhang and Chris Beaumont and Chris Hallacy and Chris Koch and Christian Gibson and Christina Kim and Christine Choi and Christine McLeavey and Christopher Hesse and Claudia Fischer and Clemens Winter and Coley Czarnecki and Colin Jarvis and Colin Wei and Constantin Koumouzelis and Dane Sherburn and Daniel Kappler and Daniel Levin and Daniel Levy and David Carr and David Farhi and David Mely and David Robinson and David Sasaki and Denny Jin and Dev Valladares and Dimitris Tsipras and Doug Li and Duc Phong Nguyen and Duncan Findlay and Edede Oiwoh and Edmund Wong and Ehsan Asdar and Elizabeth Proehl and Elizabeth Yang and Eric Antonow and Eric Kramer and Eric Peterson and Eric Sigler and Eric Wallace and Eugene Brevdo and Evan Mays and Farzad Khorasani and Felipe Petroski Such and Filippo Raso and Francis Zhang and Fred von Lohmann and Freddie Sulit and Gabriel Goh and Gene Oden and Geoff Salmon and Giulio Starace and Greg Brockman and Hadi Salman and Haiming Bao and Haitang Hu and Hannah Wong and Haoyu Wang and Heather Schmidt and Heather Whitney and Heewoo Jun and Hendrik Kirchner and Henrique Ponde de Oliveira Pinto and Hongyu Ren and Huiwen Chang and Hyung Won Chung and Ian Kivlichan and Ian O'Connell and Ian O'Connell and Ian Osband and Ian Silber and Ian Sohl and Ibrahim Okuyucu and Ikai Lan and Ilya Kostrikov and Ilya Sutskever and Ingmar Kanitscheider and Ishaan Gulrajani and Jacob Coxon and Jacob Menick and Jakub Pachocki and James Aung and James Betker and James Crooks and James Lennon and Jamie Kiros and Jan Leike and Jane Park and Jason Kwon and Jason Phang and Jason Teplitz and Jason Wei and Jason Wolfe and Jay Chen and Jeff Harris and Jenia Varavva and Jessica Gan Lee and Jessica Shieh and Ji Lin and Jiahui Yu and Jiayi Weng and Jie Tang and Jieqi Yu and Joanne Jang and Joaquin Quinonero Candela and Joe Beutler and Joe Landers and Joel Parish and Johannes Heidecke and John Schulman and Jonathan Lachman and Jonathan McKay and Jonathan Uesato and Jonathan Ward and Jong Wook Kim and Joost Huizinga and Jordan Sitkin and Jos Kraaijeveld and Josh Gross and Josh Kaplan and Josh Snyder and Joshua Achiam and Joy Jiao and Joyce Lee and Juntang Zhuang and Justyn Harriman and Kai Fricke and Kai Hayashi and Karan Singhal and Katy Shi and Kavin Karthik and Kayla Wood and Kendra Rimbach and Kenny Hsu and Kenny Nguyen and Keren Gu-Lemberg and Kevin Button and Kevin Liu and Kiel Howe and Krithika Muthukumar and Kyle Luther and Lama Ahmad and Larry Kai and Lauren Itow and Lauren Workman and Leher Pathak and Leo Chen and Li Jing and Lia Guy and Liam Fedus and Liang Zhou and Lien Mamitsuka and Lilian Weng and Lindsay McCallum and Lindsey Held and Long Ouyang and Louis Feuvrier and Lu Zhang and Lukas Kondraciuk and Lukasz Kaiser and Luke Hewitt and Luke Metz and Lyric Doshi and Mada Aflak and Maddie Simens and Madelaine Boyd and Madeleine Thompson and Marat Dukhan and Mark Chen and Mark Gray and Mark Hudnall and Marvin Zhang and Marwan Aljubeh and Mateusz Litwin and Matthew Zeng and Max Johnson and Maya Shetty and Mayank Gupta and Meghan Shah and Mehmet Yatbaz and Meng Jia Yang and Mengchao Zhong and Mia Glaese and Mianna Chen and Michael Janner and Michael Lampe and Michael Petrov and Michael Wu and Michele Wang and Michelle Fradin and Michelle Pokrass and Miguel Castro and Miguel Oom Temudo de Castro and Mikhail Pavlov and Miles Brundage and Miles Wang and Minal Khan and Mira Murati and Mo Bavarian and Molly Lin and Murat Yesildal and Nacho Soto and Natalia Gimelshein and Natalie Cone and Natalie Staudacher and Natalie Summers and Natan LaFontaine and Neil Chowdhury and Nick Ryder and Nick Stathas and Nick Turley and Nik Tezak and Niko Felix and Nithanth Kudige and Nitish Keskar and Noah Deutsch and Noel Bundick and Nora Puckett and Ofir Nachum and Ola Okelola and Oleg Boiko and Oleg Murk and Oliver Jaffe and Olivia Watkins and Olivier Godement and Owen Campbell-Moore and Patrick Chao and Paul McMillan and Pavel Belov and Peng Su and Peter Bak and Peter Bakkum and Peter Deng and Peter Dolan and Peter Hoeschele and Peter Welinder and Phil Tillet and Philip Pronin and Philippe Tillet and Prafulla Dhariwal and Qiming Yuan and Rachel Dias and Rachel Lim and Rahul Arora and Rajan Troll and Randall Lin and Rapha Gontijo Lopes and Raul Puri and Reah Miyara and Reimar Leike and Renaud Gaubert and Reza Zamani and Ricky Wang and Rob Donnelly and Rob Honsby and Rocky Smith and Rohan Sahai and Rohit Ramchandani and Romain Huet and Rory Carmichael and Rowan Zellers and Roy Chen and Ruby Chen and Ruslan Nigmatullin and Ryan Cheu and Saachi Jain and Sam Altman and Sam Schoenholz and Sam Toizer and Samuel Miserendino and Sandhini Agarwal and Sara Culver and Scott Ethersmith and Scott Gray and Sean Grove and Sean Metzger and Shamez Hermani and Shantanu Jain and Shengjia Zhao and Sherwin Wu and Shino Jomoto and Shirong Wu and Shuaiqi and Xia and Sonia Phene and Spencer Papay and Srinivas Narayanan and Steve Coffey and Steve Lee and Stewart Hall and Suchir Balaji and Tal Broda and Tal Stramer and Tao Xu and Tarun Gogineni and Taya Christianson and Ted Sanders and Tejal Patwardhan and Thomas Cunninghman and Thomas Degry and Thomas Dimson and Thomas Raoux and Thomas Shadwell and Tianhao Zheng and Todd Underwood and Todor Markov and Toki Sherbakov and Tom Rubin and Tom Stasi and Tomer Kaftan and Tristan Heywood and Troy Peterson and Tyce Walters and Tyna Eloundou and Valerie Qi and Veit Moeller and Vinnie Monaco and Vishal Kuo and Vlad Fomenko and Wayne Chang and Weiyi Zheng and Wenda Zhou and Wesam Manassra and Will Sheu and Wojciech Zaremba and Yash Patil and Yilei Qian and Yongjik Kim and Youlong Cheng and Yu Zhang and Yuchen He and Yuchen Zhang and Yujia Jin and Yunxing Dai and Yury Malkov},
      year={2024},
      eprint={2410.21276},
      archivePrefix={arXiv},
      primaryClass={cs.CL},
      url={https://arxiv.org/abs/2410.21276}, 
}

@misc{openai2025gpt41,
  author       = {OpenAI},
  title        = {Introducing GPT-4.1 in the API},
  year         = {2025},
  month        = {April},
  url          = {https://openai.com/index/gpt-4-1/}
}

@misc{deepseekai2025deepseekv3technicalreport,
      title={DeepSeek-V3 Technical Report}, 
      author={DeepSeek-AI and Aixin Liu and Bei Feng and Bing Xue and Bingxuan Wang and Bochao Wu and Chengda Lu and Chenggang Zhao and Chengqi Deng and Chenyu Zhang and Chong Ruan and Damai Dai and Daya Guo and Dejian Yang and Deli Chen and Dongjie Ji and Erhang Li and Fangyun Lin and Fucong Dai and Fuli Luo and Guangbo Hao and Guanting Chen and Guowei Li and H. Zhang and Han Bao and Hanwei Xu and Haocheng Wang and Haowei Zhang and Honghui Ding and Huajian Xin and Huazuo Gao and Hui Li and Hui Qu and J. L. Cai and Jian Liang and Jianzhong Guo and Jiaqi Ni and Jiashi Li and Jiawei Wang and Jin Chen and Jingchang Chen and Jingyang Yuan and Junjie Qiu and Junlong Li and Junxiao Song and Kai Dong and Kai Hu and Kaige Gao and Kang Guan and Kexin Huang and Kuai Yu and Lean Wang and Lecong Zhang and Lei Xu and Leyi Xia and Liang Zhao and Litong Wang and Liyue Zhang and Meng Li and Miaojun Wang and Mingchuan Zhang and Minghua Zhang and Minghui Tang and Mingming Li and Ning Tian and Panpan Huang and Peiyi Wang and Peng Zhang and Qiancheng Wang and Qihao Zhu and Qinyu Chen and Qiushi Du and R. J. Chen and R. L. Jin and Ruiqi Ge and Ruisong Zhang and Ruizhe Pan and Runji Wang and Runxin Xu and Ruoyu Zhang and Ruyi Chen and S. S. Li and Shanghao Lu and Shangyan Zhou and Shanhuang Chen and Shaoqing Wu and Shengfeng Ye and Shengfeng Ye and Shirong Ma and Shiyu Wang and Shuang Zhou and Shuiping Yu and Shunfeng Zhou and Shuting Pan and T. Wang and Tao Yun and Tian Pei and Tianyu Sun and W. L. Xiao and Wangding Zeng and Wanjia Zhao and Wei An and Wen Liu and Wenfeng Liang and Wenjun Gao and Wenqin Yu and Wentao Zhang and X. Q. Li and Xiangyue Jin and Xianzu Wang and Xiao Bi and Xiaodong Liu and Xiaohan Wang and Xiaojin Shen and Xiaokang Chen and Xiaokang Zhang and Xiaosha Chen and Xiaotao Nie and Xiaowen Sun and Xiaoxiang Wang and Xin Cheng and Xin Liu and Xin Xie and Xingchao Liu and Xingkai Yu and Xinnan Song and Xinxia Shan and Xinyi Zhou and Xinyu Yang and Xinyuan Li and Xuecheng Su and Xuheng Lin and Y. K. Li and Y. Q. Wang and Y. X. Wei and Y. X. Zhu and Yang Zhang and Yanhong Xu and Yanhong Xu and Yanping Huang and Yao Li and Yao Zhao and Yaofeng Sun and Yaohui Li and Yaohui Wang and Yi Yu and Yi Zheng and Yichao Zhang and Yifan Shi and Yiliang Xiong and Ying He and Ying Tang and Yishi Piao and Yisong Wang and Yixuan Tan and Yiyang Ma and Yiyuan Liu and Yongqiang Guo and Yu Wu and Yuan Ou and Yuchen Zhu and Yuduan Wang and Yue Gong and Yuheng Zou and Yujia He and Yukun Zha and Yunfan Xiong and Yunxian Ma and Yuting Yan and Yuxiang Luo and Yuxiang You and Yuxuan Liu and Yuyang Zhou and Z. F. Wu and Z. Z. Ren and Zehui Ren and Zhangli Sha and Zhe Fu and Zhean Xu and Zhen Huang and Zhen Zhang and Zhenda Xie and Zhengyan Zhang and Zhewen Hao and Zhibin Gou and Zhicheng Ma and Zhigang Yan and Zhihong Shao and Zhipeng Xu and Zhiyu Wu and Zhongyu Zhang and Zhuoshu Li and Zihui Gu and Zijia Zhu and Zijun Liu and Zilin Li and Ziwei Xie and Ziyang Song and Ziyi Gao and Zizheng Pan},
      year={2025},
      eprint={2412.19437},
      archivePrefix={arXiv},
      primaryClass={cs.CL},
      url={https://arxiv.org/abs/2412.19437}, 
}

@misc{zhang2025llavavideovideoinstructiontuning,
      title={LLaVA-Video: Video Instruction Tuning With Synthetic Data}, 
      author={Yuanhan Zhang and Jinming Wu and Wei Li and Bo Li and Zejun Ma and Ziwei Liu and Chunyuan Li},
      year={2025},
      eprint={2410.02713},
      archivePrefix={arXiv},
      primaryClass={cs.CV},
      url={https://arxiv.org/abs/2410.02713}, 
}

@misc{ru2026generationknowsunlockingfinegrained,
      title={What Generation Knows: Unlocking Fine-Grained Understanding in Unified Multimodal Models}, 
      author={Tengju Ru and Weitong Lian and Zecong Tang and Zhiheng Liu and Lingyi Meng and Haoran Li and Kaixuan Wang and Bohao Li and Xihui Liu},
      year={2026},
      eprint={2610.23987},
      archivePrefix={arXiv},
      primaryClass={cs.CV},
      url={https://arxiv.org/abs/2610.23987}, 
}

@misc{meng2026imugbenchbenchmarkingunifiedmultimodal,
      title={IMUG-Bench: Benchmarking Unified Multimodal Models on Interleaved Understanding and Generation}, 
      author={Lingyi Meng and Zecong Tang and Haoran Li and Tengju Ru and Zhejun Cui and Weitong Lian and Qi Kang and Hangshuo Cao and Yichen Zhu and Yechi Liu and Kaixuan Wang and Yu-Jie Yuan and Chunwei Wang and Yu Zhang and Bo Dai},
      year={2026},
      eprint={2606.09169},
      archivePrefix={arXiv},
      primaryClass={cs.AI},
      url={https://arxiv.org/abs/2606.09169}, 
}

@misc{zhang2026pelicanunify10unifiedembodied,
      title={Pelican-Unify 1.0: A Unified Embodied Intelligence Model for Understanding, Reasoning, Imagination and Action}, 
      author={Yi Zhang and Yinda Chen and Che Liu and Zeyuan Ding and Jin Xu and Shilong Zou and Junwei Liao and Jiayu Hu and Xiancong Ren and Xiaopeng Zhang and Yechi Liu and Haoyuan Shi and Zecong Tang and Haosong Sun and Renwen Cui and Kuishu Wu and Wenhai Liu and Yang Xu and Yingji Zhang and Yidong Wang and Senkang Hu and Jinpeng Lu and Nga Teng Chan and Yechen Wu and Zeting Liu and Xianzhou Hou and Yong Dai and Jian Tang and Xiaozhu Ju},
      year={2026},
      eprint={2605.15153},
      archivePrefix={arXiv},
      primaryClass={cs.RO},
      url={https://arxiv.org/abs/2605.15153}, 
}

@misc{chen2024rightwayevaluatinglarge,
      title={Are We on the Right Way for Evaluating Large Vision-Language Models?}, 
      author={Lin Chen and Jinsong Li and Xiaoyi Dong and Pan Zhang and Yuhang Zang and Zehui Chen and Haodong Duan and Jiaqi Wang and Yu Qiao and Dahua Lin and Feng Zhao},
      year={2024},
      eprint={2403.20330},
      archivePrefix={arXiv},
      primaryClass={cs.CV},
      url={https://arxiv.org/abs/2403.20330}, 
}

@misc{yang2025thinkingspacemultimodallarge,
      title={Thinking in Space: How Multimodal Large Language Models See, Remember, and Recall Spaces}, 
      author={Jihan Yang and Shusheng Yang and Anjali W. Gupta and Rilyn Han and Li Fei-Fei and Saining Xie},
      year={2025},
      eprint={2412.14171},
      archivePrefix={arXiv},
      primaryClass={cs.CV},
      url={https://arxiv.org/abs/2412.14171}, 
}

\appendix
\section*{Appendix}

\section{VidUEU-Agent Pipeline}
\label{app:VidUEU_agent_pipeline}

\subsection{Stage 1: Global Perception}
\label{app:stage1_global_perception}
The prompts for shot-level summarization, chapter construction, chapter-level summarization, video-level summarization, and Gate A shot admission are shown in Fig.~\ref{fig:shot_profiling_prompt}, Fig.~\ref{fig:stage1_summary_prompts}, and Fig.~\ref{fig:shot_admission_prompt}.

\begin{figure}[!htbp]
\centering
\begin{tcolorbox}[
    width=0.98\linewidth,
    colback=white,
    colframe=black,
    boxrule=0.4pt,
    arc=0pt,
    left=7pt,
    right=7pt,
    top=7pt,
    bottom=7pt
]
\footnotesize
\textbf{Shot Profiling Prompt}\par
\medskip\hrule\medskip
{\ttfamily\raggedright\setlength{\parindent}{0pt}\setlength{\parskip}{4pt}%
You are building a visual-only shot profile for a video pipeline.\par
Use only the images.\par
Return strict JSON with keys:\par
vlm\_summary, shot\_role, scene\_type, visual\_quality\_hint, informativeness\_hint, candidate\_sampling\_hint.\par
Allowed hint values for visual\_quality\_hint, informativeness\_hint, and candidate\_sampling\_hint: low, medium, high.\par
}
\medskip\hrule
\end{tcolorbox}
\caption{Prompt for shot-level summarization.}
\label{fig:shot_profiling_prompt}
\end{figure}

\begin{figure}[!htbp]
\centering
\begin{tcolorbox}[
    width=0.98\linewidth,
    colback=white,
    colframe=black,
    boxrule=0.4pt,
    arc=0pt,
    left=7pt,
    right=7pt,
    top=7pt,
    bottom=7pt
]
\footnotesize
\textbf{Chapter Construction and Summary Prompts}\par
\medskip\hrule\medskip
{\ttfamily\raggedright\setlength{\parindent}{0pt}\setlength{\parskip}{3pt}%
[Chapter Construction Prompt / Pairwise Merge Decision Prompt]\par
You are a strict JSON decision engine for merging two adjacent video shots into chapters.\par
Decide whether the two shots belong to the same chapter.\par
Return exactly one valid JSON object and nothing else. Do not use markdown fences.\par
The JSON schema is:\par
\{\par
  "same\_chapter": true,\par
  "confidence": 0.0,\par
  "hard\_boundary": false,\par
  "reason\_type": "visual\_continuation",\par
  "notes": "short reason"\par
\}\par
Rules: same\_chapter and hard\_boundary must be JSON booleans; confidence must be in [0.0, 1.0]; reason\_type must be a short snake\_case string; notes must be under 20 words; if evidence is mixed or uncertain, still return valid JSON with lower confidence; base the decision only on the provided shot summaries and hints.\par
LEFT role: \textless{}left\_shot.shot\_role\textgreater{}\par
LEFT vlm: \textless{}left\_shot.vlm\_summary\textgreater{}\par
LEFT scene\_type: \textless{}left\_shot.scene\_type\textgreater{}\par
LEFT informativeness\_hint: \textless{}left\_shot.informativeness\_hint\textgreater{}\par
RIGHT role: \textless{}right\_shot.shot\_role\textgreater{}\par
RIGHT vlm: \textless{}right\_shot.vlm\_summary\textgreater{}\par
RIGHT scene\_type: \textless{}right\_shot.scene\_type\textgreater{}\par
RIGHT informativeness\_hint: \textless{}right\_shot.informativeness\_hint\textgreater{}\par
\vspace{0.35em}
[Chapter Summary Prompt]\par
You are a robust chapter summarizer for a structured video understanding pipeline.\par
Read all provided shot summaries for this chapter and merge them into one concise, fluent, informative English paragraph. Summarize the chapter as a coherent local segment, not disconnected bullet points. Preserve important subjects, actions, stages, transitions, and outcomes; remove redundancy; reflect progression when present; stay grounded in the shot summaries only.\par
Writing requirements: write plain English only; output exactly one short paragraph; do not use bullet points, numbering, JSON, markdown, labels, or quotation marks; keep the summary compact, specific, and easy for downstream planning, scoring, and visualization.\par
Return the chapter summary only.\par
\vspace{0.35em}
[Video-level Summary Prompt]\par
You are a robust video summarizer for a structured video understanding pipeline.\par
Read all provided chapter summaries and merge them into one concise, fluent, informative English paragraph. Summarize the global content of the whole video rather than isolated chapter fragments. Capture the overall subject, major stages, important transitions, and final outcomes when supported; remove redundancy; reflect global progression when present; stay grounded in the chapter summaries only.\par
Writing requirements: write plain English only; output exactly one short paragraph; do not use bullet points, numbering, JSON, markdown, labels, or quotation marks; keep the summary compact, globally informative, and easy for downstream planning, scoring, and visualization.\par
Return the video summary only.\par
}
\medskip\hrule
\end{tcolorbox}
\caption{Prompt for chapter construction, chapter-level summarization, and video-level summarization.}
\label{fig:stage1_summary_prompts}
\end{figure}

\begin{figure}[!htbp]
\centering
\begin{tcolorbox}[
    width=0.98\linewidth,
    colback=white,
    colframe=black,
    boxrule=0.4pt,
    arc=0pt,
    left=7pt,
    right=7pt,
    top=7pt,
    bottom=7pt
]
\footnotesize
\textbf{Shot Admission Prompt}\par
\medskip\hrule\medskip
{\ttfamily\raggedright\setlength{\parindent}{0pt}\setlength{\parskip}{4pt}%
You are a strict shot-admission judge for a video frame selection pipeline.\par
Judge whether the current shot is worth sending to downstream frame sampling.\par
Keep shots only if they are visually usable and likely to yield informative frames.\par
Reject shots that are mainly low-information filler, weak transitions, visually poor, or not worth spending frame-sampling budget on.\par
Return strict JSON with keys usable\_score, quality\_score, informativeness\_score, decision, reason, evidence.\par
Video summary: \textless{}video\_profile.video\_summary\textgreater{}\par
Video type: \textless{}video\_profile.video\_type\textgreater{}\par
Chapter role: \textless{}getattr(chapter, 'chapter\_role', '')\textgreater{}\par
Chapter summary: \textless{}getattr(chapter, 'chapter\_summary', '')\textgreater{}\par
Shot summary: \textless{}shot.vlm\_summary\textgreater{}\par
Shot role: \textless{}shot.shot\_role\textgreater{}\par
Scene type: \textless{}shot.scene\_type\textgreater{}\par
Visual quality hint: \textless{}shot.visual\_quality\_hint\textgreater{}\par
Informativeness hint: \textless{}shot.informativeness\_hint\textgreater{}\par
Sampling hint: \textless{}shot.candidate\_sampling\_hint\textgreater{}\par
Representative frame count: \textless{}len(shot.rep\_frames or [])\textgreater{}\par
}
\medskip\hrule
\end{tcolorbox}
\caption{Prompt for Gate A shot admission.}
\label{fig:shot_admission_prompt}
\end{figure}

\begin{figure}[!htbp]
\centering
\begin{tcolorbox}[
    width=0.98\linewidth,
    colback=white,
    colframe=black,
    boxrule=0.4pt,
    arc=0pt,
    left=7pt,
    right=7pt,
    top=7pt,
    bottom=7pt
]
\footnotesize
\textbf{Clip Summary Generation Prompt}\par
\medskip\hrule\medskip
{\ttfamily\raggedright\setlength{\parindent}{0pt}\setlength{\parskip}{4pt}%
You are writing a concise \textless{}level\textgreater{} summary for a video segment.\par
Summarize ONLY from the provided \textless{}level\textgreater{} evidence.\par
Keep it factual, concrete, and short.\par
Return 1-3 sentences only.\par
\vspace{0.35em}
\textless{}level\_title\textgreater{} evidence:\par
\textless{}evidence\_lines\textgreater{}\par
}
\medskip\hrule
\end{tcolorbox}
\caption{Prompt for clip summary generation.}
\label{fig:stage3_summary_prompt}
\end{figure}

\subsection{Stage 2: Keyframe Mining}
\label{app:stage2_keyframe_mining}
The implementation prompts and rules for hierarchical tool planning, tool signal verbalization, frame value scoring, and quota allocation with weighted facility-location selection are shown in Fig.~\ref{fig:tool_planning_prompts}, Fig.~\ref{fig:tool_signal_semanticizer_prompt}, Fig.~\ref{fig:value_scoring_prompt}, and Fig.~\ref{fig:quota_planning_prompt}.

\begin{figure}[!htbp]
\centering
\begin{tcolorbox}[
    width=0.98\linewidth,
    colback=white,
    colframe=black,
    boxrule=0.4pt,
    arc=0pt,
    left=7pt,
    right=7pt,
    top=7pt,
    bottom=7pt
]
\footnotesize
\textbf{Tool Planning Prompts}\par
\medskip\hrule\medskip
{\ttfamily\raggedright\setlength{\parindent}{0pt}\setlength{\parskip}{4pt}%
[Tool Planning Prompt (chapter-level)]\par
You are the chapter-level tool planner for a structured video keyframe mining pipeline.\par
\vspace{0.35em}
A chapter is a contiguous semantic segment ... usually covering one local topic, stage, or scene block.\par
\vspace{0.35em}
Your job is NOT to choose frames. Your job is to choose the 2 most important tools for each chapter ... so the pipeline can understand that chapter efficiently ...\par
\vspace{0.35em}
Planning principles:\par
- Choose the smallest sufficient tool set for each chapter.\par
- Reflect what needed to understand the whole chapter broadly, not every possible detail.\par
- Use the video summary as global context, but judge each chapter independently.\par
- This stage does not decide score\_focus.\par
- Reasons must be short, concrete, and grounded in the provided summaries.\par
\vspace{0.35em}
Output requirements:\par
- Return valid JSON only.\par
- Use English only.\par
\vspace{0.35em}
[Tool Planning Prompt (shot-level)]\par
You are the shot-level tool planner for a structured video keyframe mining pipeline.\par
\vspace{0.35em}
The video has been decomposed into chapters and shots. Later stages will choose keyframes ...\par
\vspace{0.35em}
Your job is NOT to choose frames. Your job is to choose the final tool set for each shot so the pipeline can understand that specific shot ...\par
\vspace{0.35em}
You will be given:\par
- the current chapter summary\par
- the chapter-level recommended tools\par
- all shot summaries inside that chapter\par
\vspace{0.35em}
Planning principles:\par
- Evaluate each shot independently based on its own summary.\par
- Treat chapter-level tools as recommendations, not fixed final output.\par
- Prefer adding at least 1 extra suitable tool when the shot content supports it; skip only when none is suitable.\par
- Remove a chapter-level tool only when it is unsuitable for this shot.\par
- Add only the minimum necessary extra tools from the provided extra candidate list.\par
- Keep the final tool set small but sufficient.\par
- final tool\_actions must contain the complete final tool set for the shot, with no duplicates.\par
- score\_focus means the 1 or 2 most important tools inside the selected set; it must be a subset of selected tools.\par
- Reasons must be short, concrete, and grounded in the provided summaries.\par
\vspace{0.35em}
Output requirements:\par
- Return valid JSON only.\par
- Use English only.\par
}
\medskip\hrule
\end{tcolorbox}
\caption{Prompt for hierarchical tool planning.}
\label{fig:tool_planning_prompts}
\end{figure}

\begin{figure}[!htbp]
\centering
\begin{tcolorbox}[
    width=0.98\linewidth,
    colback=white,
    colframe=black,
    boxrule=0.4pt,
    arc=0pt,
    left=7pt,
    right=7pt,
    top=7pt,
    bottom=7pt
]
\footnotesize
\textbf{Tool Signal Semanticizer Prompt}\par
\medskip\hrule\medskip
{\ttfamily\raggedright\setlength{\parindent}{0pt}\setlength{\parskip}{4pt}%
You are a frame-level tool-signal semanticizer for structured video keyframe mining.\par
\vspace{0.35em}
You will receive multiple candidate frames from the same shot, the tools actually executed for this shot, and each active tool's purpose.\par
\vspace{0.35em}
Your task is NOT to score frames.\par
Your task is to convert raw tool evidence, values, scores, and tool-specific hints into frame-specific semantic descriptions that are faithful to the actual tool outputs and easy to understand.\par
\vspace{0.35em}
Core requirements:\par
- Be strictly evidence-grounded.\par
- Explain the meaning of tool outputs rather than merely copying raw snippets.\par
- Use tool purpose as context, but base every description on actual per-frame evidence.\par
- Explain numeric values, categorical levels, or structured fields in plain language when present.\par
- Preserve supported directionality, such as stronger vs weaker motion or higher vs lower confidence.\par
- Do not invent thresholds, hidden formulas, unseen objects, actions, relations, or text.\par
- Do not confuse detector confidence, motion magnitude, or tool values with semantic importance.\par
- Use only tools that were actually executed and provided in the input.\par
- For tools not executed or without useful frame-specific evidence, return an empty string for that field.\par
- semantic\_text must be a short merged frame-specific description from useful active-tool evidence.\par
- Do not score, rank, or compare quality.\par
- Output JSON only.\par
- Use English only.\par
}
\medskip\hrule
\end{tcolorbox}
\caption{Prompt for tool signal verbalization.}
\label{fig:tool_signal_semanticizer_prompt}
\end{figure}

\begin{figure}[!htbp]
\centering
\begin{tcolorbox}[
    width=0.98\linewidth,
    colback=white,
    colframe=black,
    boxrule=0.4pt,
    arc=0pt,
    left=7pt,
    right=7pt,
    top=7pt,
    bottom=7pt
]
\scriptsize
\textbf{Value Scoring Prompt}\par
\medskip\hrule\medskip
{\ttfamily\raggedright\setlength{\parindent}{0pt}\setlength{\parskip}{4pt}%
You are a frame value scorer for structured video keyframe mining.\par
\vspace{0.35em}
You will receive multiple candidate frames from the same shot; video, chapter, and shot summaries; the tools selected for this shot and their purposes; score\_focus, which names the 1 or 2 relatively more important selected tools; frame-level semantic descriptions rewritten from raw tool outputs; and an extra factual caption for each frame.\par
\vspace{0.35em}
Your task is to assign one ABSOLUTE value score in the range [0, 1] to each frame.\par
\vspace{0.35em}
How to reason:\par
- Understand the video, chapter, and shot from the summaries.\par
- Inspect each frame's semantic\_text, tool-specific descriptions, and factual caption.\par
- Judge what concrete information each source contributes, how clear and specific it is, and how useful it is for understanding the shot.\par
- Treat tool-derived signals as auxiliary evidence; keep the frame caption as a central focus.\par
- Integrate all evidence to decide the final score.\par
\vspace{0.35em}
How to use score\_focus:\par
- score\_focus is not a score dimension list.\par
- When meaningful evidence from a score\_focus tool is present, give it extra attention.\par
- Do not ignore useful evidence from other selected tools.\par
\vspace{0.35em}
Scoring principles:\par
- Score each frame on its own absolute usefulness as a final keyframe for this shot.\par
- Differentiate frames when they differ in clarity, specificity, or decision value.\par
- Prefer frames that are clear, informative, representative, non-redundant, decision-useful, and easy to interpret.\par
- Reward frames that capture important entities, actions, states, transitions, text, speech-derived meaning, or step-specific evidence relevant to the shot.\par
- Reward frames where caption, semantic\_text, and tool semantics reinforce a clear grounded interpretation.\par
- Penalize frames that are vague, redundant, weakly grounded, semantically thin, generic, repetitive, or less useful despite technical cleanliness.\par
- Do not treat tool presence, detector confidence, motion magnitude, or other tool-side values as value by themselves.\par
\vspace{0.35em}
Scoring meaning:\par
- 0.0-0.1: almost no useful evidence.\par
- 0.1-0.25: very weak; evidence is vague, redundant, or not useful enough.\par
- 0.25-0.45: limited; partially useful but missing clarity or informativeness.\par
- 0.45-0.65: moderate; understandable and useful but not especially strong.\par
- 0.65-0.8: strong; clearly informative, representative, or decision-worthy.\par
- 0.8-0.92: very strong; strong candidate for final keyframe selection.\par
- 0.92-1.0: exceptional; unusually clear, information-dense, and highly useful.\par
\vspace{0.35em}
High-value frames usually clearly show the key subject, state, action, step, transition, result, important text, speech-derived meaning, object evidence, or motion/change evidence; provide non-trivial information beyond generic background or redundant nearby frames; and help justify retaining the frame as a representative.\par
\vspace{0.35em}
Output JSON only.\par
Use English only.\par
}
\medskip\hrule
\end{tcolorbox}
\caption{Prompt for frame value scoring.}
\label{fig:value_scoring_prompt}
\end{figure}

\begin{figure}[!htbp]
\centering
\begin{tcolorbox}[
    width=0.98\linewidth,
    colback=white,
    colframe=black,
    boxrule=0.4pt,
    arc=0pt,
    left=7pt,
    right=7pt,
    top=7pt,
    bottom=7pt
]
\footnotesize
\textbf{Quota Planning Prompt}\par
\medskip\hrule\medskip
{\ttfamily\raggedright\setlength{\parindent}{0pt}\setlength{\parskip}{4pt}%
You are a hierarchical quota planner for structured video keyframe mining.\par
\vspace{0.35em}
You must plan quotas at three levels:\par
- global\par
- chapter\par
- shot\par
\vspace{0.35em}
Goal:\par
Under the allowed total quota range, choose one fixed final global quota and allocate it across shots so later stages can retain the most valuable representative frames. You are not choosing frames directly.\par
\vspace{0.35em}
Critical planning rules:\par
1. Choose one integer global\_budget inside the allowed total quota range.\par
2. Do not default to quota=1 merely because there is only one chapter or shot.\par
3. The sum of all shot target\_quota values must equal global\_budget exactly.\par
4. Use video, chapter, and shot summaries to judge where more quota is needed for distinct coverage, richer content, or stronger representativeness.\par
5. Duration is supporting evidence, not the only reason; longer duration justifies more quota only when summaries suggest broader coverage, multiple stages, transitions, or meaningful states/results.\par
6. Weak, redundant, or semantically flat shots may receive quota 0.\par
7. Prefer coverage of distinct stages, states, results, events, or subtopics over near-duplicates.\par
8. Judge each shot independently while keeping the final allocation globally balanced.\par
\vspace{0.35em}
How to use the input:\par
- Start from the video summary to understand broad content and final selection coverage.\par
- Use chapter summaries to judge which chapters deserve broader coverage.\par
- Reason shot by shot using each shot summary and duration.\par
- Assign more quota when a shot suggests multiple steps, transitions, states, results, or distinct moments.\par
- Assign low or zero quota to weak, repetitive, or semantically flat shots.\par
- Reason fields should explain quotas in terms of summary meaning and coverage need.\par
\vspace{0.35em}
Output JSON only.\par
Use English only.\par
}
\medskip\hrule
\end{tcolorbox}
\caption{Prompt and rule for quota allocation and weighted facility-location selection.}
\label{fig:quota_planning_prompt}
\end{figure}

\subsection{Stage 3: Metadata Construction}
\label{app:stage3_metadata_construction}
The prompts for clip summary generation and pairwise transition generation are shown in Fig.~\ref{fig:stage3_summary_prompt} and Fig.~\ref{fig:pairwise_transition_prompt}.

\begin{figure}[!htbp]
\centering
\begin{tcolorbox}[
    width=0.98\linewidth,
    colback=white,
    colframe=black,
    boxrule=0.4pt,
    arc=0pt,
    left=7pt,
    right=7pt,
    top=7pt,
    bottom=7pt
]
\footnotesize
\textbf{Pairwise Transition Prompt and Normalization Rules}\par
\medskip\hrule\medskip
{\ttfamily\raggedright\setlength{\parindent}{0pt}\setlength{\parskip}{4pt}%
You are analyzing two consecutive keyframes A -\textgreater{} B from a short video segment.\par
Write exactly ONE transition string describing meaningful visible changes from A to B.\par
\vspace{0.35em}
Use images as primary evidence. Captions, video type, global summary, and modality advice are secondary hints; if they conflict with the images, trust the images.\par
\vspace{0.35em}
Describe only visible changes that affect scene understanding, including object appearance/disappearance, position, state, visible text, camera/view change, or shot change. Ignore tiny flicker, compression noise, blur differences, glare, reflections, and exposure shifts unless they affect meaningful content. Do not infer hidden causes, intentions, or off-screen events.\par
\vspace{0.35em}
Normalize changes into atomic edits. Each edit should use one target and one allowed verb:\par
add, remove, move, rotate, open, close, turn on, turn off, change text, zoom, pan, tilt, cut.\par
\vspace{0.35em}
Use image-frame locations such as left side, center-right, upper-left, or bottom center. Avoid object-centric or anatomical left/right.\par
\vspace{0.35em}
Decision rules:\par
- If B is a clear new shot, use cut to new shot.\par
- If the main change is camera/view motion, prefer zoom/pan/tilt over object move.\par
- Use add/remove for appearance changes, change text for visible text changes, and open/close or turn on/off for clear state changes.\par
- For driving or dashcam scenes, do not treat ego-motion or optics artifacts as object edits.\par
- Edits(k) must match the number of edits; if there is no meaningful change, output Edits(0): none.\par
\vspace{0.35em}
Context:\par
- Video type: \textless{}video\_type\textgreater{}\par
- Global summary: \textless{}summary\textgreater{}\par
- Modality advice: \textless{}advice\textgreater{}\par
\vspace{0.35em}
Captions:\par
A: \textless{}cap\_a\textgreater{}\par
B: \textless{}cap\_b\textgreater{}\par
\vspace{0.35em}
OUTPUT FORMAT:\par
Edits(k): 1) ...; 2) ... | Shot: \textless{}none/minor/scene\_cut\textgreater{} | View: \textless{}none/minor/major\textgreater{} | Subject: \textless{}same/unclear/different\textgreater{} | Uncertainty: \textless{}none or short note\textgreater{}\par
\vspace{0.35em}
Return ONLY one line in the exact format above.\par
}
\medskip\hrule
\end{tcolorbox}
\caption{Prompt for pairwise transition generation.}
\label{fig:pairwise_transition_prompt}
\end{figure}

\subsection{Stage 4: Instruction Data Construction}
\label{app:stage4_instruction_data_construction}
The prompts for task routing and instruction generation are shown in Fig.~\ref{fig:task_routing_prompt_appendix} and Fig.~\ref{fig:task_generation_prompts}.
\begin{figure}[!htbp]
\centering
\begin{tcolorbox}[
    width=0.98\linewidth,
    colback=white,
    colframe=black,
    boxrule=0.4pt,
    arc=0pt,
    left=7pt,
    right=7pt,
    top=7pt,
    bottom=7pt
]
\scriptsize
\textbf{Task Routing Prompt}\par
\medskip\hrule\medskip
{\ttfamily\raggedright\setlength{\parindent}{0pt}\setlength{\parskip}{3pt}%
You are a quota allocator for dataset instance generation.\par
You MUST base your decision ONLY on the EVIDENCE PACK lines below.\par
Each evidence line begins with an evidence\_id in brackets, e.g. [S\_SEGMENT\_SUM], [F00\_CAP], [P00].\par
\vspace{0.25em}
GOAL (highest priority: maximize VALID OUTPUT COUNT): assign quotas within the limits for THIS segment; prefer producing MORE dialog instances when feasible ...; if allocating edit quota, choose suitable frame pairs ONLY from the provided candidate pair list.\par
\vspace{0.25em}
\textless{}global\_mix\_note\textgreater{}\par
IMPORTANT: GLOBAL\_MIX is SOFT guidance. Do NOT set dialog quotas to 0 just to satisfy a mix ratio. If unsure, choose MORE dialog rather than fewer.\par
\vspace{0.25em}
EDIT ALLOCATION POLICY (do NOT be overly conservative): candidate edit pairs are already pre-screened; do NOT re-apply stricter hidden rules. If there is at least 1 candidate pair and no explicit evidence saying edits are unsuitable, allocate a NON-ZERO edit quota. When uncertain, prefer a SMALL non-zero edit quota ...; allocate edit ONLY from the provided candidate pair list.\par
\vspace{0.25em}
Practical heuristic: if candidate\_pairs\_count \textgreater{}= 4, usually allocate edit 3-4 unless hard evidence says unsuitable; if candidate\_pairs\_count is 1-3, usually allocate edit 2-3; if candidate\_pairs\_count == 0, set edit = 0.\par
\vspace{0.25em}
CRITICAL INDEXING: this segment has N keyframes indexed by POSITION 0..N-1. All frame and pair indices in your output refer to these keyframe POSITIONS.\par
\vspace{0.25em}
EVIDENCE USAGE POLICY:\par
1) ALWAYS read and use segment-level background evidence FIRST: [S\_SEGMENT\_SUM].\par
2) Then use frame-level evidence [Fxx\_*] and pair/change evidence [Pxx]/[Cxx] for specific instances.\par
3) Every type with quota \textgreater{} 0 MUST cite at least ONE segment-level evidence id (S\_*) in its rationale.\par
4) Do NOT treat any tool signal as ground truth; treat all evidence as imperfect.\par
\vspace{0.25em}
EDIT PAIR QUALITY: edit generation requires small, controllable edits. ONLY choose pairs from candidate pairs; prefer reviewed pairs [Pxx] when available; if no candidate pair satisfies small-edit constraints, set edit quota to 0.\par
\vspace{0.25em}
DIALOG TYPES: dialogs must remain grounded in evidence and not invent details.\par
TYPE-SPECIFIC GUIDELINES: edit only when a small-edit pair exists; unify only when reviewed edit pairs exist.\par
\vspace{0.25em}
INPUTS:\par
Candidate instruction task types: \textless{}candidate\_types\_json\textgreater{}\par
Quota limits: \textless{}quota\_lines\textgreater{}\par
Edit candidate pairs: \textless{}mb\_pairs\_txt\textgreater{}\par
Reviewed edit pairs: \textless{}reviewed\_txt\textgreater{}\par
EVIDENCE PACK: \textless{}evidence\_pack\_text\textgreater{}\par
\vspace{0.25em}
OUTPUT FORMAT (strict JSON only, no extra text):\par
\{\par
"quotas": \{\par
"fine\_grained\_perception": 0,\par
"visual\_reasoning": 0,\par
"multi\_image\_logic": 0,\par
"editing": 1,\par
"grounded\_edit\_to\_select": 0,\par
"interleaved\_recall\_generate\_select": 0,\par
"stepwise\_generate\_select": 0,\par
"middle\_step\_gap\_fill": 0,\par
"instructional\_generate\_select": 0\par
\},\par
"edit\_pairs": [\{"from": 0, "to": 1\}],\par
"rationales": [\{"type": "editing", "evidence\_refs": ["S\_SEGMENT\_SUM", "P00"], "reason": "Grounded edit pair."\}]\par
\}\par
\vspace{0.25em}
RATIONALES REQUIREMENTS: provide one rationale entry per type with quota \textgreater{} 0; each rationale MUST include at least one segment-level evidence id (S\_*); for dialog rationales, cite at least one frame-level evidence id (Fxx\_*).\par
}
\medskip\hrule
\end{tcolorbox}
\caption{Prompt for task routing.}
\label{fig:task_routing_prompt_appendix}
\end{figure}

\begin{figure}[!htbp]
\centering
\begin{tcolorbox}[
    width=0.98\linewidth,
    colback=white,
    colframe=black,
    boxrule=0.4pt,
    arc=0pt,
    left=7pt,
    right=7pt,
    top=7pt,
    bottom=7pt
]
\scriptsize
\textbf{Task Generation Prompts and Rules}\par
\medskip\hrule\medskip
{\ttfamily\raggedright\setlength{\parindent}{0pt}\setlength{\parskip}{3pt}%
[Understanding Generation]\par
Generate instruction-tuning instances from an EVIDENCE PACK, where each line starts with an evidence\_id such as [F00\_CAP], [S\_SEGMENT\_SUM], or [P00].\par
\vspace{0.25em}
\textless{}fewshot\_block\textgreater{}\par
Rules: cite evidence\_refs for every item; use keyframe positions 0..N-1; do not invent unsupported details; output concise JSON only.\par
SEGMENT META: \textless{}seg\_meta\_json\textgreater{}\par
DEFAULT IMAGE: image\_pos=\textless{}default\_pos\textgreater{}, image\_path=\textless{}default\_img\textgreater{}\par
\textless{}body\textgreater{}\par
EVIDENCE PACK: \textless{}evidence\_pack\_text\textgreater{}\par
Target tasktype: \textless{}tasktype\textgreater{}; target item count: \textless{}quota\textgreater{}\par
Output JSON schema: \textless{}schema\textgreater{}\par
\vspace{0.25em}
[Single-Image and Multiple-Choice Rules]\par
Single-image questions must be answerable from image\_pos only, without process or timeline wording. Segment evidence may be used only as light background; evidence\_refs must include frame-level evidence for the same image\_pos. Multiple-choice items must have exactly one correct option and three plausible same-type distractors. Avoid fragile exact counting; use "from the camera viewpoint" for left/right relations.\par
\vspace{0.25em}
[Dialog Generation and Repair]\par
\textless{}kind\_rules\textgreater{}\par
\textless{}base\_rules\textgreater{}\par
\textless{}image\_block\textgreater{}\par
EVIDENCE PACK: \textless{}dlg\_ev\textgreater{}\par
Generate \textless{}quota\textgreater{} dialog(s) of type: \textless{}kind\textgreater{}.\par
Output JSON schema: \textless{}schema\textgreater{}\par
Repair rule: output exactly \textless{}want\textgreater{} turns, alternate user/assistant from user, cite evidence\_refs in every turn, include frame-level evidence per turn, and avoid before/after/earlier/later/next/previous.\par
CURRENT TURNS: \textless{}turns\_json\textgreater{}\par
\vspace{0.25em}
[Edit Pair Review and Instruction]\par
Validate consecutive keyframe pairs for localized instruction-based image editing. Accept only small, controllable, same-shot changes with at most 1-2 visible semantic edits. Reject scene cuts, major viewpoint changes, camera-motion-only changes, and unsupported target content. The instruction must cover the true source-target change, add no extra changes, and end with: "Do not change anything else."\par
Segment summary: \textless{}segment\_summary\_short\textgreater{}\par
INPUT PAIRS: \textless{}input\_pairs\_json\textgreater{}\par
Output JSON fields: pairs[from, to, ok, edit\_count, edits, instruction, reason]\par
\vspace{0.25em}
[Unified Task Generation]\par
\textless{}system\_style\textgreater{}\par
Generate a grounded multi-step vision-language task from structured evidence. Use only the provided evidence, keep fields concise, and keep explanations consistent with evidence\_refs.\par
TASK TYPE: \textless{}task\_type\textgreater{}\par
TASK PAYLOAD: \textless{}task\_payload\_json\textgreater{}\par
EVIDENCE PACK: \textless{}evidence\_pack\_text\textgreater{}\par
Style: write as a natural multimodal instruction sample; make select.question choose a candidate frame/image; explain why the target is uniquely supported; avoid speculation; use select.answer format target:FXX.\par
Return JSON schema: \textless{}schema\_json\textgreater{}\par
}
\medskip\hrule
\end{tcolorbox}
\caption{Prompt for instruction generation.}
\label{fig:task_generation_prompts}
\end{figure}

\FloatBarrier

\subsection{LLM--VLM Design and Model Selection}
\label{app:llm_vlm_comparison}

\noindent\textbf{LLM--VLM Design Comparison.}
We compare the metadata-driven LLM design with a direct VLM alternative for filtering, task routing, and instruction generation. Both settings use the same input data, prompt templates, generation parameters, and GPT-4o evaluation protocol, each producing 200 final samples. As shown in Table~\ref{tab:llm_vlm_comparison}, the LLM design outperforms the VLM alternative in both efficiency (12.4 vs. 5.6 samples/min) and quality (4.18 vs. 2.72), supporting the effectiveness of metadata-driven routing and instruction synthesis.

\begin{table}[!htbp]
\centering
\small
\caption{Comparison between the metadata-driven LLM design and the direct VLM alternative.}
\label{tab:llm_vlm_comparison}
\begin{tabular}{lcc}
\toprule
Setting & Final Samples/min $\uparrow$ & Quality $\uparrow$ \\
\midrule
Original LLM setting & \textbf{12.4} & \textbf{4.18} \\
Direct VLM setting & 5.6 & 2.72 \\
\bottomrule
\end{tabular}
\end{table}

\noindent\textbf{VLM Model Selection.}
To examine the efficiency--quality trade-off in VLM selection, we compare Qwen3.5-9B and Qwen3.5-27B under the same agent workflow, input data, prompt templates, and generation parameters. Both models are evaluated on a single H20 GPU using their respective maximum stable concurrency settings. Each setting generates 200 samples, with quality assessed by GPT-4o on a five-point scale. As shown in Table~\ref{tab:vlm_model_selection}, increasing the model size from 9B to 27B yields only a marginal quality improvement (4.16 to 4.21), while reducing throughput from 12.4 to 4.1 samples/min/GPU. Therefore, we select Qwen3.5-9B as the VLM in our pipeline for its favorable balance between generation quality and computational efficiency.

\begin{table}[!htbp]
\centering
\small
\caption{Efficiency and quality comparison of VLM candidates under the same agent workflow.}
\label{tab:vlm_model_selection}
\begin{tabular}{lcc}
\toprule
VLM & Samples/min/GPU $\uparrow$ & GPT-4o Quality $\uparrow$ \\
\midrule
Qwen3.5-9B & \textbf{12.4} & 4.16 \\
Qwen3.5-27B & 4.1 & \textbf{4.21} \\
\bottomrule
\end{tabular}
\end{table}
\section{Tool Invocation Framework}
\label{app:tool_invocation_system}

\subsection{Tool Pool and Output Types}
\label{app:tool_registry}

Tool invocation supplements frame-level semantic evidence when visual captions alone are insufficient.
The current tool pool includes PaddleOCR~\cite{cui2025paddleocr30technicalreport}, Whisper ASR~\cite{radford2022whisper}, Ultralytics YOLOv8~\cite{jocher_2024_10983461}, RAFT optical flow~\cite{10.1007/978-3-030-58536-5_24}, and OmniParser~\cite{lu2024omniparserpurevisionbased}.
Table~\ref{tab:auxiliary_tool_list} summarizes the primary role, applicable scenarios, and retained output of each tool.

\begin{table*}[!htbp]
\centering
\caption{Summary of auxiliary perception tools used for frame-level and shot-level information extraction.}
\label{tab:auxiliary_tool_list}
\small
\setlength{\tabcolsep}{3.5pt}
\renewcommand{\arraystretch}{1.45}
\begin{tabularx}{\textwidth}{
>{\raggedright\arraybackslash}p{0.18\textwidth}
>{\raggedright\arraybackslash}p{0.28\textwidth}
>{\raggedright\arraybackslash}X}
\toprule
\textbf{Tool} & \textbf{Role} & \textbf{Applicable scenarios} \\
\midrule

\textbf{PaddleOCR}
&
Scene-text extraction from video frames
&
Used to read titles, labels, menu items, tables, subtitles, document text, and screen text, especially in slides, screen recordings, documents, and software demonstrations
\\

\textbf{Whisper ASR}
&
Speech-content recovery around a shot
&
Used when speech provides semantic cues near candidate frames, including narration, spoken instructions, or dialogue context
\\

\textbf{YOLOv8}
&
Object and entity detection
&
Used to identify people, objects, tools, and scene entities that provide visual grounding for frame understanding
\\

\textbf{RAFT}
&
Motion estimation between adjacent candidate frames
&
Used when temporal dynamics or motion strength is informative for candidate-frame analysis
\\

\textbf{OmniParser}
&
Structured GUI understanding
&
Used for GUI screenshots and software-interface frames where interface layout and interactive regions are important
\\

\bottomrule
    \end{tabularx}
\end{table*}

\subsection{Tool Evidence Verbalization and Scoring Rules}
\label{app:tool_evidence_interpretation_rules}
Table~\ref{tab:tool_evidence_interpretation_rules} summarizes the rules used in stage 2 to verbalize tool signals and score each frame.
\begin{table*}[!htbp]
\centering
\caption{Rules used to verbalize tool signals and score each frame.}
\label{tab:tool_evidence_interpretation_rules}
\small
\setlength{\tabcolsep}{3.0pt}
\renewcommand{\arraystretch}{1.45}
\begin{tabularx}{\textwidth}{
>{\raggedright\arraybackslash}p{0.13\textwidth}
>{\raggedright\arraybackslash}p{0.30\textwidth}
>{\raggedright\arraybackslash}p{0.32\textwidth}
>{\raggedright\arraybackslash}X}
\toprule
\textbf{Tool evidence} & \textbf{Verbalization rule} & \textbf{Value-scoring use} & \textbf{Reliability boundary} \\
\midrule

\textbf{OCR}
&
Use recognized textual content as primary evidence for readable text; describe the visible meaning, function, or role of the text rather than enumerating all detected tokens.
&
Reward readable and specific text evidence that reveals task-relevant titles, steps, labels, states, commands, or interface cues; assign higher value when text evidence substantially improves understanding beyond generic visual appearance.
&
Keep noisy, incomplete, or ambiguous text evidence conservative; do not reward repeated boilerplate, recognition artifacts, or isolated fragments unless they reveal important state or intent.
\\

\textbf{ASR}
&
Use nearby spoken content as temporal contextual evidence; treat speech as information associated with the frame interval, rather than as on-screen text or direct visual evidence.
&
Reward specific and temporally relevant speech that contributes concrete instructions, explanations, actions, or contextual meaning; assign higher value when spoken content matches the apparent step, state, or action around the frame.
&
Stay conservative when the alignment between speech and frame is uncertain; do not over-reward vague transcripts, weakly aligned speech, or information already evident from the visual content alone.
\\

\textbf{YOLOv8}
&
Use detected object categories as primary evidence for visible entities; use confidence or location only as supporting information, and describe only entities that are visually supported.
&
Reward clear and relevant visible entities that explain the subject, objects in use, workspace context, or interaction target; assign higher value when detected entities are specific, diverse, and informative.
&
Do not treat detection confidence, quantity, or coverage alone as evidence of value, and do not infer unsupported relationships, intentions, or actions.
\\

\textbf{RAFT}
&
Use optical-flow evidence as primary support for motion or temporal change; interpret motion strength as an indicator of change intensity rather than semantic importance or action type.
&
Reward motion evidence when it marks a meaningful transition, interaction moment, or state change; assign higher value when motion helps distinguish the frame from static or redundant neighboring frames.
&
Describe staticness or motion-heavy appearance only when supported by motion evidence; do not reward motion magnitude alone without semantic relevance.
\\

\textbf{OmniParser}
&
Use parsed interface structure and element descriptions as primary evidence for GUI layout, component meaning, and visible application state; prioritize actionable or state-revealing elements over low-level visual labels.
&
Reward actionable interface structure such as buttons, menus, form fields, toggles, tabs, or clear application states; assign higher value when parsed elements materially improve understanding of what the interface shows or allows.
&
Treat element counts, positions, and bounding regions as supporting evidence rather than meaning by themselves; do not reward dense but uninformative interface clutter.
\\

\bottomrule
\end{tabularx}
\end{table*}

\section{Metadata Schema Definition}
\label{app:metadata_schema_definition}
A minimal metadata example is shown in Fig.~\ref{fig:minimal_metadata_schema_appendix}.
The \texttt{segment} field comes from Stage 1 and contains the video summary, start and end timestamps, segment summary, and a coarse label.
The \texttt{chapters} field comes from Stage 1 and contains ordered shots, their time ranges, chapter summaries, and labels indicating each chapter's role in the video.
The \texttt{shots} field comes from Stage 1 and contains shot-level time ranges, shot summaries, and hints about image quality and informativeness.
The \texttt{keyframes} field comes from Stage 2 and contains selected representative frames with their timestamps, captions, quality judgments, tool-produced semantic signals, and value scores.
The \texttt{transitions} field comes from Stage 3 and contains ordered frame pairs, whether they remain within the same shot, whether the main subject is consistent, the concrete transition between the frames, and a reliability estimate.

\begin{figure}[!htbp]
  \centering
  \setlength{\fboxsep}{7pt}
  \setlength{\fboxrule}{0.4pt}

  \fbox{%
  \begin{minipage}{\dimexpr0.98\linewidth-2\fboxsep-4\fboxrule\relax}
    \footnotesize

    \textbf{Minimal Metadata Schema}\par
    \medskip\hrule\medskip

    {\ttfamily\raggedright\setlength{\parindent}{0pt}\setlength{\parskip}{4pt}%
    \begin{tabular}{@{}p{0.18\linewidth}@{\hspace{0.025\linewidth}}p{0.755\linewidth}@{}}
      \textbf{segment} &
      \texttt{video\_id}, \texttt{start}, \texttt{end}, \texttt{summary}, \texttt{type} \\[3pt]

      \textbf{chapters} &
      \texttt{chapter\_id}, \texttt{time\_span}, \texttt{summary}, \texttt{role} \\[3pt]

      \textbf{shots} &
      \texttt{shot\_id}, \texttt{chapter\_id}, \texttt{time\_span}, \texttt{summary}, \texttt{role}, \texttt{scene\_type}, \texttt{quality\_hint}, \texttt{info\_hint} \\[3pt]

      \textbf{keyframes} &
      \texttt{frame\_id}, \texttt{shot\_id}, \texttt{timestamp}, \texttt{caption}, \texttt{quality}, \texttt{value\_score}, \texttt{tool\_semantics}, \texttt{evidence\_id} \\[3pt]

      \textbf{transitions} &
      \texttt{source}, \texttt{target}, \texttt{atomic\_changes}, \texttt{same\_shot}, \texttt{viewpoint\_change}, \texttt{subject\_consistent}, \texttt{reliability}
    \end{tabular}%
    }

    \medskip\hrule
  \end{minipage}
  }

  \caption{Minimal metadata schema used to bind summaries, keyframes, tool evidence, and transitions.}
  \label{fig:minimal_metadata_schema_appendix}
\end{figure}

\section{Instruction Templates and Examples}
\label{app:instruction_synthesis_templates}

\subsection{Instruction-Data Synthesis Templates}
This subsection gives the detailed templates used to generate task instructions.

\paragraph{Fine-Grained Perception.}
This template produces single-image perception questions, answered as yes/no or multiple choice, to test recognition of visible entities, attributes, positions, text, counts, and local details.

\begin{lstlisting}[style=schema]
{
  "image_pos": <Frame[ID]>,
  "category": "presence|attribute|spatial|text|count",
  "question": "<single-image yes/no or multiple-choice question>",
  "options": ["<A>", "<B>", "<C>", "<D>"],
  "answer": "<Yes|No|A|B|C|D>",
  "evidence_refs": ["F00_CAP"]
}
\end{lstlisting}

\paragraph{Visual Reasoning.}
This template produces single-image visual reasoning questions, answered as open-ended QA, to test inference from concrete visual cues.
The answer gives a concise conclusion followed by concrete visual cues.

\begin{lstlisting}[style=schema]
{
  "image_pos": <Frame[ID]>,
  "category": "reasoning|function|activity|relation",
  "question": "<why/how/what can be inferred from the image>",
  "answer": "<2-4 sentences grounded in visible cues>",
  "evidence_refs": ["F00_CAP"]

}
\end{lstlisting}

\paragraph{Multi-Image Logic.}
This template produces multi-image, multi-turn logic questions, answered through ordered dialogue, to test cross-image comparison, verification, and evidence binding.
It asks the model to compare, connect, or jointly verify facts across images while maintaining strict role order and evidence grounding.

\begin{lstlisting}[style=schema]
{
  "image_pos_list": [<Frame[ID]>, <Frame[ID]>, ...],
  "turns": [
    {
      "role": "user",
      "text": "Goal: <cross-image verification goal>.
       Plan: Step 1: <inspect Image1>. 
       Step 2: <inspect Image2>.
       Step 3: <compare them>.
       Images: Image1:<ImageHere> Image2:<ImageHere>.
       Step 1: <question>",
       "evidence_refs": ["S_SEGMENT_SUM","F00_CAP","F01_CAP","C00"]
    },
    {
      "role": "assistant",
      "text": "<answer grounded in the referenced image>",
      "evidence_refs": ["S_SEGMENT_SUM","F00_CAP","F01_CAP","C00"]
    },
    {
      "role": "user",
      "text": "Step 2: <pairwise comparison question>",
      "evidence_refs": ["F00_CAP", "F01_CAP", "C00"]
    },
    {
      "role": "assistant",
      "text": "<cross-image answer>",
      "evidence_refs": ["S_SEGMENT_SUM","F00_CAP","F01_CAP","C00"]
    }
  ]
}
\end{lstlisting}

\paragraph{Editing}
This template produces image editing instructions from a source--target pair, answered by generating the edited image, to test localized and controllable visual generation.
The change must be small and controllable, preserve stable scene content, and be expressible as one or two atomic edits.

\begin{lstlisting}[style=schema]
{
  "from_pos": <source Frame[ID]>,
  "to_pos": <target Frame[ID]>,
  "ok": true,
  "edit_count": 1|2,
  "edits": ["<atomic local edit>", ...],
  "instruction": "<local edit or infilling instruction>. 
                 Do not change anything else.",
  "reason": "<why the pair is small and controllable>",
  "evidence_refs": ["F00_CAP", "F01_CAP", "P00"]
}
\end{lstlisting}

\paragraph{Grounded Edit-to-Select Generation.}
This template produces a two-image grounded editing task from a source image and a target image.
The answer is a generated target image followed by a single-choice candidate image selection.
It evaluates whether the model can preserve stable visual invariants, identify grounded visible edits, generate the target visual state, and select the candidate image matching the intended edited result.

\begin{lstlisting}[style=schema]
{
  "Round_Plan": {
    "question": {
      "text": "What should be changed from the source image to reach the tar
      -get state?",
      "Frames": ["Frame[SOURCE_ID]", "Frame[TARGET_ID]"]
    },
    "answer": {
      "goal": "<objective target visual state>",
      "invariants": ["<stable entity/background>", ...],
      "edits": ["<grounded visible change>", ...]
    }
  },
  "Round_Gen": {
    "question": {
      "text": "Generate the target image after applying the grounded edit.",
      "Frames": ["Frame[SOURCE_ID]"]
    },
    "answer": {
      "image": "<generated target image>"
    }
  },
  "Round_Select": {
    "question": {
      "text": "Which candidate image best matches the generated target image?",
      "Frames": ["Frame[CANDIDATE_ID]", "Frame[CANDIDATE_ID]", ...]
    },
    "answer": "target:FXX",
    "rationale": "<why the candidate matches the generated target image>"
  },
  "evidence_refs": ["F00_CAP", "F01_CAP", "P00"]
}
\end{lstlisting}

\paragraph{Interleaved Understand--Recall--Generate--Select.}
This template produces a two-image visual understanding and recall task.
The answer contains an open-ended QA response, a generated later-state image, and a single-choice candidate image selection.
It evaluates whether the model can extract visible facts, reason about earlier-to-later changes, generate the recalled later state, and select the candidate frame consistent with that generated state.

\begin{lstlisting}[style=schema]
{
  "Round_Understand": {
    "question": {
      "text": "What visible facts describe the earlier and later image states?",
      "Frames": ["Frame[EARLIER_ID]", "Frame[LATER_ID]"]
    },
    "answer": {
      "facts": ["<visible earlier-state fact>", "<visible later-state fact>"]
    }
  },
  "Round_Plan": {
    "question": {
      "text": "What grounded visual changes connect the earlier state to the later state?",
      "Frames": ["Frame[EARLIER_ID]", "Frame[LATER_ID]"]
    },
    "answer": {
      "edits": ["<grounded earlier-to-later change>", ...]
    }
  },
  "Round_QA": {
    "question": {
      "text": "<open question about the recalled later detail>",
      "Frames": ["Frame[EARLIER_ID]"]
    },
    "answer": "<grounded later-state answer>",
    "rubric": ["<answer constraint>", ...]
  },
  "Round_Gen": {
    "question": {
      "text": "Generate the recalled later-state image.",
      "Frames": ["Frame[EARLIER_ID]"]
    },
    "answer": {
      "image": "<generated later-state image >"
    }
  },
  "Round_Select": {
    "question": {
      "text": "Which candidate frame matches the generated later-state image?",
      "Frames": ["Frame[CANDIDATE_ID]", "Frame[CANDIDATE_ID]", ...]
    },
    "answer": "target:FXX",
    "rationale": "<why the selected candidate matches the generated later state>"
  },
  "evidence_refs": ["F00_CAP", "F01_CAP", "C00"]
}
\end{lstlisting}

\paragraph{Stepwise Sequence Generate-and-Select.}
This template produces a multi-image step reasoning task over a coherent visual sequence.
The answer is a list of generated step-state image followed by single-choice selections for each step.
It evaluates whether the model can decompose a visual process into grounded steps, generate each intermediate state, track temporal order, and select the correct candidate frame for each step.

\begin{lstlisting}[style=schema]
{
  "Round_Overall": {
    "question": {
      "text": "What is the overall coherent visual sequence?",
      "Frames": ["Frame[START_ID]", "Frame[END_ID]"]
    },
    "answer": "<overall description of the visual sequence>"
  },
  "Round_Steps": {
    "question": {
      "text": "What are the grounded step-by-step state changes?",
      "Frames": ["Frame[STEP_0]", "Frame[STEP_1]", "Frame[STEP_2]", ...]
    },
    "answer": [
      "<step 1 grounded transition>",
      "<step 2 grounded transition>"
    ]
  },
  "Round_Gen": [
    {
      "question": {
        "text": "Generate the target image for step 1.",
        "Frames": ["Frame[STEP_0]"]
      },
      "answer": {
        "image": "<generated step-1 image >"
      }
    },
    {
      "question": {
        "text": "Generate the target image for step 2.",
        "Frames": ["Frame[STEP_1]"]
      },
      "answer": {
        "image": "<generated step-2 image >"
      }
    }
  ],
  "Round_Select": [
    {
      "question": {
        "text": "Which candidate frame matches the generated image for step 1?",
        "Frames": ["Frame[CANDIDATE_1]", "Frame[CANDIDATE_2]", ...]
      },
      "answer": "target:FXX",
      "rationale": "<why this frame is the correct step-1 state>"
    },
    {
      "question": {
        "text": "Which candidate frame matches the generated image for step 2?",
        "Frames": ["Frame[CANDIDATE_1]", "Frame[CANDIDATE_2]", ...]
      },
      "answer": "target:FXX",
      "rationale": "<why this frame is the correct step-2 state>"
    }
  ],
  "evidence_refs": ["F00_CAP", "F01_CAP", "F02_CAP", "C00", "C01"]
}
\end{lstlisting}

\paragraph{Middle-Step Gap Generate-and-Fill.}
This template produces a three-image temporal gap-filling task.
The answer is a generated middle-state image followed by a single-choice selection of the real middle candidate frame.
It evaluates whether the model can reason bidirectionally from both adjacent transitions, generate the visual state that fits between two endpoints, and select the candidate satisfying both sides of the sequence.

\begin{lstlisting}[style=schema]
{
  "Round_Explain": {
    "question": {
      "text": "What visual state should fit between the earlier and later frames?",
      "Frames": ["Frame[EARLIER_ID]", "Frame[LATER_ID]"]
    },
    "answer": "<why the middle state fits between earlier and later frames>"
  },
  "Round_Gen": {
    "question": {
      "text": "Generate the middle-step image that fits between the earlier and later frames.",
      "Frames": ["Frame[EARLIER_ID]", "Frame[LATER_ID]"]
    },
    "answer": {
      "image": "<generated middle-state image >"
    }
  },
  "Round_Select": {
    "question": {
      "text": "Which candidate frame matches the generated middle-step image?",
      "Frames": ["Frame[CANDIDATE_1]", "Frame[CANDIDATE_2]", ...]
    },
    "answer": "target:FXX",
    "rationale": "<why the selected frame matches both adjacent transitions>"
  },
  "evidence_refs": ["F00_CAP", "F01_CAP", "F02_CAP", "C00", "C01"]
}
\end{lstlisting}

\paragraph{Instructional Sequential Generate-and-Select.}
This template produces a multi-image future-state instruction generation task from context frames and future target frames.
The answer combines generated edit instructions, generated future-state image, and per-step single-choice candidate selections.
It evaluates whether the model can infer immediate future visual changes from context, generate each future state, define visual checks, and select the matching future frame.

\begin{lstlisting}[style=schema]
{
  "Round_Overall": {
    "question": {
      "text": "What future visual sequence should follow from the context frames?",
      "Frames": ["Frame[CONTEXT_1]", "Frame[CONTEXT_2]"]
    },
    "answer": "<overall future-step instruction from the context>"
  },
  "Round_Edit": {
    "question": {
      "text": "What visible edit should be made at each future step?",
      "Frames": ["Frame[CONTEXT_1]", "Frame[CONTEXT_2]"]
    },
    "answer": ["<step 1 edit instruction>", "<step 2 edit instruction>"]
  },
  "Round_Reason": {
    "question": {
      "text": "Why does each future step follow from the context or previous state?",
      "Frames": ["Frame[CONTEXT_1]", "Frame[CONTEXT_2]"]
    },
    "answer": ["<reason for step 1>", "<reason for step 2>"]
  },
  "Round_Target": {
    "question": {
      "text": "What visual state should each step reach?",
      "Frames": ["Frame[CONTEXT_1]", "Frame[CONTEXT_2]"]
    },
    "answer": ["<desired visual state after step 1>", "<desired visual state after step 2>"]
  },
  "Round_Check": {
    "question": {
      "text": "How can each generated future step be visually verified?",
      "Frames": ["Frame[CONTEXT_1]", "Frame[CONTEXT_2]"]
    },
    "answer": ["<visible condition for step 1>", "<visible condition for step 2>"]
  },
  "Round_Gen": [
    {
      "question": {
        "text": "Generate the target image for future step 1.",
        "Frames": ["Frame[CONTEXT_1]", "Frame[CONTEXT_2]"]
      },
      "answer": {
        "image": "<generated future-step-1 image>"
      }
    },
    {
      "question": {
        "text": "Generate the target image for future step 2.",
        "Frames": ["Frame[CONTEXT_1]", "Frame[CONTEXT_2]", "Frame[FUTURE_1]"]
      },
      "answer": {
        "image": "<generated future-step-2 image>"
      }
    }
  ],
  "Round_Select": [
    {
      "question": {
        "text": "Which candidate frame matches the generated image for future step 1?",
        "Frames": ["Frame[CANDIDATE_1]", "Frame[CANDIDATE_2]", ...]
      },
      "answer": "target:FXX",
      "rationale": "<why this frame is the immediate next future state>"
    },
    {
      "question": {
        "text": "Which candidate frame matches the generated image for future step 2?",
        "Frames": ["Frame[CANDIDATE_1]", "Frame[CANDIDATE_2]", ...]
      },
      "answer": "target:FXX",
      "rationale": "<why this frame matches the next future state>"
    }
  ],
  "evidence_refs": ["F00_CAP", "F01_CAP", "F02_CAP", "C00", "C01"]
}
\end{lstlisting}



\subsection{Examples}
\label{app:instruction_examples}

This section gives one example for each template.

\newcommand{\questionanswer}[2]{%
  \vspace{0.6em}\hrule\vspace{0.55em}
  {\small\raggedright\begin{tabular}{@{}p{\linewidth}@{}}\textbf{Question:} #1\end{tabular}\par}
  \vspace{0.45em}
  {\small\raggedright\begin{tabular}{@{}p{\linewidth}@{}}\textbf{Answer:} #2\end{tabular}\par}
}
\newcommand{\editquestionanswerimage}[3]{%
  \vspace{0.6em}\hrule\vspace{0.55em}
  {\small\bfseries Question:}\par\vspace{0.35em}
  \centering #1\par\vspace{0.45em}
  {\small\raggedright\begin{tabular}{@{}p{\linewidth}@{}}\textbf{Instruction:} #2\end{tabular}\par}
  \vspace{0.55em}{\small\bfseries Answer:}\par\vspace{0.35em}
  \centering #3\par
}
\newcommand{\roundtext}[3]{%
  \vspace{0.65em}\hrule\vspace{0.5em}
  {\small\bfseries #1}\par\vspace{0.35em}
  {\small\raggedright\begin{tabular}{@{}p{\linewidth}@{}}\textbf{Question:} #2\end{tabular}\par}
  \vspace{0.35em}
  {\small\raggedright\begin{tabular}{@{}p{\linewidth}@{}}\textbf{Answer:} #3\end{tabular}\par}
}
\newcommand{\roundtextqimages}[4]{%
  \vspace{0.65em}\hrule\vspace{0.5em}
  {\small\bfseries #1}\par\vspace{0.35em}
  {\small\raggedright\begin{tabular}{@{}p{\linewidth}@{}}\textbf{Question:} #2\end{tabular}\par}
  \vspace{0.35em}\centering #3\par\vspace{0.35em}
  {\small\raggedright\begin{tabular}{@{}p{\linewidth}@{}}\textbf{Answer:} #4\end{tabular}\par}
}
\newcommand{\roundimageqimages}[4]{%
  \vspace{0.65em}\hrule\vspace{0.5em}
  {\small\bfseries #1}\par\vspace{0.35em}
  {\small\raggedright\begin{tabular}{@{}p{\linewidth}@{}}\textbf{Question:} #2\end{tabular}\par}
  \vspace{0.35em}\centering #3\par\vspace{0.35em}
  {\small\bfseries Answer:}\par\vspace{0.35em}
  \centering #4\par
}

\newsavebox{\schemafigbox}
\newenvironment{schemafig}[3]{%
  \def\schemafigcaption{#2}%
  \def\schemafiglabel{#3}%
  \begin{figure}[!htbp]\centering
  \setlength{\fboxsep}{7pt}\setlength{\fboxrule}{0.4pt}
  \begin{lrbox}{\schemafigbox}
  \begin{minipage}{\dimexpr\textwidth-2\fboxsep-2\fboxrule\relax}
  \small\textbf{Instruction-Data Example (#1)}\par\medskip\hrule\medskip
}{%
  \medskip\hrule
  \end{minipage}
  \end{lrbox}
  \fbox{\usebox{\schemafigbox}}
  \caption{\schemafigcaption}
  \label{\schemafiglabel}
  \end{figure}
}

\begin{schemafig}
{Fine-Grained Perception}
{Fine-grained perception example. The sample asks the model to identify a foreground object from a selected visual frame.}
{fig:fine_grained_perception}
\centering \begin{minipage}{0.42\linewidth}\centering \includegraphics[width=\linewidth,height=1.8in,keepaspectratio]{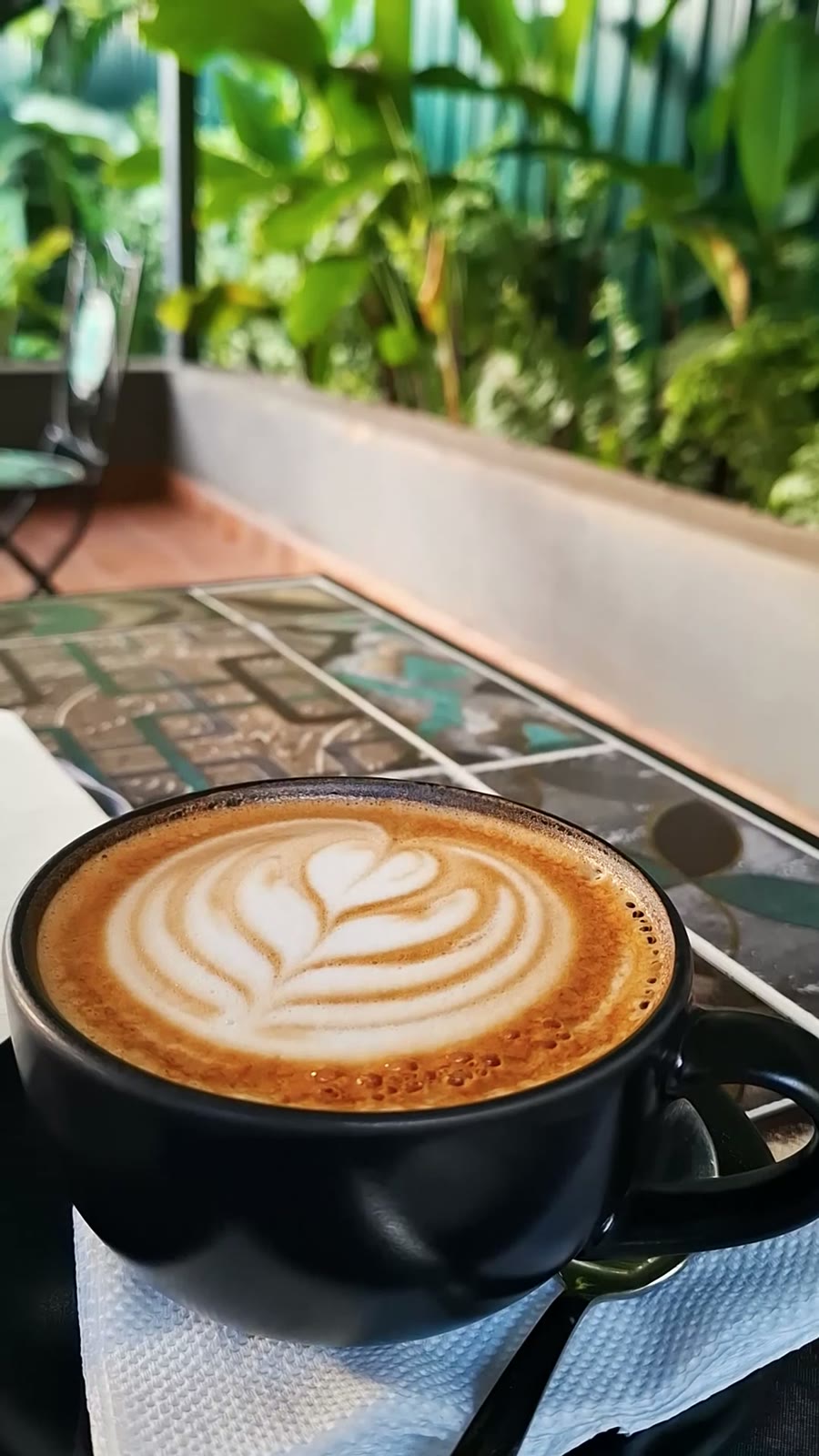}\\[-0.5mm]{\scriptsize\bfseries Input Image}\end{minipage}\par
\questionanswer
{Which object is visible in the foreground? A.\ Plate. B.\ Coffee cup. C.\ Phone. D.\ Book.}
{B.\ Coffee cup.}
\end{schemafig}

\begin{schemafig}
{Visual Reasoning}
{Visual reasoning example. The sample requires inferring the ongoing activity from visual evidence in the image.}
{fig:visual_reasoning}
\centering \begin{minipage}{0.42\linewidth}\centering \includegraphics[width=\linewidth,height=1.28in,keepaspectratio]{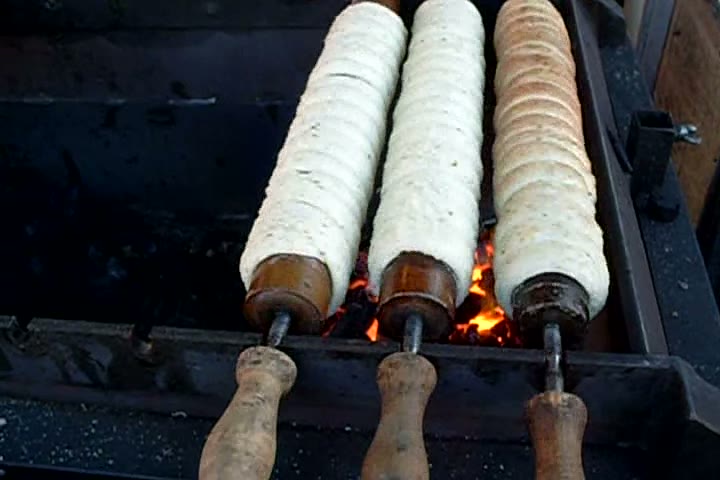}\\[-0.5mm]{\scriptsize\bfseries Input Image}\end{minipage}\par
\questionanswer
{What can be inferred about the activity in the image?}
{Someone is cooking food over heat. The food is positioned above a cooking surface and the visible setup supports an ongoing preparation process.}
\end{schemafig}

\begin{schemafig}
{Multi-Image Logic}
{Multi-image logic example. The sample compares multiple related frames and verifies the shared scene structure and temporal changes.}
{fig:multi_image_logic}
\centering \begin{minipage}{0.28\linewidth}\centering \includegraphics[width=\linewidth,height=1.08in,keepaspectratio]{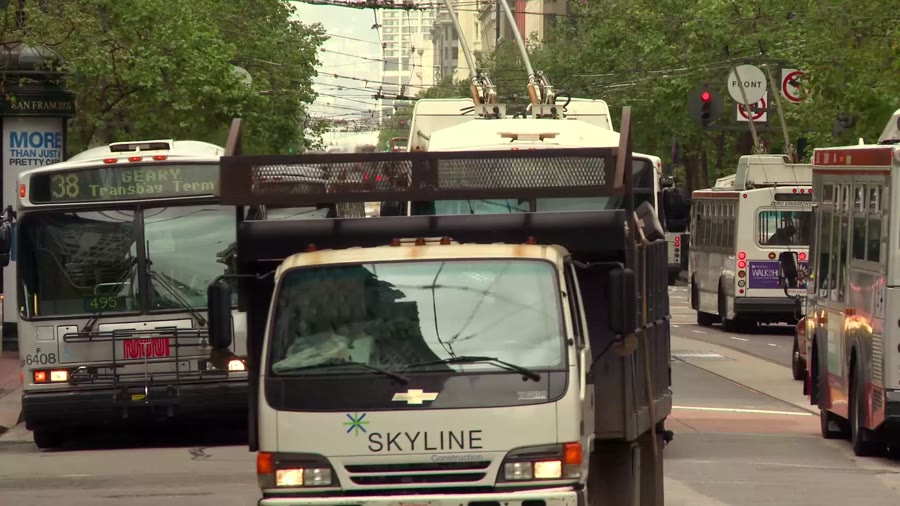}\\[-0.5mm]{\scriptsize\bfseries Image 1}\end{minipage}\hfill \begin{minipage}{0.28\linewidth}\centering \includegraphics[width=\linewidth,height=1.08in,keepaspectratio]{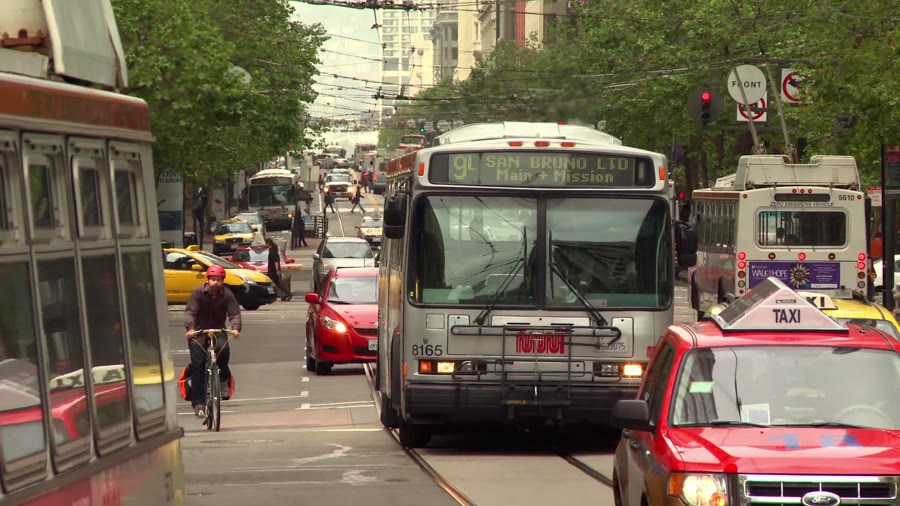}\\[-0.5mm]{\scriptsize\bfseries Image 2}\end{minipage}\hfill \begin{minipage}{0.28\linewidth}\centering \includegraphics[width=\linewidth,height=1.08in,keepaspectratio]{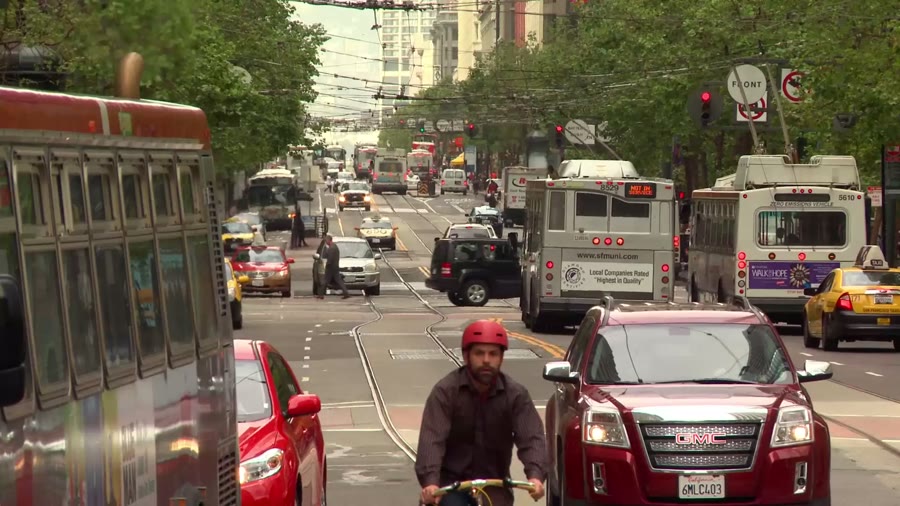}\\[-0.5mm]{\scriptsize\bfseries Image 3}\end{minipage}\par
\roundtext{Round 1: Inspect}
{Do the first two images show the same street environment?}
{Yes. Both show a city street with moving traffic and similar road-side structures.}
\roundtext{Round 2: Compare}
{What changes across the images?}
{The positions of vehicles and pedestrians shift along the street while the road layout remains stable.}
\roundtext{Round 3: Verify}
{Is the comparison supported by all images?}
{Yes. The shared street layout binds the images together, and the moving road users provide the changing evidence.}
\end{schemafig}

\begin{schemafig}
{Editing}
{Image editing example. The sample edits the source image according to a localized visual instruction while preserving unrelated content.}
{fig:edit_example}
\editquestionanswerimage
{\begin{minipage}{0.42\linewidth}\centering \includegraphics[width=\linewidth,height=1.08in,keepaspectratio]{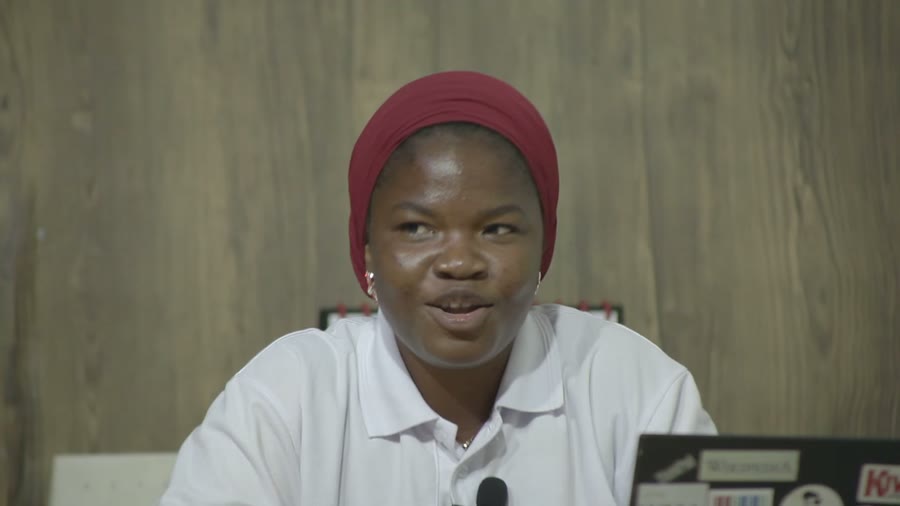}\\[-0.5mm]{\scriptsize\bfseries Source}\end{minipage}}
{Edit the source image so the woman's facial expression changes from neutral/speaking to smiling. Do not change anything else.}
{\includegraphics[width=0.36\linewidth,height=1.58in,keepaspectratio]{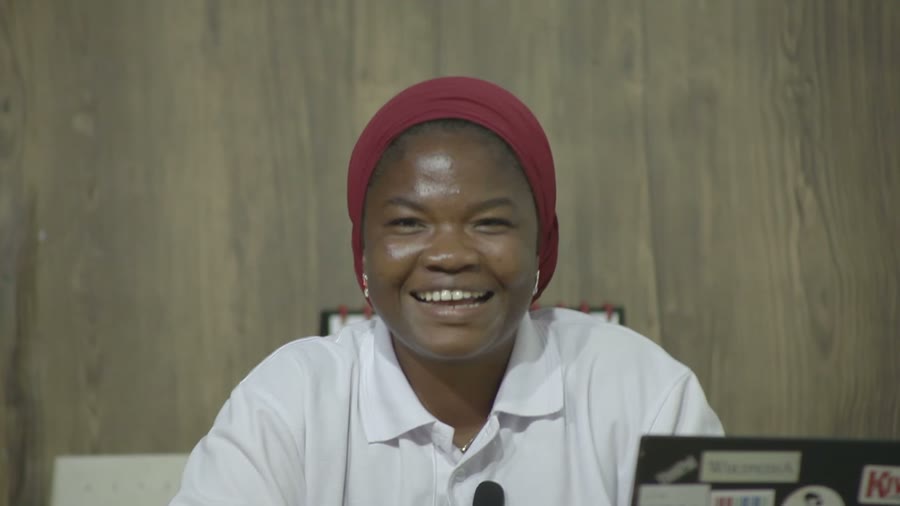}}
\end{schemafig}

\begin{schemafig}
{Grounded Edit-to-Select Generation}
{Grounded edit-to-select generation example. The sample plans a grounded visual edit, generates the target state, and selects the matching candidate.}
{fig:grounded_edit_to_select}
\roundtextqimages{Round 1: Plan}
{What should be changed from the source image to reach the target state?}
{\begin{minipage}{0.38\linewidth}\centering \includegraphics[width=\linewidth,height=1.08in,keepaspectratio]{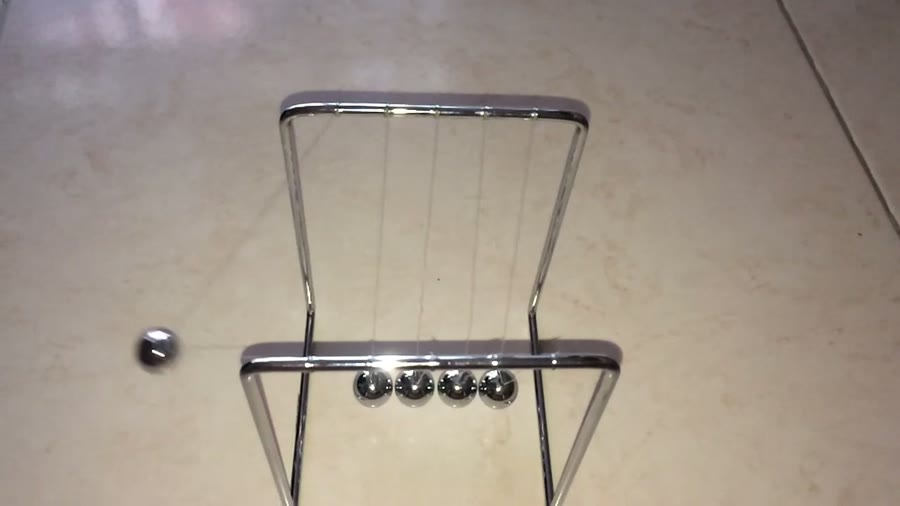}\\[-0.5mm]{\scriptsize\bfseries Source}\end{minipage}\hspace{0.04\linewidth}\begin{minipage}{0.38\linewidth}\centering \includegraphics[width=\linewidth,height=1.08in,keepaspectratio]{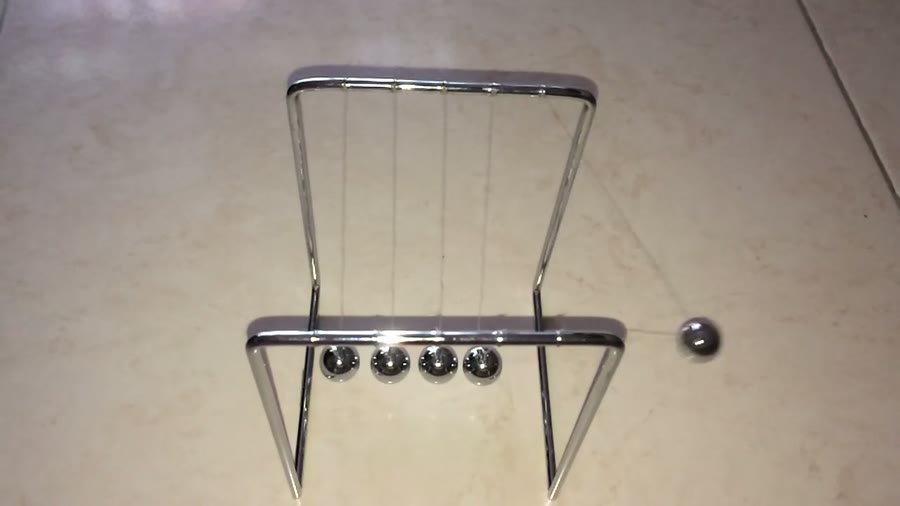}\\[-0.5mm]{\scriptsize\bfseries Target}\end{minipage}}
{Keep the same Newton's cradle setup, but transfer the motion from the left displaced ball to the right displaced ball after the collision.}
\roundimageqimages{Round 2: Generate}
{Generate the target image after applying the grounded edit.}
{\begin{minipage}{0.42\linewidth}\centering \includegraphics[width=\linewidth,height=1.08in,keepaspectratio]{figures/examples/grounded_source.jpg}\\[-0.5mm]{\scriptsize\bfseries Source}\end{minipage}}
{\includegraphics[width=0.36\linewidth,height=1.58in,keepaspectratio]{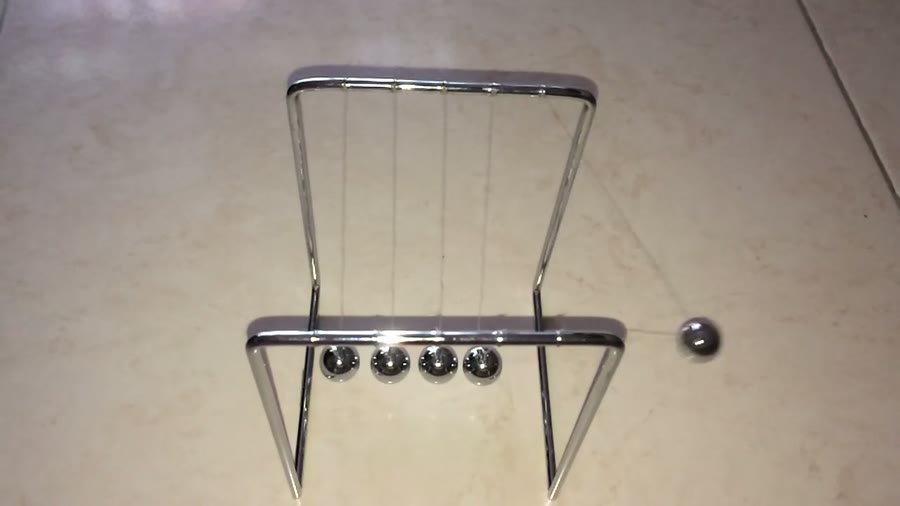}}
\roundtextqimages{Round 3: Select}
{Which candidate image best matches the generated target image?}
{\begin{minipage}{0.25\linewidth}\centering \includegraphics[width=\linewidth,height=1.08in,keepaspectratio]{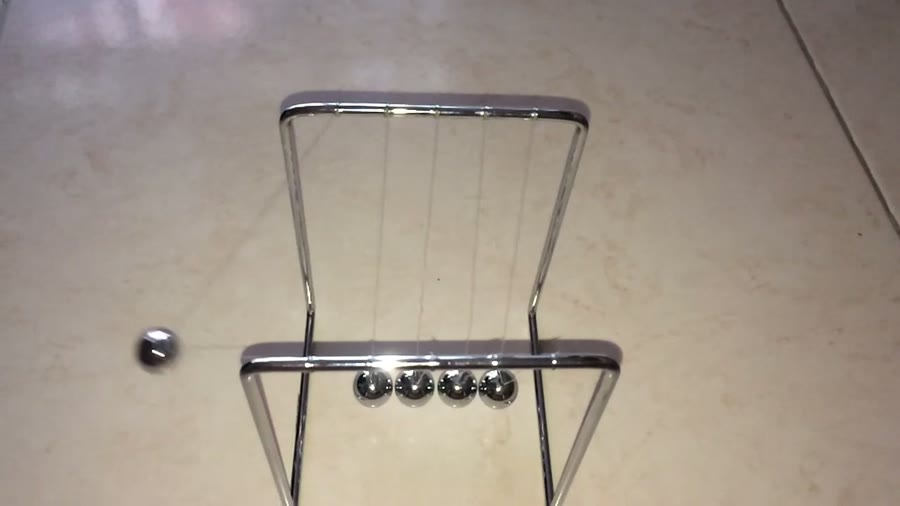}\\[-0.5mm]{\scriptsize\bfseries Candidate A}\end{minipage}\hfill \begin{minipage}{0.25\linewidth}\centering \includegraphics[width=\linewidth,height=1.08in,keepaspectratio]{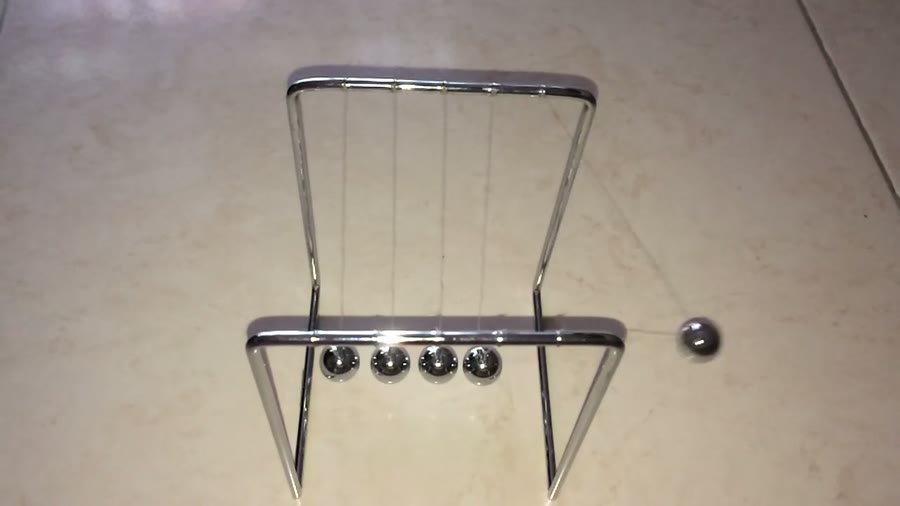}\\[-0.5mm]{\scriptsize\bfseries Candidate B}\end{minipage}\hfill \begin{minipage}{0.25\linewidth}\centering \includegraphics[width=\linewidth,height=1.08in,keepaspectratio]{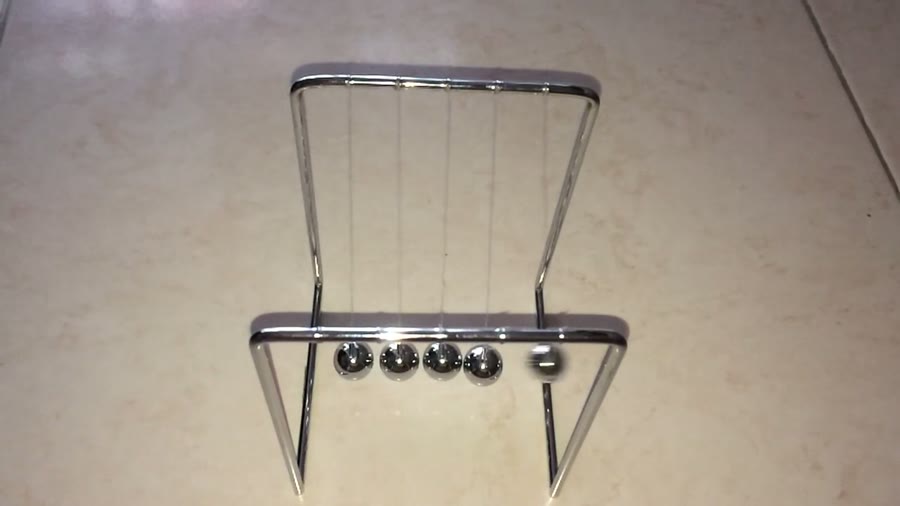}\\[-0.5mm]{\scriptsize\bfseries Candidate C}\end{minipage}}
{Candidate B. It best matches the later state where the right metal ball swings outward while preserving the same cradle and floor background.}
\end{schemafig}

\begin{schemafig}
{Interleaved Understand--Recall--Generate--Select}
{Interleaved understand-recall-generate-select example. The sample alternates between visual understanding, grounded recall, generation, and candidate selection.}
{fig:interleaved_understand_recall_generate_select}
\roundtextqimages{Round 1: Understand}
{What visible facts describe the earlier and later image states?}
{\begin{minipage}{0.38\linewidth}\centering \includegraphics[width=\linewidth,height=1.08in,keepaspectratio]{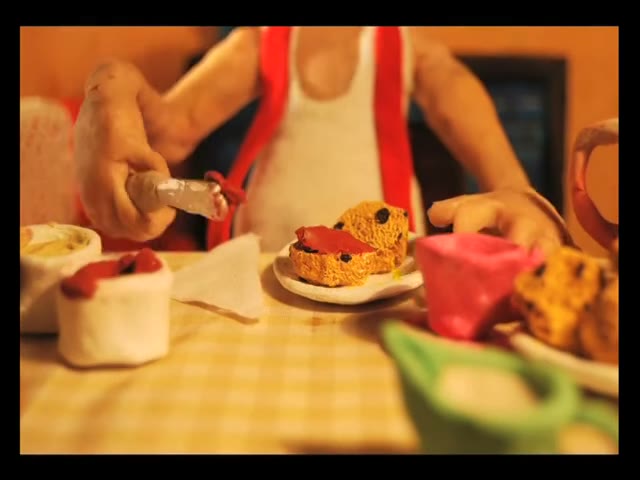}\\[-0.5mm]{\scriptsize\bfseries Earlier}\end{minipage}\hspace{0.04\linewidth} \begin{minipage}{0.38\linewidth}\centering \includegraphics[width=\linewidth,height=1.08in,keepaspectratio]{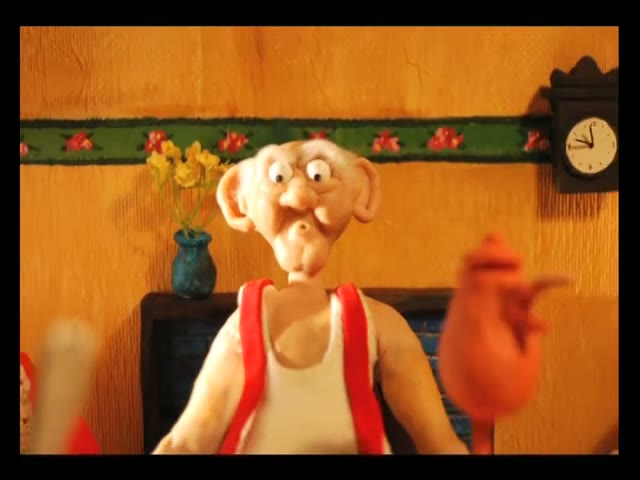}\\[-0.5mm]{\scriptsize\bfseries Later}\end{minipage}}
{The earlier image shows food and hands around a table. The later image shows a clay character in the same warm stop-motion style, now framed as an upper-body view.}
\roundtext{Round 2: Recall}
{What grounded visual change connects the two states?}
{The scene changes from a food-focused tabletop view to a character-focused upper-body view.}
\roundimageqimages{Round 3: Generate}
{Generate the recalled later-state image.}
{\begin{minipage}{0.42\linewidth}\centering \includegraphics[width=\linewidth,height=1.08in,keepaspectratio]{figures/examples/inter_earlier.jpg}\\[-0.5mm]{\scriptsize\bfseries Earlier}\end{minipage}}
{\includegraphics[width=0.36\linewidth,height=1.08in,keepaspectratio]{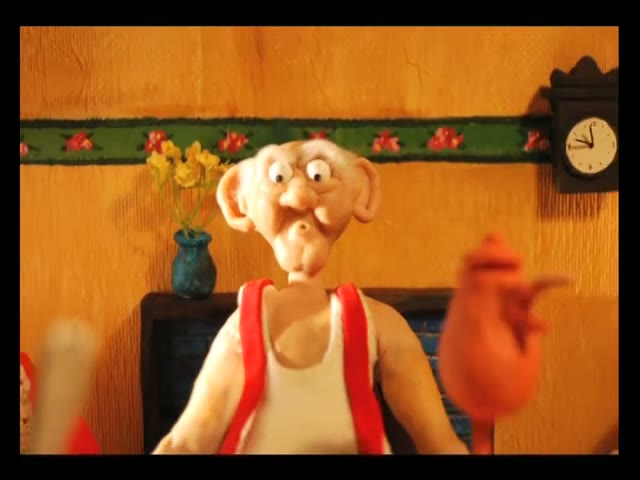}}
\roundtextqimages{Round 4: Select}
{Which candidate matches the generated later-state image?}
{\begin{minipage}{0.25\linewidth}\centering \includegraphics[width=\linewidth,height=1.08in,keepaspectratio]{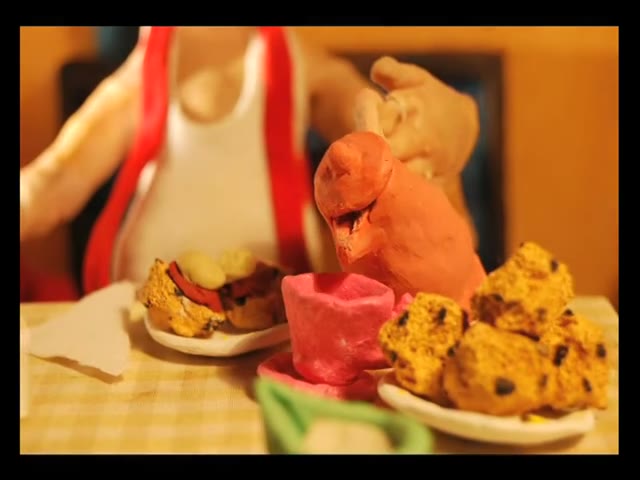}\\[-0.5mm]{\scriptsize\bfseries Candidate A}\end{minipage}\hfill \begin{minipage}{0.25\linewidth}\centering \includegraphics[width=\linewidth,height=1.08in,keepaspectratio]{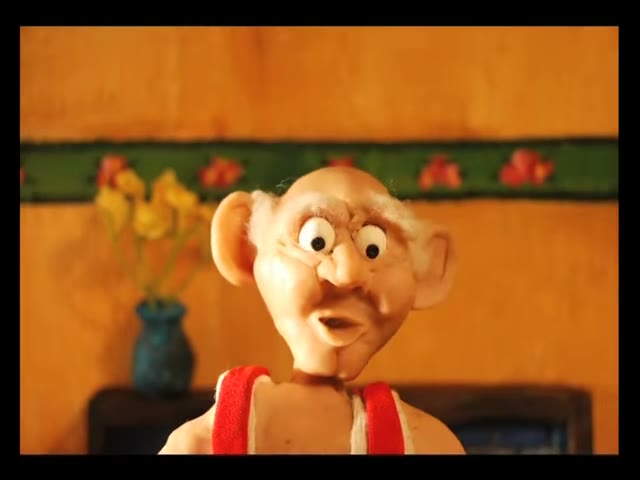}\\[-0.5mm]{\scriptsize\bfseries Candidate B}\end{minipage}\hfill \begin{minipage}{0.25\linewidth}\centering \includegraphics[width=\linewidth,height=1.08in,keepaspectratio]{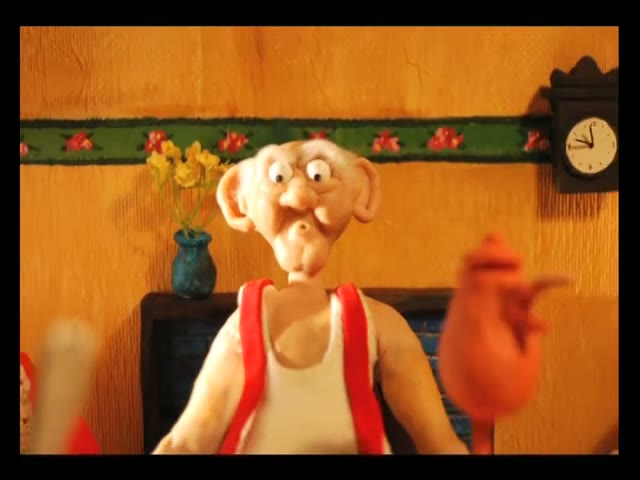}\\[-0.5mm]{\scriptsize\bfseries Candidate C}\end{minipage}}
{Candidate C. It matches the recalled later character-focused view.}
\end{schemafig}

\begin{schemafig}
{Stepwise Sequence Generate-and-Select}
{Stepwise sequence generate-and-select example. The sample follows a continuous action sequence, generates the next step, and selects the matching candidate.}
{fig:stepwise_sequence_generate_select}
\centering \begin{minipage}{0.21\linewidth}\centering \includegraphics[width=\linewidth,height=1.08in,keepaspectratio]{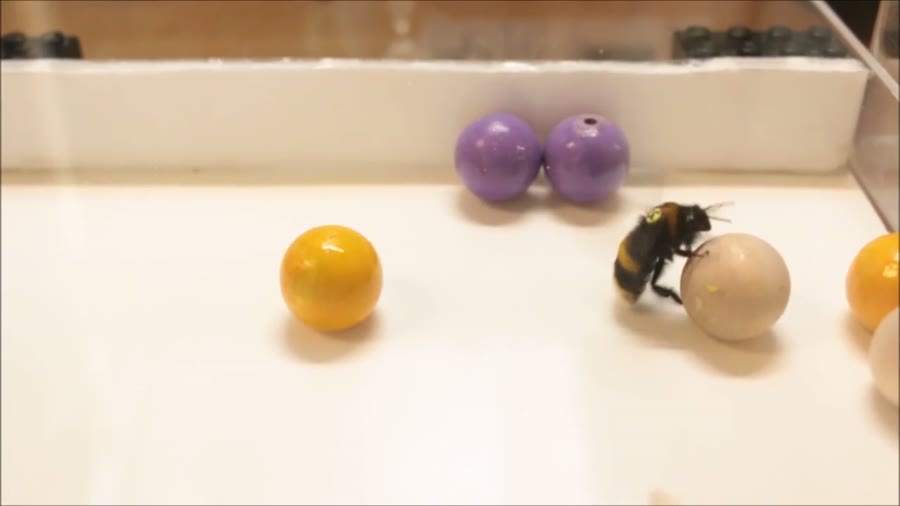}\\[-0.5mm]{\scriptsize\bfseries Start}\end{minipage}\hfill \begin{minipage}{0.21\linewidth}\centering \includegraphics[width=\linewidth,height=1.08in,keepaspectratio]{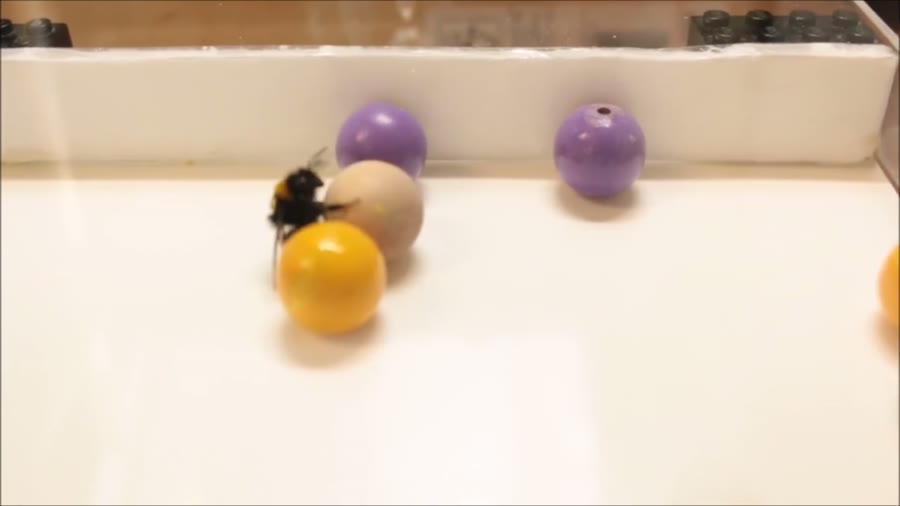}\\[-0.5mm]{\scriptsize\bfseries Step}\end{minipage}\hfill \begin{minipage}{0.21\linewidth}\centering \includegraphics[width=\linewidth,height=1.08in,keepaspectratio]{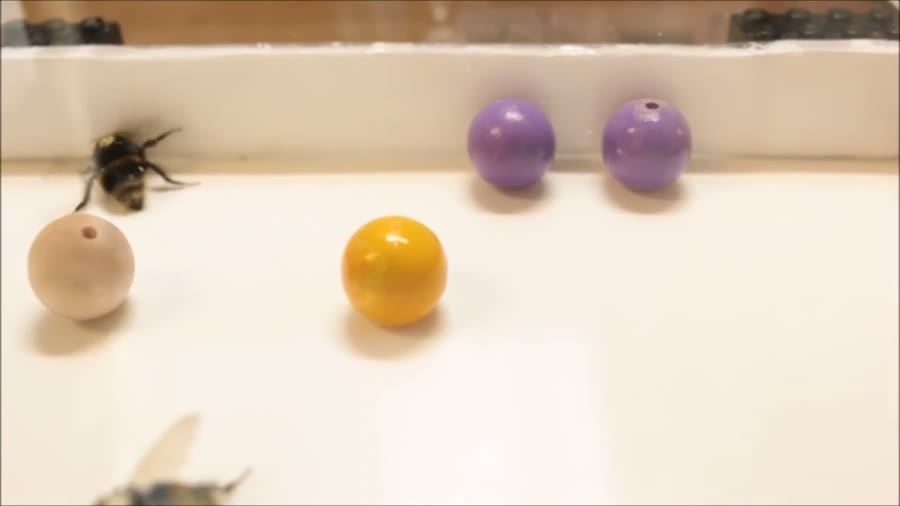}\\[-0.5mm]{\scriptsize\bfseries Step}\end{minipage}\hfill \begin{minipage}{0.21\linewidth}\centering \includegraphics[width=\linewidth,height=1.08in,keepaspectratio]{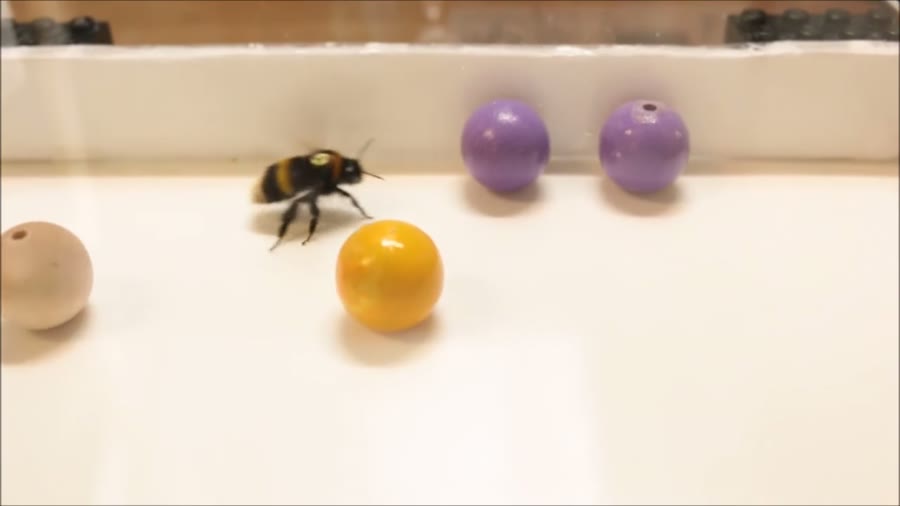}\\[-0.5mm]{\scriptsize\bfseries End}\end{minipage}\par
\roundtext{Round 1: Overall}
{What is the overall coherent visual sequence?}
{The sequence shows a bumblebee approaching a small ball, contacting it, and rolling it along the surface.}
\roundtext{Round 2: Steps}
{What are the grounded step-by-step state changes?}
{Step 1: the bee approaches the ball. Step 2: it touches and grips the ball. Step 3: it rolls the ball forward.}
\roundimageqimages{Round 3: Generate}
{Generate the next real action frame in which the bee is rolling the ball.}
{\begin{minipage}{0.42\linewidth}\centering \includegraphics[width=\linewidth,height=1.08in,keepaspectratio]{figures/examples/step_mid1.jpg}\\[-0.5mm]{\scriptsize\bfseries Current}\end{minipage}}
{\includegraphics[width=0.36\linewidth,height=1.08in,keepaspectratio]{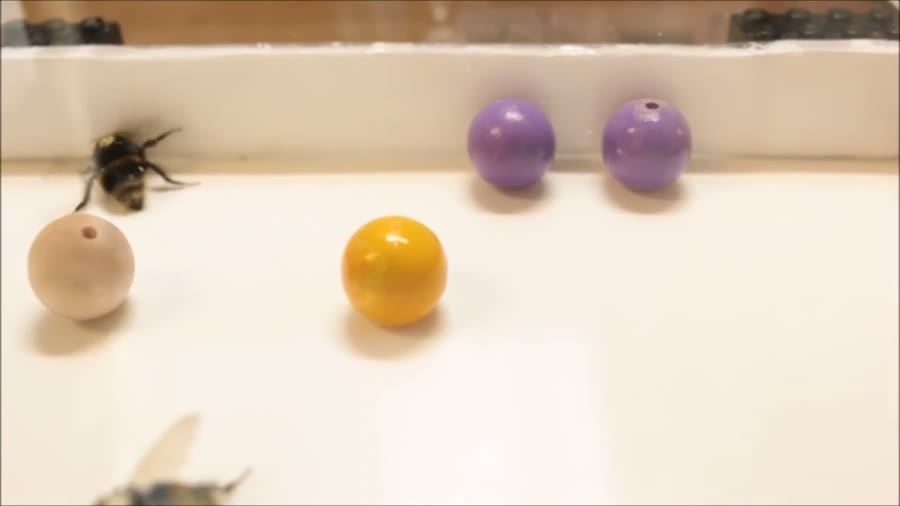}}
\roundtextqimages{Round 4: Select}
{Which candidate matches the generated step image?}
{\begin{minipage}{0.25\linewidth}\centering \includegraphics[width=\linewidth,height=1.08in,keepaspectratio]{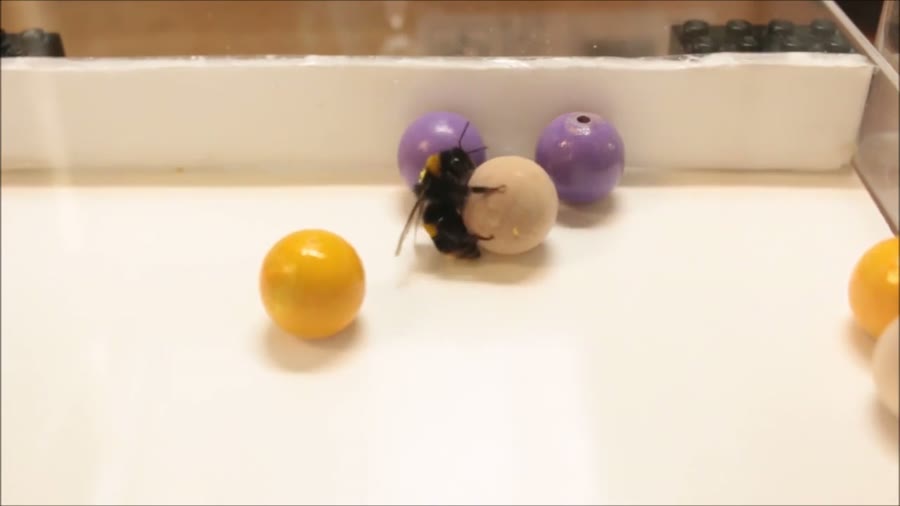}\\[-0.5mm]{\scriptsize\bfseries Candidate A}\end{minipage}\hfill \begin{minipage}{0.25\linewidth}\centering \includegraphics[width=\linewidth,height=1.08in,keepaspectratio]{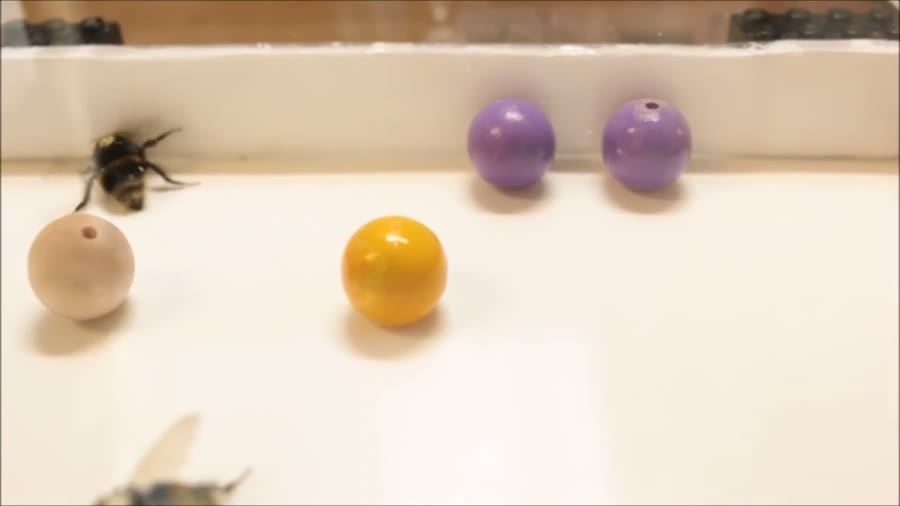}\\[-0.5mm]{\scriptsize\bfseries Candidate B}\end{minipage}\hfill \begin{minipage}{0.25\linewidth}\centering \includegraphics[width=\linewidth,height=1.08in,keepaspectratio]{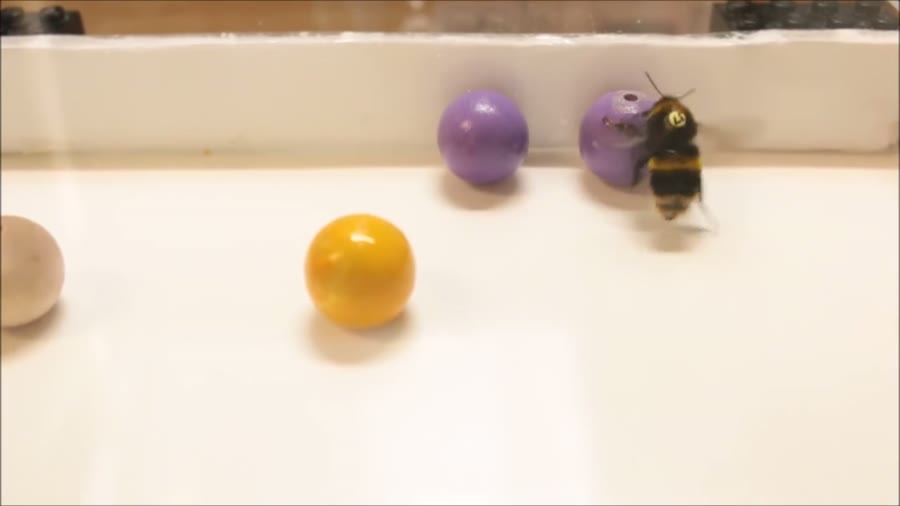}\\[-0.5mm]{\scriptsize\bfseries Candidate C}\end{minipage}}
{Candidate B. It matches the bee rolling the ball in the next visible action state.}
\end{schemafig}

\begin{schemafig}
{Middle-Step Gap Generate-and-Fill}
{Middle-step gap generate-and-fill example. The sample infers and generates the missing intermediate frame between two temporally ordered endpoints.}
{fig:middle_step_gap_generate_fill}
\roundtextqimages{Round 1: Explain}
{What visual state should fit between the earlier and later images?}
{\begin{minipage}{0.38\linewidth}\centering \includegraphics[width=\linewidth,height=0.98in,keepaspectratio]{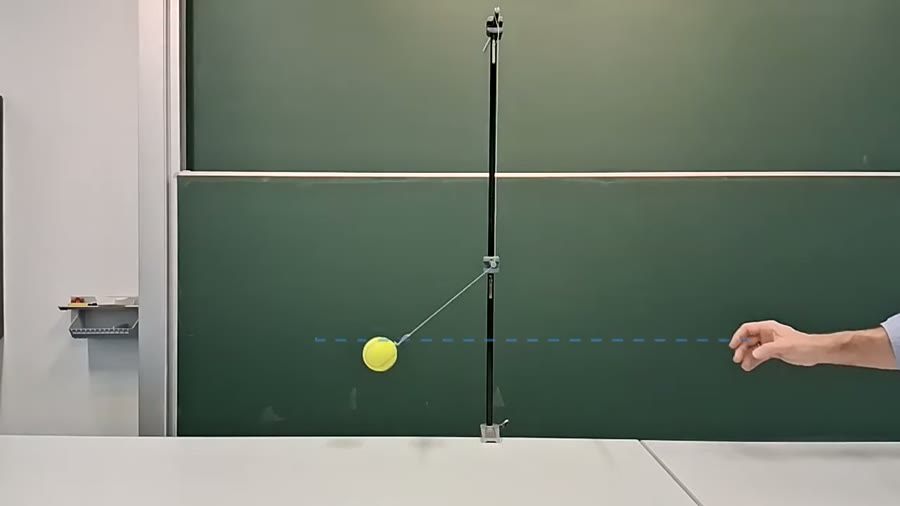}\\[-0.5mm]{\scriptsize\bfseries Earlier}\end{minipage}\hspace{0.04\linewidth} \begin{minipage}{0.38\linewidth}\centering \includegraphics[width=\linewidth,height=0.98in,keepaspectratio]{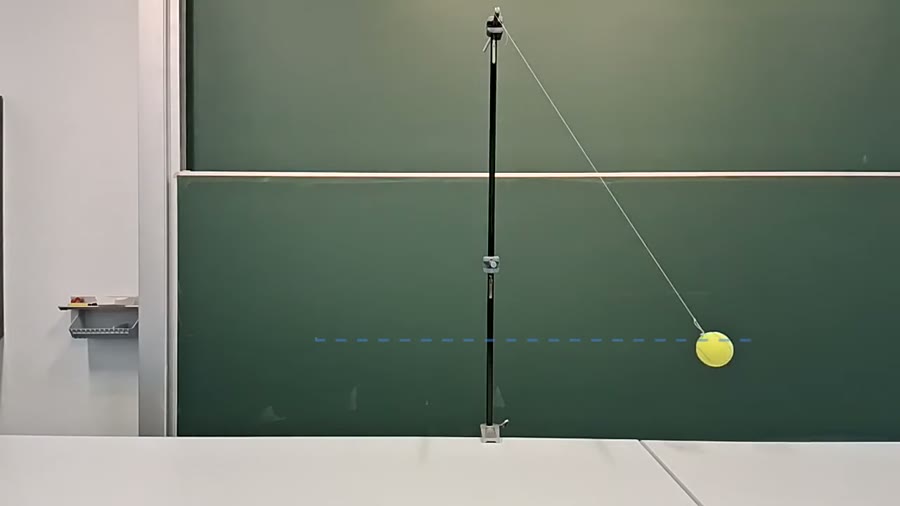}\\[-0.5mm]{\scriptsize\bfseries Later}\end{minipage}}
{The middle state should show the pendulum between the earlier left-side position and the later right-side position.}
\roundimageqimages{Round 2: Generate}
{Generate the middle-step image that fits between the endpoints.}
{\begin{minipage}{0.38\linewidth}\centering \includegraphics[width=\linewidth,height=0.98in,keepaspectratio]{figures/examples/middle_earlier.jpg}\\[-0.5mm]{\scriptsize\bfseries Earlier}\end{minipage}\hspace{0.04\linewidth} \begin{minipage}{0.38\linewidth}\centering \includegraphics[width=\linewidth,height=0.98in,keepaspectratio]{figures/examples/middle_later.jpg}\\[-0.5mm]{\scriptsize\bfseries Later}\end{minipage}}
{\includegraphics[width=0.36\linewidth,height=0.98in,keepaspectratio]{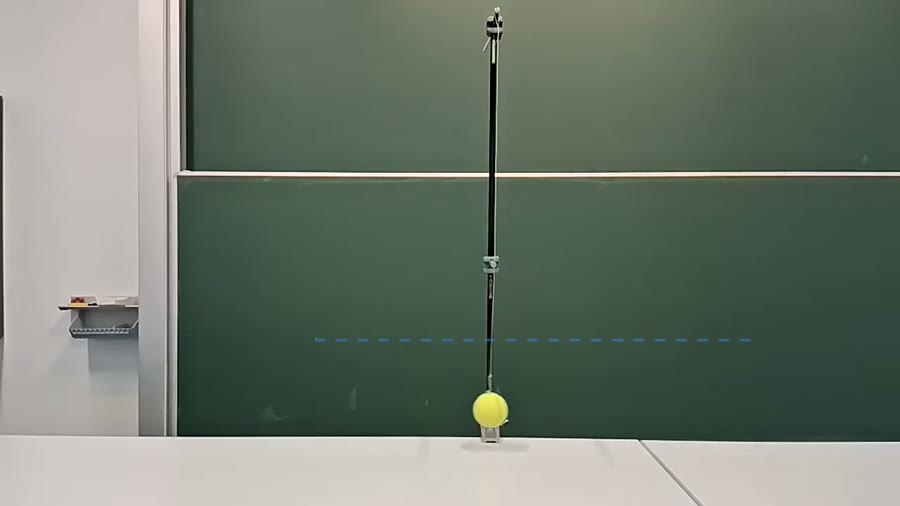}}
\roundtextqimages{Round 3: Select}
{Which candidate matches the generated middle-step image?}
{\begin{minipage}{0.25\linewidth}\centering \includegraphics[width=\linewidth,height=1.48in,keepaspectratio]{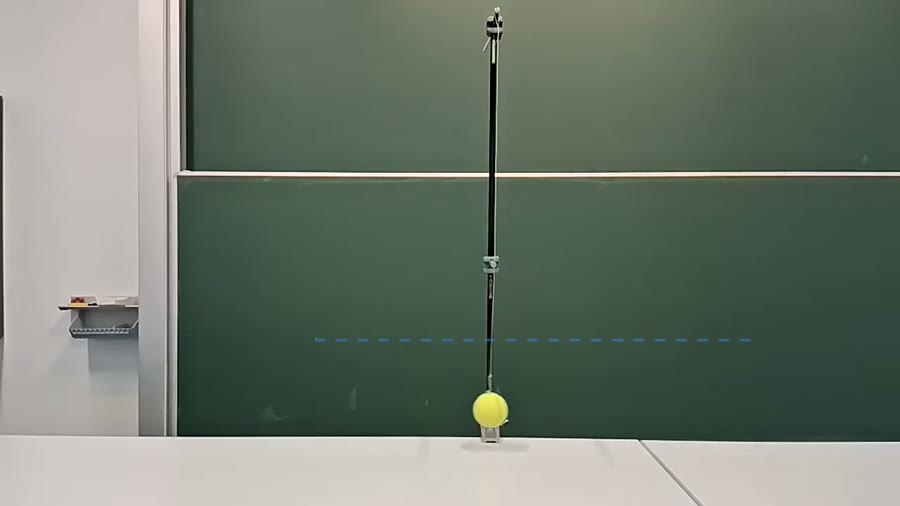}\\[-0.5mm]{\scriptsize\bfseries Candidate A}\end{minipage}\hfill \begin{minipage}{0.25\linewidth}\centering \includegraphics[width=\linewidth,height=1.48in,keepaspectratio]{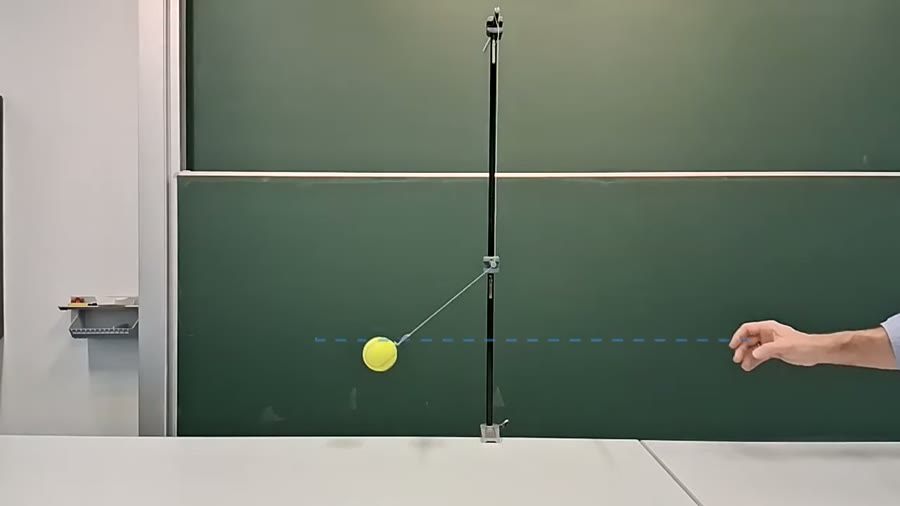}\\[-0.5mm]{\scriptsize\bfseries Candidate B}\end{minipage}\hfill \begin{minipage}{0.25\linewidth}\centering \includegraphics[width=\linewidth,height=1.48in,keepaspectratio]{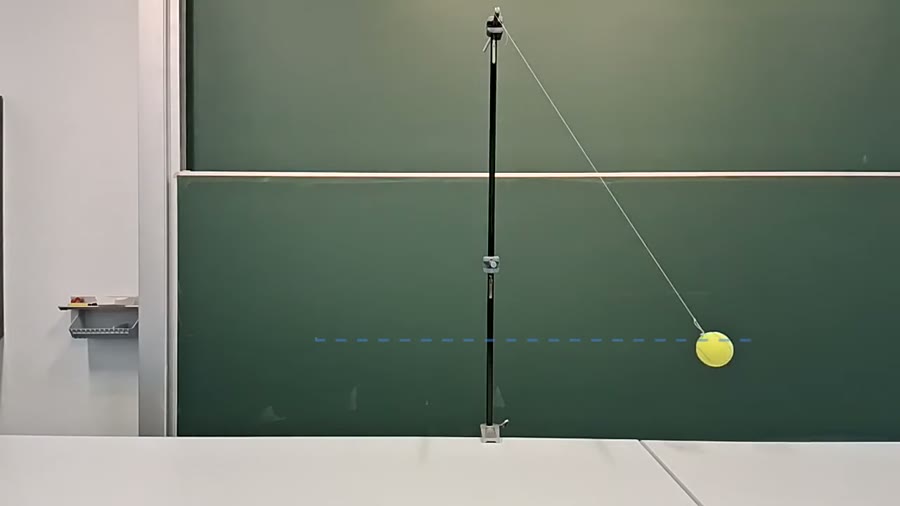}\\[-0.5mm]{\scriptsize\bfseries Candidate C}\end{minipage}}
{Candidate A. It shows the pendulum passing through the intermediate position between the earlier and later swing states.}
\end{schemafig}

\begin{schemafig}
{Instructional Sequential Generate-and-Select}
{Instructional sequential generate-and-select example. The sample follows a stable natural blooming process, generates future visual steps, and selects the matching sequence.}
{fig:instructional_sequential_generate_select}

\roundtextqimages{Round 1: Overall}
{What future visual sequence should follow from the context images?}
{\begin{minipage}{0.30\linewidth}\centering
\includegraphics[width=\linewidth,height=0.68in,keepaspectratio]{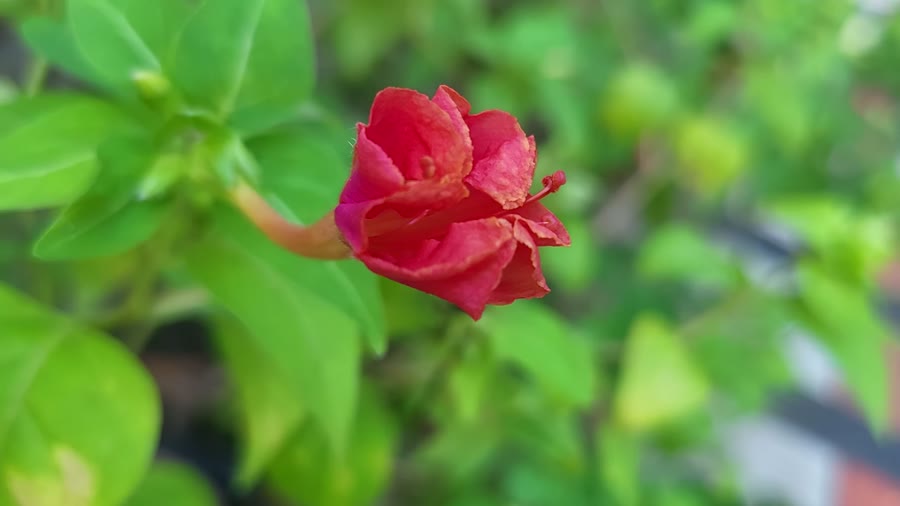}\\[-0.5mm]
{\scriptsize\bfseries Context 1}
\end{minipage}\hspace{0.045\linewidth}
\begin{minipage}{0.30\linewidth}\centering
\includegraphics[width=\linewidth,height=0.68in,keepaspectratio]{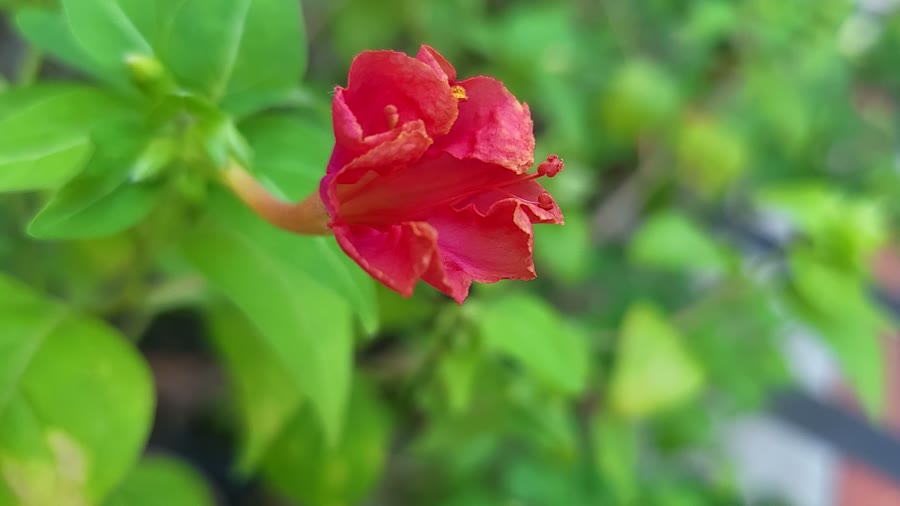}\\[-0.5mm]
{\scriptsize\bfseries Context 2}
\end{minipage}}
{The flower-blooming process should continue through three visible states in the same outdoor close-up: the petals open wider, the bloom expands, and the flower reaches a fuller open state.}

\roundtext{Round 2: Edit}
{What visible edit should be made at each future step?}
{Step 1: open the petals slightly farther. Step 2: continue unfolding the red petals. Step 3: show the flower in a fuller open bloom.}

\roundtext{Round 3: Reason}
{Why do these future steps follow from the context or previous state?}
{The context shows the same flower in a stable natural scene, so the future images should advance the blooming state while preserving the red flower and green-leaf composition.}

\roundimageqimages{Round 4: Generate}
{Generate the three future target images that follow from the context.}
{\begin{minipage}{0.30\linewidth}\centering
\includegraphics[width=\linewidth,height=0.68in,keepaspectratio]{figures/examples/future_context1.jpg}\\[-0.5mm]
{\scriptsize\bfseries Context 1}
\end{minipage}\hspace{0.045\linewidth}
\begin{minipage}{0.30\linewidth}\centering
\includegraphics[width=\linewidth,height=0.68in,keepaspectratio]{figures/examples/future_context2.jpg}\\[-0.5mm]
{\scriptsize\bfseries Context 2}
\end{minipage}}
{\begin{minipage}{0.20\linewidth}\centering
\includegraphics[width=\linewidth,height=0.68in,keepaspectratio]{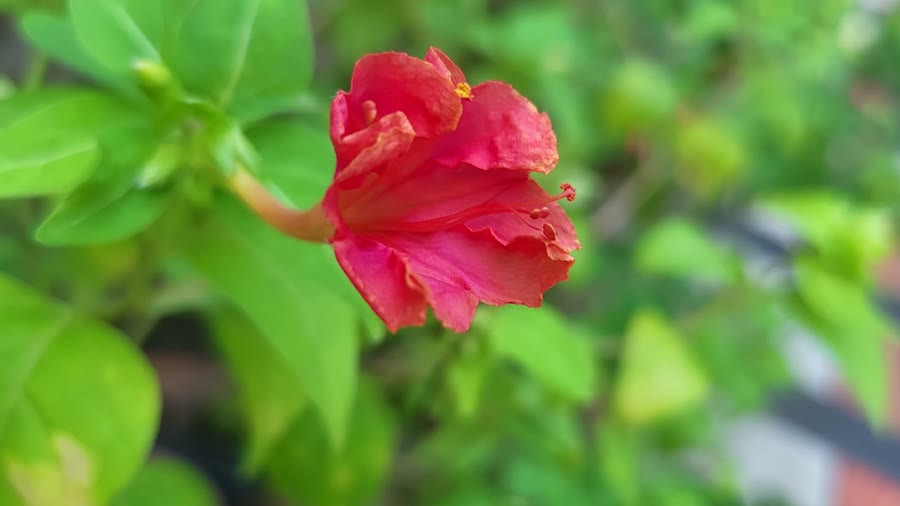}\\[-0.5mm]
{\scriptsize\bfseries }
\end{minipage}\hspace{0.045\linewidth}
\begin{minipage}{0.20\linewidth}\centering
\includegraphics[width=\linewidth,height=0.68in,keepaspectratio]{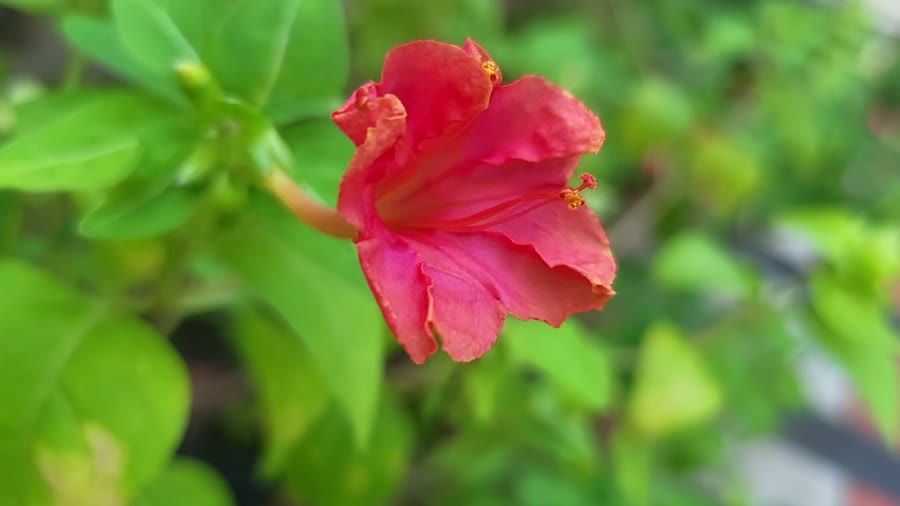}\\[-0.5mm]
{\scriptsize\bfseries }
\end{minipage}\hspace{0.045\linewidth}
\begin{minipage}{0.20\linewidth}\centering
\includegraphics[width=\linewidth,height=0.68in,keepaspectratio]{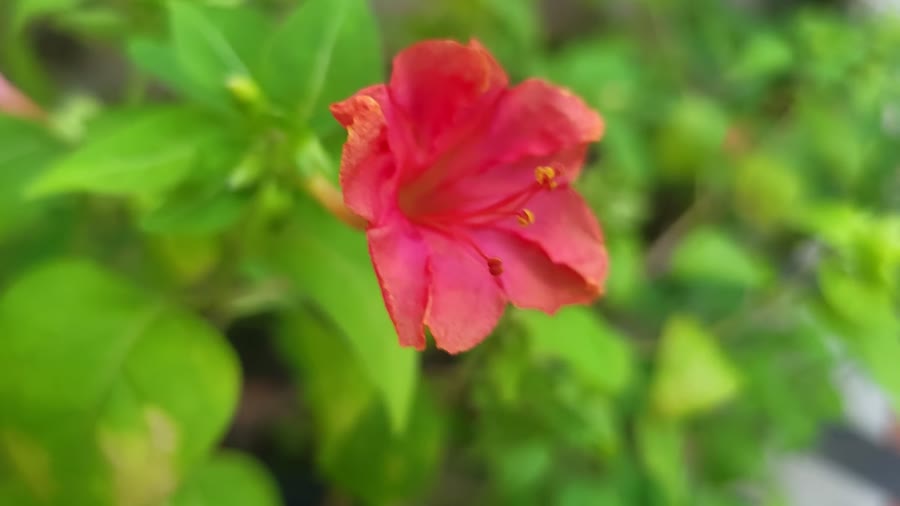}\\[-0.5mm]
{\scriptsize\bfseries }
\end{minipage}}

\roundtextqimages{Round 5: Select}
{Which three candidate images match the generated future sequence?}
{\centering
\begin{minipage}{0.16\linewidth}\centering
\includegraphics[width=\linewidth,height=0.58in,keepaspectratio]{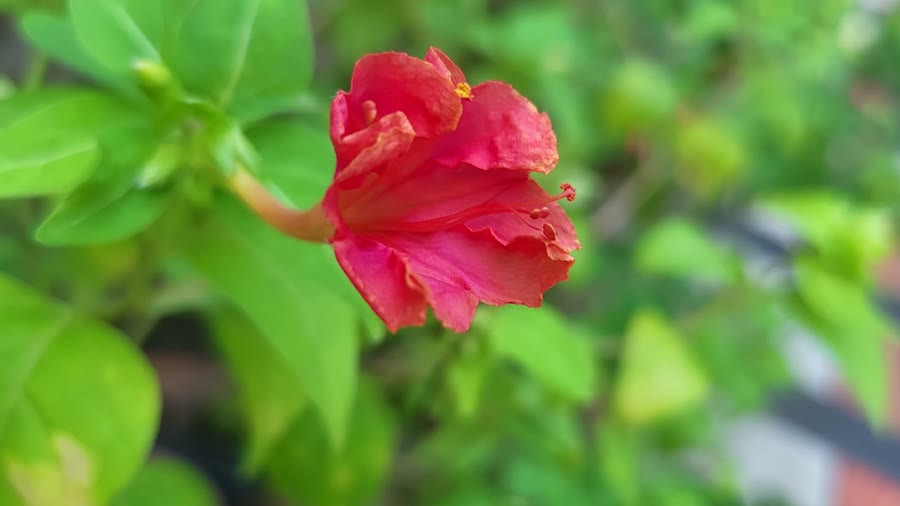}\\[-0.5mm]
{\scriptsize\bfseries A1}
\end{minipage}\hspace{0.055\linewidth}
\begin{minipage}{0.16\linewidth}\centering
\includegraphics[width=\linewidth,height=0.58in,keepaspectratio]{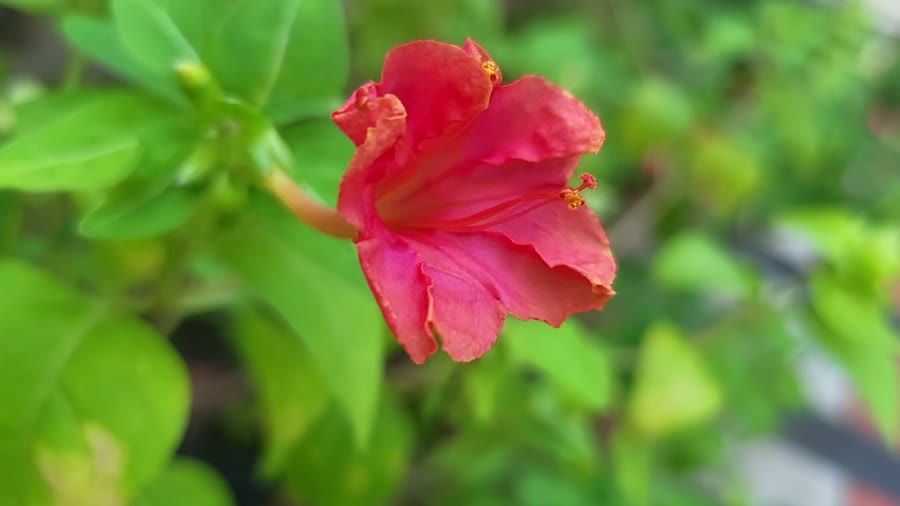}\\[-0.5mm]
{\scriptsize\bfseries A2}
\end{minipage}\hspace{0.055\linewidth}
\begin{minipage}{0.16\linewidth}\centering
\includegraphics[width=\linewidth,height=0.58in,keepaspectratio]{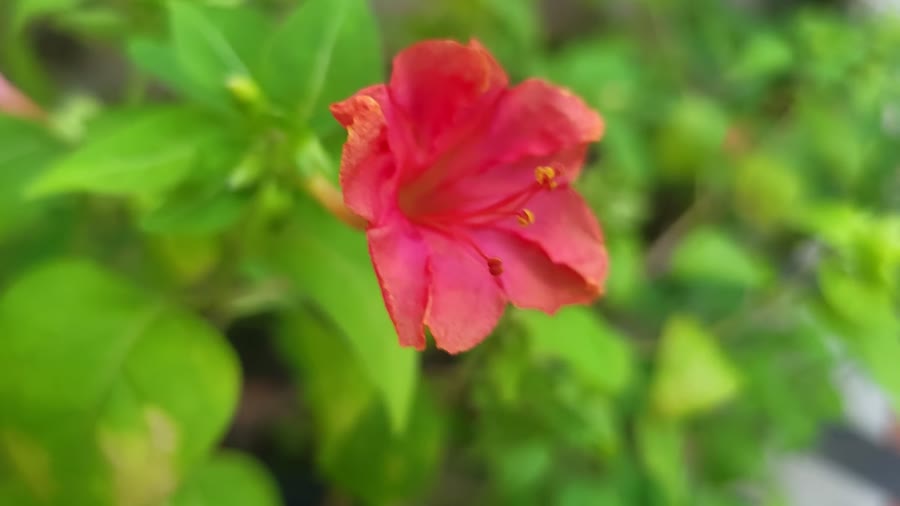}\\[-0.5mm]
{\scriptsize\bfseries A3}
\end{minipage}\par\vspace{0.18em}
\begin{minipage}{0.16\linewidth}\centering
\includegraphics[width=\linewidth,height=0.58in,keepaspectratio]{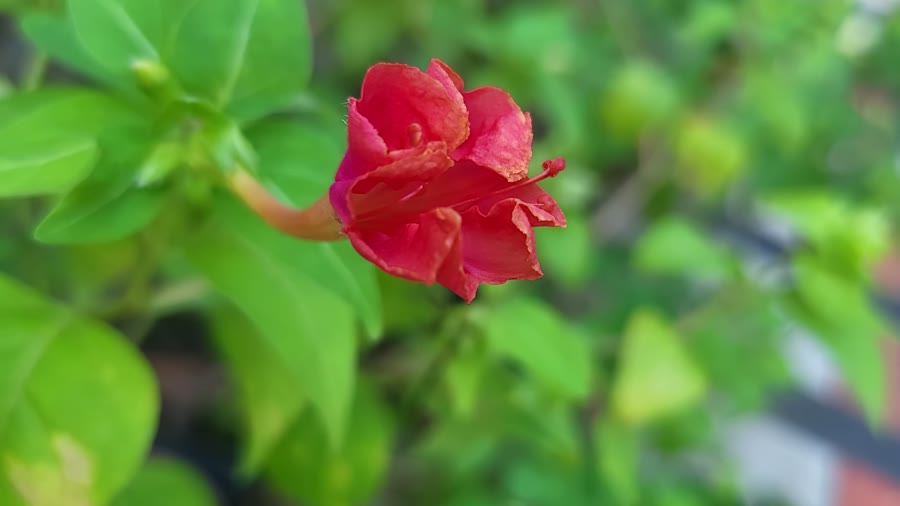}\\[-0.5mm]
{\scriptsize\bfseries B1}
\end{minipage}\hspace{0.055\linewidth}
\begin{minipage}{0.16\linewidth}\centering
\includegraphics[width=\linewidth,height=0.58in,keepaspectratio]{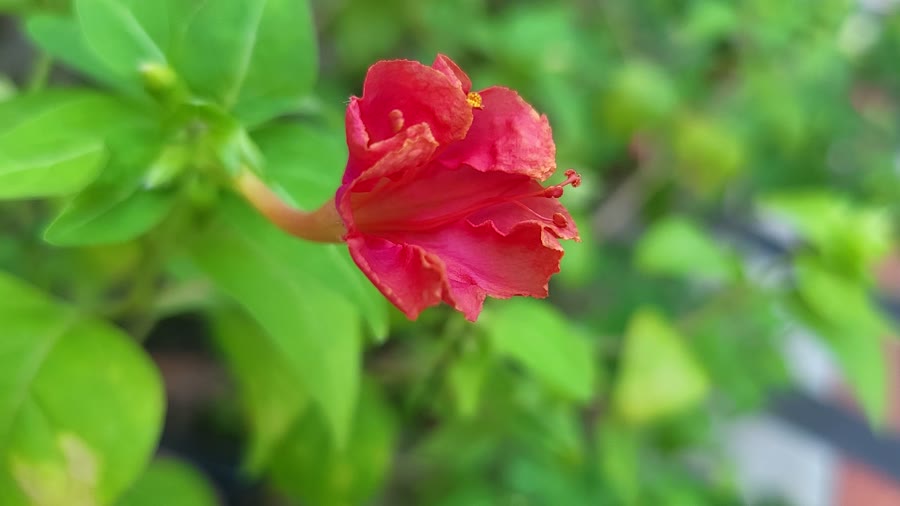}\\[-0.5mm]
{\scriptsize\bfseries B2}
\end{minipage}\hspace{0.055\linewidth}
\begin{minipage}{0.16\linewidth}\centering
\includegraphics[width=\linewidth,height=0.58in,keepaspectratio]{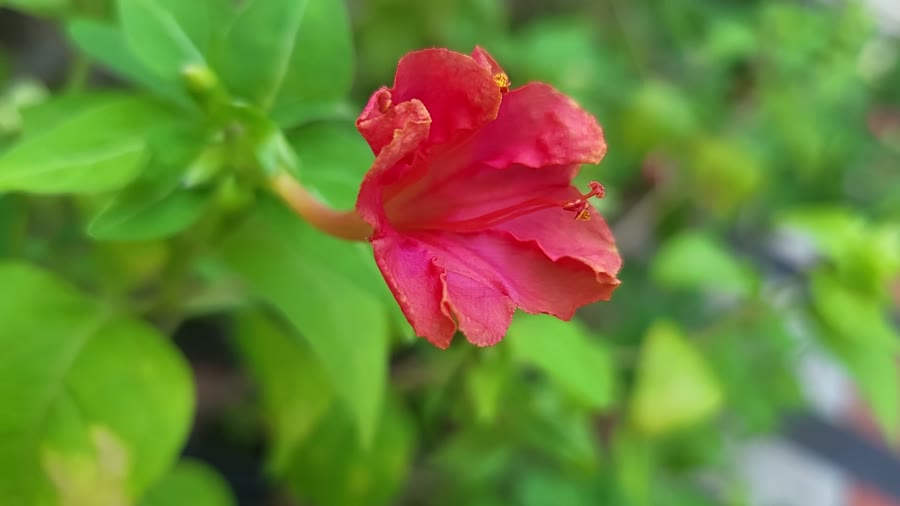}\\[-0.5mm]
{\scriptsize\bfseries B3}
\end{minipage}\par\vspace{0.18em}
\begin{minipage}{0.16\linewidth}\centering
\includegraphics[width=\linewidth,height=0.58in,keepaspectratio]{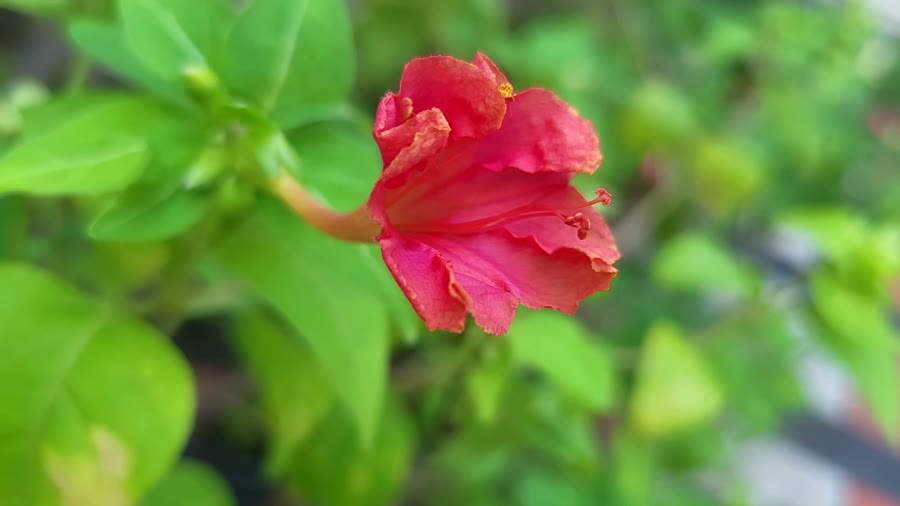}\\[-0.5mm]
{\scriptsize\bfseries C1}
\end{minipage}\hspace{0.055\linewidth}
\begin{minipage}{0.16\linewidth}\centering
\includegraphics[width=\linewidth,height=0.58in,keepaspectratio]{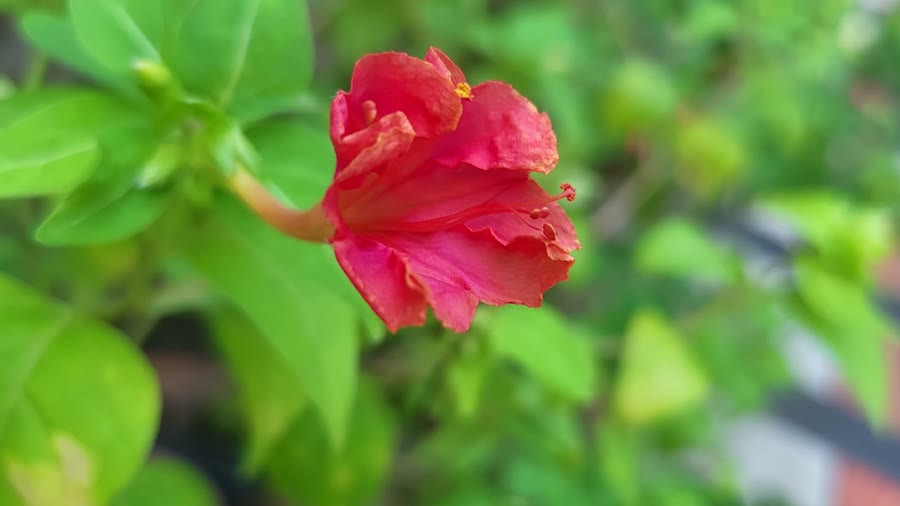}\\[-0.5mm]
{\scriptsize\bfseries C2}
\end{minipage}\hspace{0.055\linewidth}
\begin{minipage}{0.16\linewidth}\centering
\includegraphics[width=\linewidth,height=0.58in,keepaspectratio]{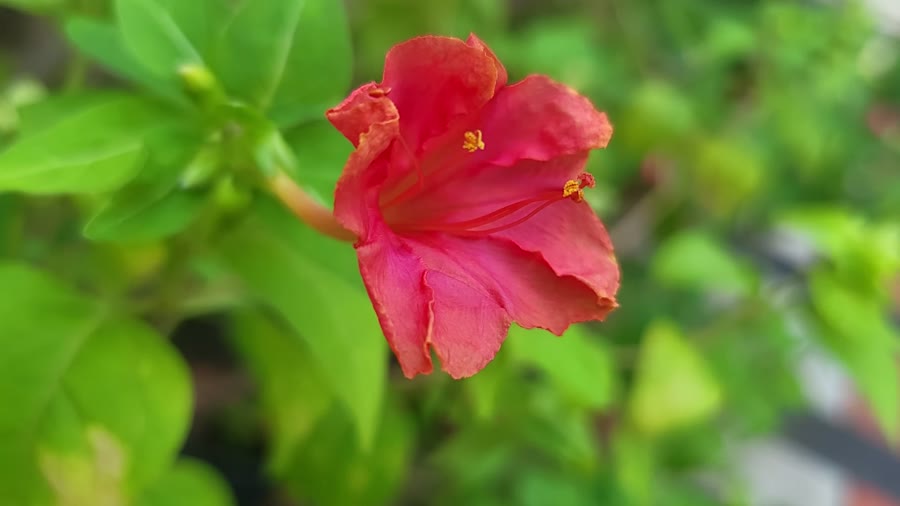}\\[-0.5mm]
{\scriptsize\bfseries C3}
\end{minipage}}
{Select A1, A2, and A3. They match the generated blooming sequence: petals open wider, the flower expands, and the bloom reaches a fuller open state.}

\end{schemafig}

\FloatBarrier

\section{Quality Control and Filtering Rules}
\label{app:quality_control_filtering_rules}

\subsection{Gate A Prompt}
\label{app:shot_level_admission}
The detailed Gate A prompt is shown in Fig.~\ref{fig:shot_admission_prompt_appendix}.

\begin{figure}[!htbp]
\centering
\begin{tcolorbox}[
    width=0.98\linewidth,
    colback=white,
    colframe=black,
    boxrule=0.4pt,
    arc=0pt,
    left=7pt,
    right=7pt,
    top=7pt,
    bottom=7pt
]
\footnotesize
\textbf{Shot Admission Prompt}\par
\medskip\hrule\medskip
{\ttfamily\raggedright\setlength{\parindent}{0pt}\setlength{\parskip}{4pt}%
You are a strict shot-admission judge for a video frame selection pipeline.\par
Judge whether the current shot is worth sending to downstream frame sampling.\par
Keep shots only if they are visually usable and likely to yield informative frames.\par
Reject shots that are mainly low-information filler, weak transitions, visually poor, or not worth spending frame-sampling budget on.\par
Return strict JSON with keys usable\_score, quality\_score, informativeness\_score, decision, reason, evidence.\par
\vspace{0.35em}
INPUTS:\par
Video summary: \textless{}video\_summary\textgreater{}\par
Video type: \textless{}video\_type\textgreater{}\par
Chapter role: \textless{}chapter\_role\textgreater{}\par
Chapter summary: \textless{}chapter\_summary\textgreater{}\par
Shot summary: \textless{}shot\_summary\textgreater{}\par
Shot role: \textless{}shot\_role\textgreater{}\par
Scene type: \textless{}scene\_type\textgreater{}\par
Visual quality hint: \textless{}visual\_quality\_hint\textgreater{}\par
Informativeness hint: \textless{}informativeness\_hint\textgreater{}\par
Sampling hint: \textless{}candidate\_sampling\_hint\textgreater{}\par
Representative frame count: \textless{}representative\_frame\_count\textgreater{}\par
\vspace{0.35em}
OUTPUT FORMAT (strict JSON only):\par
\{\par
"usable\_score": 0.8,\par
"quality\_score": 0.8,\par
"informativeness\_score": 0.8,\par
"decision": "accept",\par
"reason": "The shot is usable and informative.",\par
"evidence": "The visible scene contains meaningful content."\par
\}\par
Return JSON only. Scores and decisions must reflect the input.\par

\vspace{0.35em}
FIELD RULES:\par
- decision must be either "accept" or "reject".\par
- reason must concisely explain the decision.\par
- Return exactly one valid JSON object without any additional text.\par

}
\medskip\hrule
\end{tcolorbox}
\caption{Prompt for shot admission.}
\label{fig:shot_admission_prompt_appendix}
\end{figure}

\subsection{Gate B Rules \& Prompt}
\label{app:frame_level_quality}

After sampling, candidate frames enter a two-stage technical quality filter. 
The fast stage uses OpenCV statistics, including Laplacian variance, brightness, grayscale standard deviation, and histogram entropy, to reject frames with obvious blur, extreme exposure, low contrast, or low informativeness.
The model stage uses VLM to inspect higher-level visual usability.
The detailed Gate B prompt is shown in Fig.~\ref{fig:quality_assessment_prompt_appendix}.

\begin{figure}[!htbp]
\centering
\begin{tcolorbox}[
    width=0.98\linewidth,
    colback=white,
    colframe=black,
    boxrule=0.4pt,
    arc=0pt,
    left=7pt,
    right=7pt,
    top=7pt,
    bottom=7pt
]
\fontsize{8}{9}\selectfont
\textbf{Quality Assessment Prompt}\par
\medskip\hrule\medskip
{\ttfamily\raggedright\setlength{\parindent}{0pt}\setlength{\parskip}{4pt}%
You are a strict technical image quality inspector for dataset filtering.\par
Judge only technical quality and cleanliness of the visible image, not aesthetics, narrative value, or semantic importance. Inspect the image using domain-appropriate standards for real photos, animation, game renders, screen/UI frames, and mixed cases. Do not treat intentional stylization, high/low-key lighting, bloom, flat shading, or non-photorealism as defects unless they harm usability.\par
Evaluate only meaningful regions, including important subjects, informative structures, readable text, interfaces, faces, hands, and products. Borderline, ambiguous, or mildly degraded cases should fail under strict filtering. For each dimension, output pass or fail with one short reason, and generate a 1--2 sentence factual caption for downstream frame-value scoring.\par
QUALITY DIMENSIONS:\par
1. exposure\_brightness\par
2. focus\_sharpness\par
3. motion\_blur\_camera\_shake\_ghosting\par
4. noise\_grain\_low\_light\_degradation\par
5. artifacts\_compression\_resampling\_ai\_generated\par
6. color\_fidelity\_white\_balance\_color\_cast\par
7. contrast\_dynamic\_range\_detail\_visibility\par
8. occlusion\_obstruction\_crop\_integrity\par
9. corruption\_abnormal\_regions\_overlay\_contamination\par
OUTPUT FORMAT (strict JSON only):\par
Return exactly one valid JSON object, without any preceding text, markdown fences, or trailing explanation.\par
Use pass or fail for all nine quality dimensions. Provide one concise reason for each judgment in the reasons object.\par
\{\par
"caption": "A factual description of the visible scene.",\par
"exposure\_brightness": "pass",\par
"focus\_sharpness": "pass",\par
"motion\_blur\_camera\_shake\_ghosting": "pass",\par
"noise\_grain\_low\_light\_degradation": "pass",\par
"artifacts\_compression\_resampling\_ai\_generated": "pass",\par
"color\_fidelity\_white\_balance\_color\_cast": "pass",\par
"contrast\_dynamic\_range\_detail\_visibility": "pass",\par
"occlusion\_obstruction\_crop\_integrity": "pass",\par
"corruption\_abnormal\_regions\_overlay\_contamination": "pass",\par
"reasons": \{\par
"exposure\_brightness": "The subject is clearly visible.",\par
"focus\_sharpness": "Important details are sharp.",\par
"motion\_blur\_camera\_shake\_ghosting": "No significant blur is visible.",\par
"noise\_grain\_low\_light\_degradation": "No major degradation is visible.",\par
"artifacts\_compression\_resampling\_ai\_generated": "No major artifacts are visible.",\par
"color\_fidelity\_white\_balance\_color\_cast": "Colors appear consistent.",\par
"contrast\_dynamic\_range\_detail\_visibility": "Important details remain visible.",\par
"occlusion\_obstruction\_crop\_integrity": "Important content is unobstructed.",\par
"corruption\_abnormal\_regions\_overlay\_contamination": "No visible corruption is detected."\par
\}\par
\}\par
}
\smallskip\hrule
\end{tcolorbox}
\caption{Prompt for quality assessment.}
\label{fig:quality_assessment_prompt_appendix}
\end{figure}

\subsection{Facility-location Greedy Selection }
\label{app:diversity_aware_keyframe_selection}

We use standard weighted facility-location greedy selection to choose representative frames within each shot.
For each shot, let the candidate-frame set be $F=\{f_i\}_{i=1}^{n}$, and let $k$ denote the selection budget assigned to this shot by the quota planner.
The goal of RepresentativeSelector is to select a representative subset $S$ from the candidate set under the constraint $|S|\le k$, so that the selected frames cover the important candidates within the shot as much as possible.
The objective function is defined as:

\[
\mathcal{F}(S)=\sum_i r_i \max_{j\in S}\mathrm{sim}(i,j),
\]

where $r_i$ denotes the quality weight of candidate frame $f_i$, and $\mathrm{sim}(i,j)$ denotes the similarity between candidate frame $f_i$ and the selected representative frame $f_j$.

The candidate-frame quality weight is computed from the preceding semantic value assessment:

\[
r_i = \mathrm{value\_score}_i .
\]

The value score $\mathrm{value\_score}_i$ is produced by the agent based on semanticized tool signals, while the inter-frame similarity $\mathrm{sim}(i,j)$ is computed from the semantic representations of candidate frames.
Specifically, the system first extracts image embeddings and caption embeddings, normalizes the visual and textual semantic representations, concatenates them, and then computes cosine similarity:

\[
\mathrm{sim}(i,j)=\cos(z_i,z_j).
\]

During selection, the algorithm starts from an empty set $S=\emptyset$ and maintains the current coverage degree $c_i$ of each candidate frame by the selected set.
At each iteration, the algorithm traverses all unselected candidate frames $f_j$ and computes the marginal gain of adding it to the selected set:

\[
\Delta(j)=\sum_i r_i \max\bigl(0,\mathrm{sim}(i,j)-c_i\bigr).
\]

The candidate frame with the largest marginal gain is added to $S$, and the coverage degree is updated as:

\[
c_i \leftarrow \max\bigl(c_i,\mathrm{sim}(i,j)\bigr).
\]

This process is repeated until the shot-level budget $k$ is reached or all candidate frames have been exhausted.

\subsection{Signal Filtering Prompt}
\label{app:signal_filter_prompt}

The detailed signal filtering prompt is shown in Fig.~\ref{fig:signal_filtering_prompt_appendix}.

\begin{figure}[!htbp]
\centering
\begin{tcolorbox}[
    width=0.98\linewidth,
    colback=white,
    colframe=black,
    boxrule=0.4pt,
    arc=0pt,
    left=7pt,
    right=7pt,
    top=7pt,
    bottom=7pt
]
\fontsize{8}{9}\selectfont
\textbf{Signal Filtering Prompt}\par
\medskip\hrule\medskip
{\ttfamily\raggedright\setlength{\parindent}{0pt}\setlength{\parskip}{4pt}%
You are a precise semantic filtering assistant.\par
\vspace{0.35em}
Given a list of semantic tool signals and a shot summary, filter the signals according to their relevance to the shot.\par
\vspace{0.35em}
Keep only signals that are clearly supported by, relevant to, or useful for understanding the shot summary. Remove signals that are irrelevant, redundant, too vague, unsupported, inconsistent, or only weakly related. Be conservative: if relevance is uncertain, remove the signal.\par
\vspace{0.35em}
FILTERING RULES:\par
- Keep signals grounded in the shot summary.\par
- Keep signals that help interpret actions, objects, scene context, intent, spatial relations, or temporal events.\par
- Remove signals not mentioned or implied by the shot summary.\par
- Remove generic signals when a more specific retained signal covers the same meaning.\par
- Remove contradictory signals.\par
- Do not invent new signals.\par
- Do not modify original signal wording except for minor normalization.\par
\vspace{0.35em}
INPUTS:\par
Semantic tool signals:\par
\{\{semantic\_tool\_signals\}\}\par
\vspace{0.35em}
Shot summary:\par
\{\{shot\_summary\}\}\par
\vspace{0.35em}
OUTPUT FORMAT (strict JSON only):\par
\{\par
"filtered\_signals": [\{"signal":"kept signal", "reason":"relevant"\}],\par
"removed\_signals": [\{"signal":"removed signal", "reason":"irrelevant"\}]\par
\}\par
Use original signal text; use [] if empty. Return JSON only.\par
}
\medskip\hrule
\end{tcolorbox}
\caption{Prompt for signal filtering.}
\label{fig:signal_filtering_prompt_appendix}
\end{figure}

\subsection{Gate C}
\label{app:signal_filter}
The detailed Gate C prompt is shown in Fig.~\ref{fig:metadata_admission_prompt_appendix}.

\begin{figure}[!htbp]
\centering
\begin{tcolorbox}[
    width=0.98\linewidth,
    colback=white,
    colframe=black,
    boxrule=0.4pt,
    arc=0pt,
    left=4pt,
    right=4pt,
    top=4pt,
    bottom=4pt
]
\footnotesize
\textbf{Metadata Admission Prompt}\par
\smallskip\hrule\smallskip
{\ttfamily\raggedright\setlength{\parindent}{0pt}\setlength{\parskip}{4pt}%
You are a strict metadata-level admission judge for a video-to-dataset generation pipeline.\par

Decide whether the provided metadata package is suitable for downstream dataset instance generation. Judge the metadata as a whole, using video-level summaries, chapter summaries, shot summaries, frame captions, semanticized tool signals, OCR, ASR, object, motion, UI, transition, and pairwise-change evidence when available.\par

Do NOT judge raw image quality directly unless quality problems are explicitly reflected in the metadata. Judge whether the metadata contains enough grounded, coherent, specific, and useful information to support reliable dataset generation.\par

ADMISSION GOAL:\par
Accept metadata only if it can support meaningful grounded dataset items, such as visual QA, multi-choice QA, reasoning QA, dialogue, temporal/change reasoning, or edit instructions. Reject metadata that is too weak, generic, noisy, contradictory, redundant, or poorly grounded.\par

DECISION LABELS:\par
- accept: suitable for downstream dataset generation.\par
- reject: unsuitable for downstream dataset generation.\par

JUDGMENT CRITERIA:\par
1. Grounded visual content: keep metadata with concrete visible entities, actions, states, layouts, text, UI elements, scene structure, or temporal changes; reject generic, empty, vague, or repetitive metadata.\par
2. Cross-level coherence: keep metadata whose video-, chapter-, shot-, and frame-level descriptions are broadly consistent; reject conflicting evidence that weakens grounding.\par
3. Dataset usefulness: keep metadata that supports at least one useful task type; reject metadata that only yields trivial, obvious, or hallucination-prone items.\par
4. Evidence density: keep metadata with enough distinct evidence points across frames or tools; reject sparse, duplicated, or semantically flat evidence.\par
5. Tool-signal reliability: treat tool signals as auxiliary evidence; keep them when they reinforce captions or summaries; reject incoherent, failed, irrelevant, or contradictory tool evidence.\par
6. Temporal and structural value: keep meaningful progression, state changes, interactions, procedures, spatial layouts, readable text, UI structure, or object relations; reject filler, static background, near-duplicates, or low-information motion.\par
7. Hallucination safety: reject metadata that would require inventing unseen details, intentions, hidden causes, identities, exact text, or precise attributes.\par

OUTPUT FORMAT (strict JSON only):\par
\{\par
"decision": "accept",\par
"primary\_reasons": ["Grounded metadata", "Useful task evidence"]\par
\}\par
Use accept or reject and provide 2--4 reasons. Return JSON only.\par

FIELD RULES:\par
- primary\_reasons must contain 2-4 concise reasons for the decision.\par
}
\smallskip\hrule
\end{tcolorbox}
\caption{Prompt for metadata admission.}
\label{fig:metadata_admission_prompt_appendix}
\end{figure}

\FloatBarrier

To quantify the practical filtering effect of Gate C, we randomly sampled 5,000 metadata entries before filtering. Gate C rejected 37.4\% of the entries overall, while the domain-level rejection rates are summarized in Table~\ref{tab:gate_c_rejection}. The primary rejection reasons included unsupported metadata content, noisy textual or tool-recognition signals, generic descriptions with sparse usable information, repetitive imagery with low evidence density, and inaccurate state-transition descriptions across the five domains, respectively.

\begin{table}[!htbp]
\centering
\small
\caption{Gate C rejection rates by domain.}
\label{tab:gate_c_rejection}
\begin{tabular}{lc}
\toprule
Domain & Rejection Rate (\%) \\
\midrule
Technology and Tutorial & 37.3 \\
Sports and Action & 39.8 \\
Daily Life and Human Activities & 24.8 \\
Nature and Scenery & 47.3 \\
Gaming and ACG & 38.2 \\
\midrule
Overall & 37.4 \\
\bottomrule
\end{tabular}
\end{table}

\section{Dataset Taxonomy Details}
\label{app:dataset_taxonomy}

\begin{figure}[!htbp]
    \centering
    \includegraphics[width=1.0\linewidth]{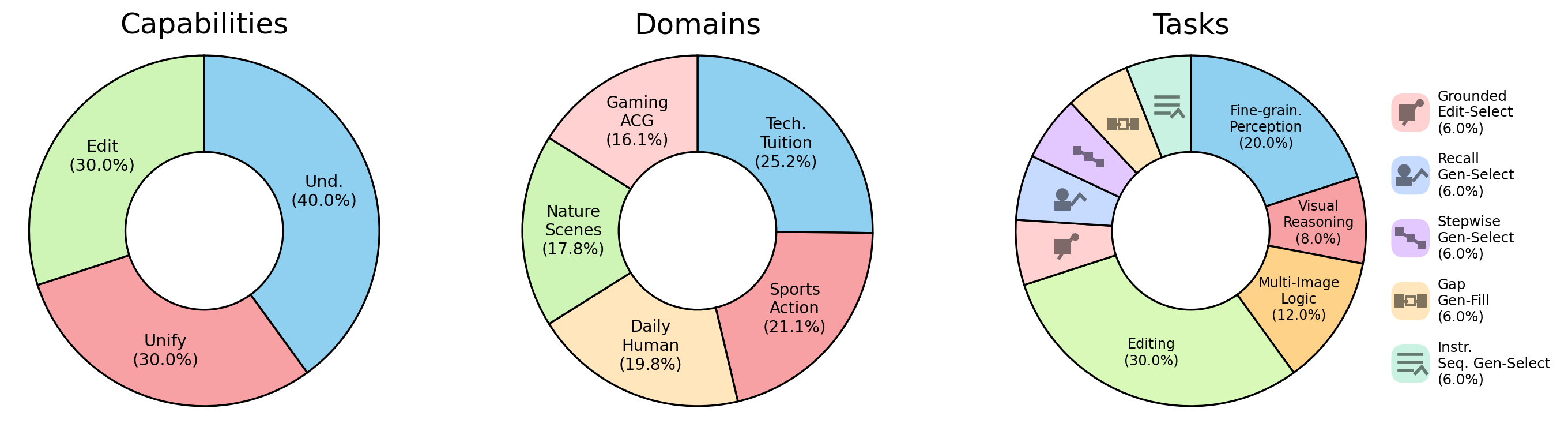}
    \caption{Distribution of the final dataset.}
    \label{fig:data_statistics}
\end{figure}

This section provides detailed definitions of the capability, domain, and task categories used to analyze VidUEU-100K.

\subsection{Capability Categories}
\label{app:capability_categories}

To improve the overall capabilities of multimodal large models, we categorize the 100K instruction-tuning samples into three capability paradigms: understanding, editing, and unified.

\begin{itemize}
\item \textbf{Understanding:} Focuses on multimodal comprehension tasks, including fine-grained visual perception and complex reasoning over visual content.

\item \textbf{Editing:} Focuses on instruction-based image editing, requiring models to modify or complete visual content according to textual instructions while preserving irrelevant visual details.

\item \textbf{Unified:} Involves interleaved text--image sequences, requiring the model to jointly perform multimodal understanding and generation capabilities within a unified stream.
\end{itemize}

\subsection{Domain Categories}
\label{app:domain_categories}

To ensure a balanced representation of both naturalistic and digitally synthesized environments, we classify the dataset into five thematic domains based on contextual semantics.
Within each domain, we maximize content diversity to mitigate representation bias and ensure broad generalization.

\begin{itemize}
\item \textbf{Technology \& Tutorial:} This category includes instructional demonstrations, software interfaces, scientific knowledge, and procedural workflows.

\item \textbf{Sports \& Action:} This category covers athletic competitions, physical training, outdoor sports, and various dynamic action sequences.

\item \textbf{Daily Life \& Human Activities:} This category encompasses routine human activities, including domestic tasks, commuting, and social interactions.

\item \textbf{Nature \& Scenery:} This category includes natural landscapes, meteorological phenomena, and various flora and fauna.

\item \textbf{Gaming \& ACG:} This category consists of gameplay recordings, digital animations, comics, and interactive interface content.

\end{itemize}

\subsection{Task Categories}
\label{app:task_categories}

Based on the generated metadata, the agent adaptively determines and generates nine task types.
The first three tasks belong to Understanding, the fourth corresponds to Editing, and the final five fall under Unified.

\begin{itemize}
\item \textbf{Fine-grained Perception:} Requires the identification of microscopic visual details, including object attributes, existence, cardinality, and precise spatial arrangements within visual scenes.

\item \textbf{Visual Reasoning:} Requires inferring logical dependencies, causal relationships, and relational dynamics between subjects across visual contexts.

\item \textbf{Multi-Image Logic:} Requires maintaining coherent co-reference and logical consistency across a sequence of incrementally introduced images.

\item \textbf{Editing:} Requires completing missing visual content or editing existing images according to textual instructions and provided visual context.

\item \textbf{Grounded Edit-to-Select Generation:} Requires generating a target visual state from grounded source--target changes and selecting the candidate image that matches the intended edit.

\item \textbf{Interleaved Understand--Recall--Generate--Select:} Requires extracting visible facts from earlier visual context, recalling a later state, generating the recalled visual state, and selecting the matching candidate frame.

\item \textbf{Stepwise Sequence Generate-and-Select:} Requires decomposing a coherent visual process into ordered steps, generating intermediate visual states, and selecting candidates that match each step.

\item \textbf{Middle-Step Gap Generate-and-Fill:} Requires reasoning over earlier and later frames to infer, generate, and select the missing middle visual state.

\item \textbf{Instructional Sequential Generate-and-Select:} Requires inferring future visual changes from context frames, generating stepwise future states, and selecting candidates that satisfy each predicted step.

\end{itemize}
\subsection{Data Statistics}

The VidUEU dataset contains 100,000 instruction samples, comprising 40,000 understanding samples, 30,000 editing samples, and 30,000 unified interleaved multimodal generation samples.
The dataset is derived from 8,268 raw videos, covering approximately 224.2 hours of measurable segment-level video duration and containing approximately 8,505 semantic segments.
The detailed distributions of capabilities, domains, and task types are shown in Figure~\ref{fig:data_statistics}.

\subsection{Intrinsic Data-Quality Audit}
\label{app:human_audit}

We randomly sampled 800 instances for human evaluation, with 200 instances per quality dimension. Two annotators independently rated each instance on a 0--5 scale. Table~\ref{tab:human_audit} reports the annotator scores, their means, and Cohen's $\kappa$ for inter-annotator agreement. The first three dimensions were assessed at the metadata level, while editing-sample validity was evaluated at the individual editing-instruction level.

\begin{table}[!htbp]
\centering
\small
\caption{Human audit of synthesized-data quality.}
\label{tab:human_audit}
\begin{tabular}{lcccc}
\toprule
Quality Dimension & Annotator A & Annotator B & Mean & Cohen's $\kappa$ \\
\midrule
Gate C Metadata Validity & 4.12 & 4.08 & 4.10 & 0.69 \\
Keyframe--Tool--Signal Alignment & 4.30 & 4.20 & 4.25 & 0.76 \\
Transition Correctness & 4.06 & 4.01 & 4.04 & 0.77 \\
Editing-Sample Validity & 3.96 & 4.02 & 3.99 & 0.71 \\
\bottomrule
\end{tabular}
\end{table}

\section{Training and Evaluation Details}
\label{app:f1_training_configuration}

\subsection{Training Details}
\label{app:training_details}

\textbf{InternVL3-8B.} For understanding, we fine-tune InternVL3-8B on the understanding subset.
We freeze the vision encoder and apply LoRA fine-tuning to the remaining target modules.
The model is trained for 2 epochs with AdamW, a learning rate of $5\times10^{-5}$, a global batch size of 32, and a cosine learning rate scheduler with a warmup ratio of 0.03.
Training uses 2 NVIDIA A100-40G with DeepSpeed ZeRO stage 2, bfloat16 precision, and gradient checkpointing.

\textbf{Bagel-7B-MoT.} For understanding, editing and unified tasks, we fine-tune Bagel-7B-MoT on the understanding, editing and unified subsets, respectively.
We freeze the VAE encoder, the VAE decoder, and the ViT encoder, apply LoRA to the LLM backbone with $r=16$ and $\alpha=32$. For editing and unified training, we additionally fully fine-tune the generation bridging heads, including \texttt{llm2vae}, \texttt{vae2llm}, and \texttt{time\_embedder}.
The model is trained for 1 epoch with AdamW, a learning rate of $5\times10^{-5}$, and a cosine learning rate scheduler, using cross-entropy loss for text supervision and MSE loss for image supervision.
Training uses 8 A100-40GB GPUs.

\subsection{Evaluation Details}
\label{app:evaluation_details}

We evaluate the fine-tuned models under three capability settings, namely understanding, editing, and unified understanding and generation.
We follow the official protocol of each benchmark.

\subsection{Compared Baselines}
For comparison, we report results against representative understanding and unified baselines, including InternVL3 with other model sizes~\cite{zhu2025internvl3exploringadvancedtraining}, Qwen2.5-VL~\cite{bai2025qwen25vltechnicalreport}, LLaMA-3.2~\cite{grattafiori2024llama3herdmodels}, Janus-Pro~\cite{chen2025janusprounifiedmultimodalunderstanding}, and TokenFlow-XL~\cite{qu2024tokenflowunifiedimagetokenizer}.
We also compare with editing and unified baselines, including Instruct-Pix2Pix~\cite{brooks2023instructpix2pixlearningfollowimage}, MagicBrush~\cite{zhang2024magicbrushmanuallyannotateddataset}, AnyEdit~\cite{yu2024anyedit}, Step1X-Edit~\cite{liu2025step1xeditpracticalframeworkgeneral}, Gemini-2.0-flash~\cite{gemini2flashimage2025}, OmniGen~\cite{xiao2024omnigenunifiedimagegeneration}, OmniGen2~\cite{wu2025omnigen2}, Ovis-U1~\cite{wang2025ovisu1technicalreport}, and Bagel-Zebra~\cite{li2026zebracot}.

\subsection{Evaluation Protocols}

\textbf{Understanding evaluation.} 
For LLaVA-Bench, we use \texttt{Qwen3.5-35B-A3B}~\cite{qwen3.5} as the VLM judge.
For MMDU, we use \texttt{Qwen3-VL-235B-A22B-Instruct}~\cite{bai2025qwen3} as the VLM judge.
   
\textbf{Editing evaluation.} The editing model is evaluated on GEdit-Bench-En.
GEdit-Bench-En is used to evaluate instruction-based image editing, with reported dimensions including G\_SC, G\_PQ, G\_O, and Average.
We use \texttt{GPT-4.1}~\cite{openai2025gpt41} as the judge for the G\_SC, G\_PQ, and G\_O metrics.

\textbf{Unified evaluation.} The unified model is evaluated on WEAVEbench. We use \texttt{GPT-4o} \cite{openai2024gpt4ocard} as the VLM judge.

\subsection{Evaluation Variability}
\label{app:automatic_judge_stability}

Table~\ref{tab:automatic_judge_stability} reports the run-to-run standard deviation of VLM-as-a-judge evaluations.
For each entry, the evaluated model outputs are fixed, and the standard deviation is computed over three independent judge runs.

\begin{table}[!htbp]
  \caption{Automatic-judge run-to-run standard deviation with fixed evaluated outputs. Mean scores are reported in the main experimental tables.}
  \label{tab:automatic_judge_stability}
  \centering
  \small
  \setlength{\tabcolsep}{5.0pt}
  \begin{tabular}{@{}llcc@{}}
    \toprule
    Capability & Model / metric & Baseline std & Ours std \\
    \midrule
    Understanding & InternVL3, LLaVA-Bench & 0.317 & 0.292 \\
    Understanding & InternVL3, MMDU & 0.150 &  0.172 \\
    Understanding & Bagel, LLaVA-Bench & 0.306 & 0.328 \\
    Understanding & Bagel, MMDU &  0.133 &  0.151 \\
    \midrule
    Editing & Bagel, GEdit-Bench-En G\_SC & 0.073 & 0.157 \\
    Editing & Bagel, GEdit-Bench-En G\_PQ & 0.076 & 0.146 \\
    Editing & Bagel, GEdit-Bench-En G\_O & 0.072 & 0.134 \\
    \midrule
    Unified & Bagel, WEAVE Science & 0.0106 & 0.0108 \\
    Unified & Bagel, WEAVE Creation & 0.0053 & 0.0051 \\
    Unified & Bagel, WEAVE Logic & 0.0119 & 0.0119 \\
    Unified & Bagel, WEAVE Game & 0.0230 & 0.0217 \\
    Unified & Bagel, WEAVE Average & 0.0129 & 0.0122 \\
    \bottomrule
  \end{tabular}
\end{table}

\section{Full Results for Autonomous Iteration and Ablation}
\label{app:iteration_ablation_full}

\subsection{Full Results of Autonomous Data Iteration}
\label{app:autonomous_iteration_full}

Table~\ref{tab:autonomous_iteration_full} reports the full results for Iterations 1--4 under 30K joint training, together with the Final results from capability-specific training on VidUEU-100K. The Final joint-training results are reported in Table~\ref{tab:autonomous_iteration}.

\begin{table}
  \caption{Full results of autonomous data iteration using Bagel.}
  \label{tab:autonomous_iteration_full}
  \centering
  \makebox[\linewidth][c]{%
  \begin{tabular}{@{}lcccccc@{}}
    \toprule
    Setting & MMT-Bench$\uparrow$ & MME$\uparrow$ & LLaVA-Bench$\uparrow$ & MMDU$\uparrow$ & GEdit G\_O$\uparrow$ & WEAVE Overall$\uparrow$ \\
    \midrule
    Iteration 1 & 0.631 & 1725.4 & 52.3 & 29.7 & 6.541 & 0.448 \\
    Iteration 2 & 0.629 & 1767.3 & 52.3 & 29.2 & 6.796 & 0.452 \\
    Iteration 3 & 0.642 & 1840.4 & 56.1 & 31.4 & 6.873 & 0.458 \\
    Iteration 4 & 0.649 & 1871.9 & 57.2 & 32.8 & 6.901 & 0.499 \\
    \midrule
    \multicolumn{1}{@{}>{\columncolor{green!12}[0pt][\tabcolsep]}l}{Final (separate)}
    & \multicolumn{1}{>{\columncolor{green!12}}c}{0.651}
    & \multicolumn{1}{>{\columncolor{green!12}}c}{1900.2}
    & \multicolumn{1}{>{\columncolor{green!12}}c}{57.9}
    & \multicolumn{1}{>{\columncolor{green!12}}c}{33.2}
    & \multicolumn{1}{>{\columncolor{green!12}}c}{6.951}
    & \multicolumn{1}{>{\columncolor{green!12}[\tabcolsep][0pt]}c@{}}{0.501} \\
    \bottomrule
  \end{tabular}%
  }
\end{table}

\subsection{Training Protocol and Data Distributions}
\label{app:iteration_training_protocol}

Iterations 1--4 each use a fixed budget of 30K samples. At each iteration, the agent randomly samples from the generated data according to the updated target proportions of capabilities, domains, and task types. Samples exceeding category quotas are removed, while additional samples are generated for underrepresented categories to form the 30K training set. Each iteration jointly fine-tunes BAGEL from the same initial checkpoint for one epoch on eight A100-40GB GPUs. The Final 100K dataset is constructed by scaling the target distribution obtained in Iteration 4.

The agent prioritizes the capability distribution, allowing deviations of at most one percentage point for domain and task-type distributions. The realized distributions across iterations are reported in Tables~\ref{tab:iteration_capability_distribution}, \ref{tab:iteration_domain_distribution}, and \ref{tab:iteration_task_distribution}.
At each iteration, the agent receives training loss and learning curves, aggregate benchmark and subcategory scores, the current data distribution, and compressed historical feedback and distributions, with optional human-audit inputs.
Baseline performance is provided only in the first iteration as a fixed reference, while the other feedback signals are updated after each round.

\begin{table}[!htbp]
\centering
\small
\caption{Realized capability distributions (\%) across autonomous iterations.}
\label{tab:iteration_capability_distribution}
\begin{tabular}{lccc}
\toprule
Setting & Understanding & Editing & Unified \\
\midrule
Iteration 1 & 50.0 & 25.0 & 25.0 \\
Iteration 2 & 35.0 & 33.0 & 32.0 \\
Iteration 3 & 39.0 & 31.0 & 30.0 \\
Iteration 4 & 40.0 & 30.0 & 30.0 \\
\bottomrule
\end{tabular}
\end{table}

\begin{table}[!htbp]
\centering
\small
\caption{Realized domain distributions (\%) across autonomous iterations.}
\label{tab:iteration_domain_distribution}
\begin{tabular}{lccccc}
\toprule
Setting & Technology & Sports & Daily Life & Nature & Gaming \\
\midrule
Iteration 1 & 22.1 & 17.8 & 28.2 & 18.1 & 13.8 \\
Iteration 2 & 18.1 & 21.9 & 24.8 & 16.1 & 19.1 \\
Iteration 3 & 23.9 & 20.2 & 21.9 & 17.9 & 16.1 \\
Iteration 4 & 25.2 & 21.1 & 19.8 & 17.8 & 16.1 \\
\bottomrule
\end{tabular}
\end{table}

\begin{table}[!htbp]
\centering
\scriptsize
\caption{Realized task-type distributions (\%) across autonomous iterations.}
\label{tab:iteration_task_distribution}
\begin{tabular}{lccccccccc}
\toprule
Setting & FP & VR & MIL & Edit & GES & RGS & SGS & GGF & ISG \\
\midrule
Iteration 1 & 28.3 & 9.6 & 12.1 & 25.0 & 4.6 & 5.3 & 4.8 & 5.1 & 5.2 \\
Iteration 2 & 22.4 & 6.8 & 5.8 & 33.0 & 9.0 & 6.4 & 6.1 & 4.8 & 5.7 \\
Iteration 3 & 20.8 & 8.3 & 9.9 & 31.0 & 7.0 & 5.6 & 5.9 & 5.3 & 6.2 \\
Iteration 4 & 19.8 & 8.3 & 11.9 & 30.0 & 5.6 & 6.3 & 5.8 & 6.2 & 6.1 \\
\bottomrule
\end{tabular}
\end{table}

\subsection{Feedback-Free Evaluation}
\label{app:feedback_free_evaluation}

MMBench, MMStar, and VSI-Bench are excluded from iterative feedback and used to assess generalization beyond the feedback benchmarks. As shown in Table~\ref{tab:feedback_free_evaluation}, the results exhibit an overall upward trend across iterations.

\begin{table}[!htbp]
\centering
\small
\caption{Evaluation on benchmarks excluded from autonomous feedback.}
\label{tab:feedback_free_evaluation}
\begin{tabular}{lccc}
\toprule
Setting & MMBench$\uparrow$ & MMStar$\uparrow$ & VSI-Bench$\uparrow$ \\
\midrule
BAGEL Base & 84.2 & 67.3 & 31.4 \\
Iteration 1 & 84.9 & 67.4 & 32.8 \\
Iteration 2 & 84.7 & 67.3 & 32.9 \\
Iteration 3 & 85.0 & 67.5 & 33.6 \\
Iteration 4 & 85.1 & 67.6 & 34.4 \\
\midrule
Final & \textbf{85.3} & \textbf{67.8} & \textbf{36.8} \\
\bottomrule
\end{tabular}
\end{table}

\subsection{Full Ablation Study}
\label{app:ablation_vidueu_agent_full}

Table~\ref{tab:ablation_vidueu_agent_full} reports the full ablation results of VidUEU-Agent components, including the additional understanding benchmarks omitted from the main table.
The results show that removing any major component degrades performance, validating the effectiveness of the full pipeline design.

\begin{table}
  \caption{Full ablation study of VidUEU-Agent components using Bagel.}
  \label{tab:ablation_vidueu_agent_full}
  \centering
  \small
  \setlength{\tabcolsep}{3pt}
  \makebox[\linewidth][c]{%
  \begin{tabular}{@{}lcccccc@{}}
    \toprule
    Setting & MMT-Bench$\uparrow$ & MME$\uparrow$ & LLaVA-Bench$\uparrow$ & MMDU$\uparrow$ & GEdit G\_O$\uparrow$ & WEAVE Overall$\uparrow$ \\
    \midrule
    \multicolumn{1}{@{}>{\columncolor{green!12}[0pt][\tabcolsep]}l}{Ours}
    & \multicolumn{1}{>{\columncolor{green!12}}c}{0.651}
    & \multicolumn{1}{>{\columncolor{green!12}}c}{1900.2}
    & \multicolumn{1}{>{\columncolor{green!12}}c}{57.9}
    & \multicolumn{1}{>{\columncolor{green!12}}c}{33.2}
    & \multicolumn{1}{>{\columncolor{green!12}}c}{6.951}
    & \multicolumn{1}{>{\columncolor{green!12}[\tabcolsep][0pt]}c@{}}{0.501} \\
    \midrule
    w/o A & 0.638 & 1772.6 & 53.6 & 32.4 & 6.773 & 0.497 \\
    w/o B & 0.605 & 1255.7 & 42.8 & 24.7 & 5.175 & 0.404 \\
    w/o S & 0.623 & 1684.3 & 50.7 & 30.8 & 5.848 & 0.462 \\
    w/o C & 0.617 & 1549.4 & 51.2 & 30.5 & 6.303 & 0.454 \\
    N & 0.537 & 1116.4 & 38.7 & 22.0 & 5.041 & 0.393 \\
    w/o R & 0.611 & 1565.5 & 51.0 & 29.4 & 6.031 & 0.470 \\
    \bottomrule
  \end{tabular}%
  }
\end{table}

Given the computational demands of multi-stage data synthesis and model training, we evaluate six ablated variants using 10K samples per capability.

We retain the capability-specific Final models trained on the 100K dataset as the full-pipeline reference, enabling broader component coverage without repeated large-scale training.

These comparisons provide indicative evidence of component-level contributions under practical data budgets, while leaving strictly budget-matched evaluation for future investigation.

\subsection{Comparison with Baseline Datasets}
\label{app:baseline_dataset_comparison}

We compare Bagel under the same training setting using our dataset and three task-specific baseline datasets: LLaVA Visual Instruct 150K~\cite{liu2023visual} for understanding, InstructPix2Pix~\cite{brooks2023instructpix2pixlearningfollowimage} for editing, and Zebra-CoT~\cite{li2026zebracot} for unified tasks, as shown in Table~\ref{tab:baseline_dataset_comparison}.
Our dataset achieves the best results across all evaluated settings, indicating stronger overall training effectiveness.
\begin{table}[!htbp]
  \caption{Comparison with baseline datasets using Bagel under the same training setting.}
  \label{tab:baseline_dataset_comparison}
  \centering
  \small
  \setlength{\tabcolsep}{3pt}
  \makebox[\linewidth][c]{%
  \begin{tabular}{@{}lcccccc@{}}
    \toprule
    Dataset & MMT-Bench$\uparrow$ & MME$\uparrow$ & LLaVA-Bench$\uparrow$ & MMDU$\uparrow$ & GEdit G\_O$\uparrow$ & WEAVE Overall$\uparrow$ \\
    \midrule
    LLaVA Visual Instruct & 0.477 & 1823.0 & 51.8 & 29.6 & -- & -- \\
    InstructPix2Pix & -- & -- & -- & -- & 6.882 & -- \\
    Zebra-CoT & -- & -- & -- & -- & -- & 0.449 \\
    \midrule
    \multicolumn{1}{@{}>{\columncolor{green!12}[0pt][\tabcolsep]}l}{Ours(Bagel)}
    & \multicolumn{1}{>{\columncolor{green!12}}c}{\underline{\textbf{0.651}}}
    & \multicolumn{1}{>{\columncolor{green!12}}c}{\underline{\textbf{1900.2}}}
    & \multicolumn{1}{>{\columncolor{green!12}}c}{\underline{\textbf{57.9}}}
    & \multicolumn{1}{>{\columncolor{green!12}}c}{\underline{\textbf{33.2}}}
    & \multicolumn{1}{>{\columncolor{green!12}}c}{\underline{\textbf{6.951}}}
    & \multicolumn{1}{>{\columncolor{green!12}[\tabcolsep][0pt]}c@{}}{\underline{\textbf{0.501}}} \\
    \bottomrule
  \end{tabular}%
  }
\end{table}

\section{Broader Impacts}
\label{sec:broader_impacts}
VidUEU-Agent and VidUEU-100K may help reduce the cost of constructing high-quality multimodal instruction data and support the development of models with stronger understanding, editing, and unified capabilities.
However, even with careful human screening, source videos may contain social biases, privacy-sensitive content, or copyright risks, which may still be inherited by the constructed dataset and downstream models.
Therefore, future use of VidUEU-Agent should include careful data filtering, license review, privacy protection, and safeguards against malicious use.

\section{Limitations}
\label{sec:limitations}
Although VidUEU-Agent covers multimodal understanding, editing, and unified tasks, it does not explicitly construct pure text-to-image generation data.
This is because the advantage of video sources lies in temporal continuity and state transitions; for pure text-to-image generation, which mainly requires isolated text--image alignment, directly curating image data would be more suitable and less costly than using videos as the primary source.
In addition, due to limited computational and annotation resources, we scale the dataset to 100K instruction samples, which remains modest compared with web-scale multimodal corpora. 
A larger-scale version may further improve data diversity, long-tail coverage, and the stability of multimodal training.


\end{document}